\documentclass[11pt]{article}

\usepackage[final]{acl}

\usepackage{times}
\usepackage{latexsym}
\usepackage[T1]{fontenc}
\usepackage[utf8]{inputenc}
\usepackage{microtype}
\usepackage{inconsolata}

\usepackage{graphicx}
\usepackage{subcaption}
\usepackage{booktabs}
\usepackage{xurl}
\usepackage{amsfonts}
\usepackage{nicefrac}
\usepackage{longtable}
\usepackage{array}
\usepackage{arydshln}
\usepackage[table,dvipsnames]{xcolor}
\usepackage{amsmath}
\usepackage{amssymb}
\usepackage{mathtools}
\usepackage{amsthm}
\usepackage{multirow}
\usepackage{bm}
\usepackage[normalem]{ulem}
\usepackage{listings}
\usepackage[most]{tcolorbox}
\tcbuselibrary{skins, breakable}
\usepackage{enumitem}
\usepackage{dashrule}
\usepackage{makecell}
\usepackage[capitalize,noabbrev]{cleveref}
\usepackage[disable,textsize=tiny]{todonotes}

\definecolor{lightgray}{gray}{0.95}
\definecolor{boxFrameGreen}{HTML}{5E9969}
\definecolor{boxTitleBack}{HTML}{89B890}
\definecolor{listNumberBlue}{HTML}{2E74B5}
\definecolor{promptframecolor}{RGB}{102,102,178}
\definecolor{promptbackcolor}{RGB}{244,251,254}
\definecolor{skyblue}{RGB}{0,102,204}

\newcommand{\ourdataset}{FullFront}

\newenvironment{wideprompt}{%
  \begin{minipage}{0.98\textwidth}\scriptsize
}{%
  \end{minipage}
}

\newtcolorbox{promptbox}[2][]{
  enhanced,
  breakable=false,
  colback=promptbackcolor,
  colframe=promptframecolor,
  fonttitle=\bfseries,
  coltitle=white,
  title=#2,
  before skip=4pt,
  after skip=4pt,
  left=6pt,
  right=6pt,
  top=5pt,
  bottom=5pt,
  #1
}

\theoremstyle{plain}

\theoremstyle{definition}

\theoremstyle{remark}

\title{FullFront: Benchmarking MLLMs Across the Full Front-End Engineering Workflow}

\author{
  \textbf{Haoyu Sun}\textsuperscript{1,2},
  \textbf{Huichen Will Wang}\textsuperscript{3},
  \textbf{Jiawei Gu}\textsuperscript{3},
  \textbf{Linjie Li}\textsuperscript{3},
  \textbf{Yu Cheng}\textsuperscript{4,2,\textdagger}
  \\[0.6em]
  \normalfont
  \textsuperscript{1}Shanghai Research Institute for Intelligent Autonomous Systems,\\
  Tongji University, Shanghai, China \\[0.2em]
  \textsuperscript{2}Shanghai AI Laboratory \quad
  \textsuperscript{3}University of Washington \\
  \textsuperscript{4}Nanyang Technological University
  \\[0.6em]
  \normalfont\small
  \textsuperscript{\textdagger}Corresponding author
}

\begin{document}
\maketitle

\begin{abstract}
Front-end engineering involves a complex workflow where engineers conceptualize designs, translate them into code, and iteratively refine the implementation. While recent benchmarks focus on converting visual designs to code, we present FullFront, a benchmark to evaluate Multimodal Large Language Models \textbf{across the full front-end development pipeline}. FullFront assesses three fundamental tasks that map directly to the front-end engineering pipeline: Webpage Design (conceptualization), Webpage Perception QA (comprehension of visual organization and elements), and Webpage Code Generation (implementation). Unlike existing benchmarks relying on scraped websites with bloated code or oversimplified LLM-generated HTML, FullFront employs a novel, two-stage process to transform real-world webpages into clean, standardized HTML while maintaining diverse visual designs and avoiding copyright issues. Extensive testing of state-of-the-art MLLMs reveals significant limitations in page perception, code generation (particularly for image handling and layout), and interaction implementation. Our results quantify performance disparities across models and tasks, highlighting a substantial gap between current MLLM capabilities and human expert performance in front-end engineering.
\end{abstract}

\begin{figure*}[t] 
  \centering 
  \includegraphics[width=0.9\linewidth]{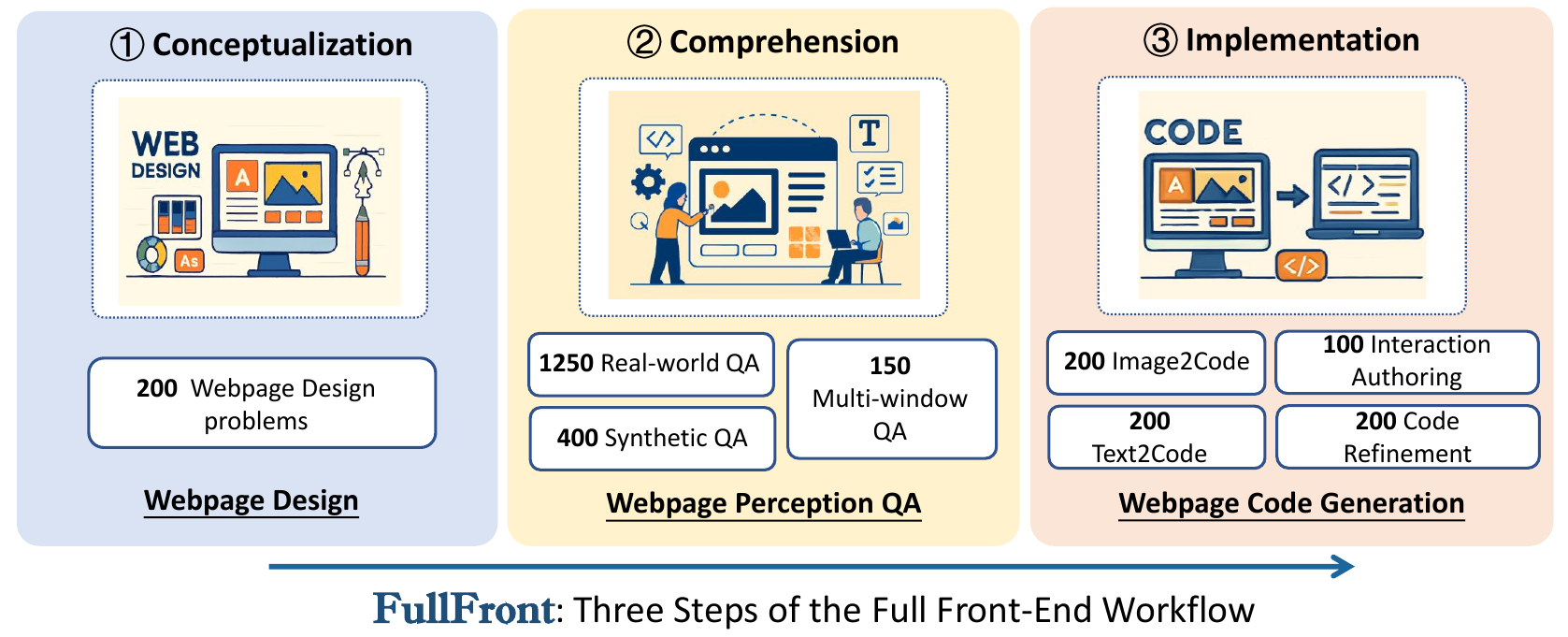}
  \caption{Mapping the full front-end workflow to FullFront's tasks: (1) Conceptualization assessed by Webpage Design, (2) Comprehension by Webpage Perception QA, and (3) Implementation by Webpage Code Generation.} 
  \label{fig:teaser}
  \vspace{-14pt}
\end{figure*}

\section{Introduction}

Front-end engineering, a cornerstone of the modern digital experience, is an intricate process, as depicted in Figure \ref{fig:teaser}. It transforms abstract concepts into initial designs (conceptualization), involves detailed visual comprehension (perception), and culminates in functional, interactive code (implementation) for web applications. This field is poised for significant transformation with the advent of Multimodal Large Language Models (MLLMs), whose capabilities in processing visual information and generating code offer compelling potential to streamline and even automate the front-end development, aligning with the aspirational goal of an ``idea-to-design-to-code'' paradigm.


Despite this burgeoning potential, a benchmark assessing MLLMs across the full front-end engineering workflow is conspicuously absent. Instead, current evaluations separately address crucial yet distinct capabilities: vision perception and code generation. For instance, benchmarks like IW-Bench~\citep{DBLP:conf/acl/GuoZCGYDCHLMZ025} and WebCode2M~\citep{DBLP:conf/www/GuiL000CSC0Z00025} scrutinize MLLMs' code generation from visual inputs but often have narrow task scopes, overlooking vital aspects such as implementing interactive features or refining existing codebases. Conversely, while WebQuest~\citep{wang2024webquest} and Webqa~\citep{chang2022webqa} investigate MLLMs' visual understanding of webpages, the focus remains on content-level reasoning, neglecting fine-grained perceptual acuity concerning element size, positioning, and layout, which is indispensable for accurate front-end implementation. Most critically, these fragmented approaches generally omit the initial conceptual ``design'' phase, and therefore fall short of gauging MLLM proficiency in end-to-end front-end engineering.

\begin{figure*}[!tb]
    \centering
    \includegraphics[width=\textwidth]{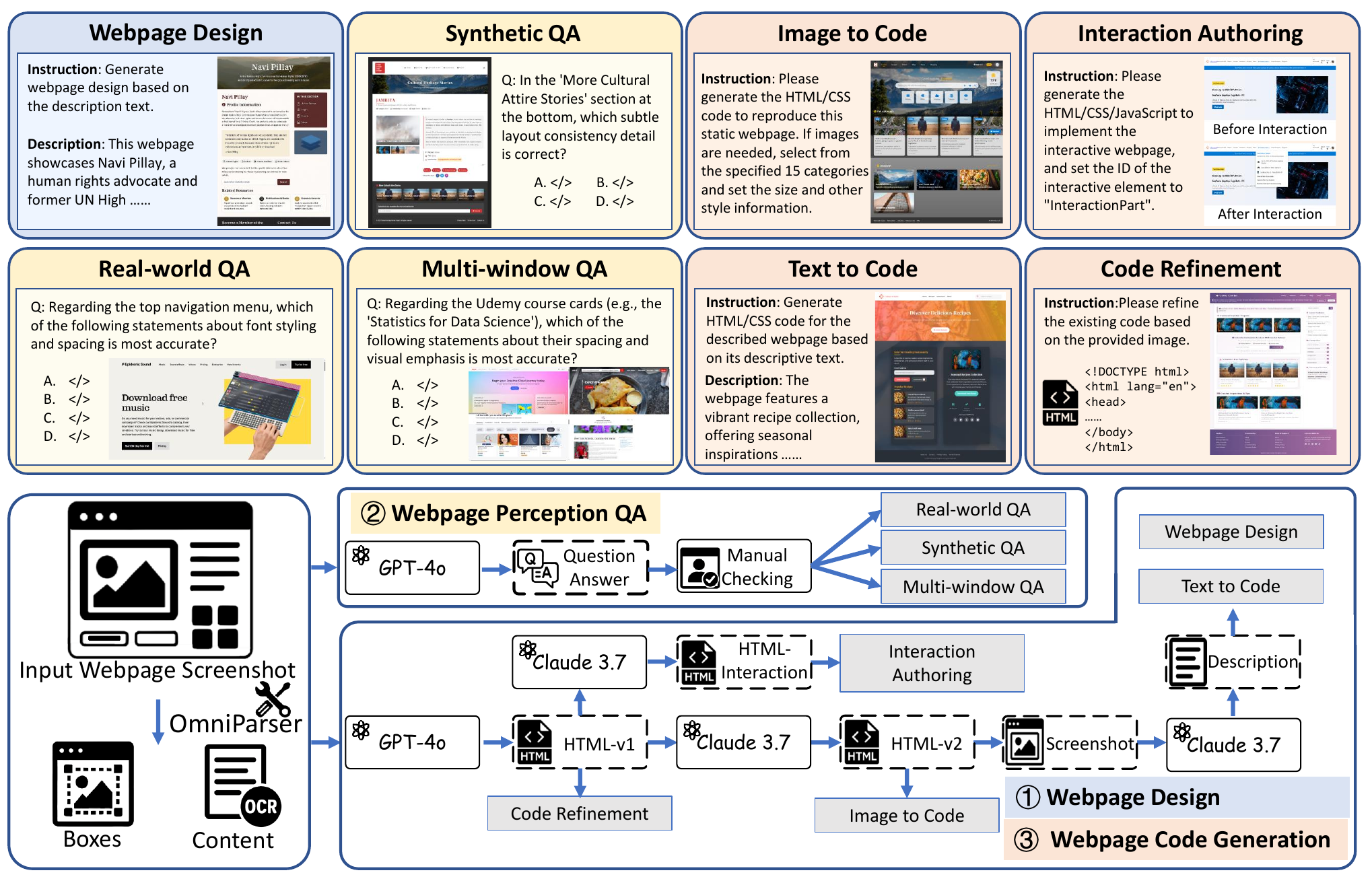}
    \caption{Overview of the eight subtasks \ourdataset~covers and our data construction pipeline.}
    \label{fig:overview}
    \vspace{-14pt}
\end{figure*}

In this work, we introduce \textbf{FullFront}, a benchmark meticulously designed to evaluate MLLMs across the full front-end engineering workflow. As depicted in Figure \ref{fig:overview}, FullFront distinctively offers a holistic assessment through three core tasks:
(1) \textbf{Webpage Design} (200 problems), which assesses the model's ability to structure and organize visual elements to present some given content;
(2) \textbf{Webpage Perception QA} (three subtasks and 1,800 multiple-choice questions), which evaluates the perception of visual organization, element characteristics, and spatial relationships within a webpage; and
(3) \textbf{Webpage Code Generation} (four subtasks and 700 code generation problems), which focuses on the accurate translation of visual designs into functional code, including interaction implementation and code refinement. We collect real-world webpages and develop an MLLM-driven pipeline to reconstruct them into clean, standardized, and copyright-free HTML, ensuring high controllability while preserving original visual diversity for robust benchmark data.
This comprehensive task structure and our evaluation framework, incorporating fine-grained visual similarity scores and detailed code-level metrics (including structural and content-based comparisons), provide a multifaceted and robust assessment of model capabilities across the full front-end engineering workflow.

Benchmarking state-of-the-art open-source and proprietary MLLMs with FullFront reveals significant challenges across the board. In the Webpage Design task, current text-to-image MLLMs demonstrate an ability to produce general layout concepts but lack the precision for high-fidelity webpage designs that accurately reflect detailed textual descriptions. In Webpage Perception QA, even leading models struggle to achieve human-comparable accuracy; for instance, the best-performing model, GPT-5, averages below 60\% across the three full-set QA tasks. On the separate FullFront-mini subset, human experts achieve 97\%, 96\%, and 94\% accuracy on Real-world, Synthetic, and Multi-window QA, respectively. Our analysis reveals that MLLMs face considerable difficulties in accurately perceiving element alignment, size, and positioning within webpages. For Webpage Code Generation, while proprietary models like Claude 3.7 Sonnet and GPT-4.1 generally outperform open-source alternatives, they still encounter difficulties, particularly in accurately handling complex front-end details such as image manipulation, layout fidelity, and interaction implementation. These findings underscore the critical need to enhance current MLLM capabilities within the front-end development workflow to bridge the substantial gap between their current performance and the requirements for expert-level engineering. In summary, our main contributions are as follows:
\begin{itemize}[leftmargin=9pt, itemsep=3pt, parsep=0pt, topsep=0pt]
    \item \textbf{Comprehensive Full Front-End Workflow Benchmark}: We propose FullFront, a benchmark that unifies Webpage Design (conceptualization), Perception QA (comprehension), and Code Generation (implementation) into a cohesive pipeline. This framework simulates the engineering lifecycle to evaluate current MLLMs and future Generalist Front-End Agents.

    \item \textbf{Robust and Bias-Resilient Evaluation}: We introduce fine-grained visual- and code-level metrics rigorously validated against human judgment. Extensive ablation studies confirm our framework's robustness against data-pipeline artifacts and model-as-judge biases, establishing a standard for cross-model comparison.

    \item \textbf{In-depth Analysis of Perception-Generation Decoupling}: Beyond performance rankings, our evaluation uncovers a critical ``Perception-Implementation Gap.'' We reveal that top-tier MLLMs often exhibit ``Grounding Failures''---relying on strong internal priors to generate correct code despite flawed perception. This insight challenges the assumption that code generation implies visual understanding, a nuance missed by isolated benchmarks.
\end{itemize}

\section{Related Work}

\paragraph{Applications of MLLMs in Web}
MLLMs have recently attracted growing attention in web applications~\citep{zhao2025worldgui, wu2024copilot, wang2025mobile}. Recent studies enable MLLMs to navigate and manipulate websites according to user instructions~\citep{zheng2024webagent, cheng2024seeclick}. For example, Mind2Web~\citep{deng2023mind2web} trains generalist web agents on diverse web tasks, demonstrating their ability to follow complex natural language commands across websites. WinClick~\citep{DBLP:journals/corr/abs-2503-04730} further studies GUI grounding, enabling more precise interaction with web elements by modeling their visual and textual properties. These efforts reflect a broader trend toward autonomous web interaction agents.

\paragraph{Webpage Benchmarks and Datasets}
Several benchmarks evaluate MLLMs on webpage-related tasks. A large body of work~\citep{wang2024webquest, chang2022webqa, DBLP:conf/acl/0007Y0WXFZ0ZXH25} uses real-world webpages to assess element grounding and content reasoning through QA. ScreenWords~\citep{wang2021screen2words} studies screen summarization, while VisualWebBench~\citep{liu2024visualwebbench} provides seven QA tasks for broader webpage understanding. Another line of work evaluates front-end code generation from screenshots. Design2Code~\citep{si2024design2code}, WebCode2M~\citep{DBLP:conf/www/GuiL000CSC0Z00025}, and IW-Bench~\citep{DBLP:conf/acl/GuoZCGYDCHLMZ025} construct datasets by scraping and simplifying existing code, whereas Web2Code~\citep{DBLP:conf/nips/YunLTBWJDWTL0NB24}, WebSight~\citep{laurenccon2024websight}, and Pix2Code~\citep{beltramelli2018pix2code} rely on generated code or synthetic UIs. Other studies extend this direction with external-resource lists~\citep{wan2024mrweb}, dynamic webpage generation~\citep{xiao2024interaction2code}, and multi-framework iterative editing and repair~\citep{xiao2025designbench}.


\begin{figure*}[t]
  \centering
  \includegraphics[width=1.0\linewidth]{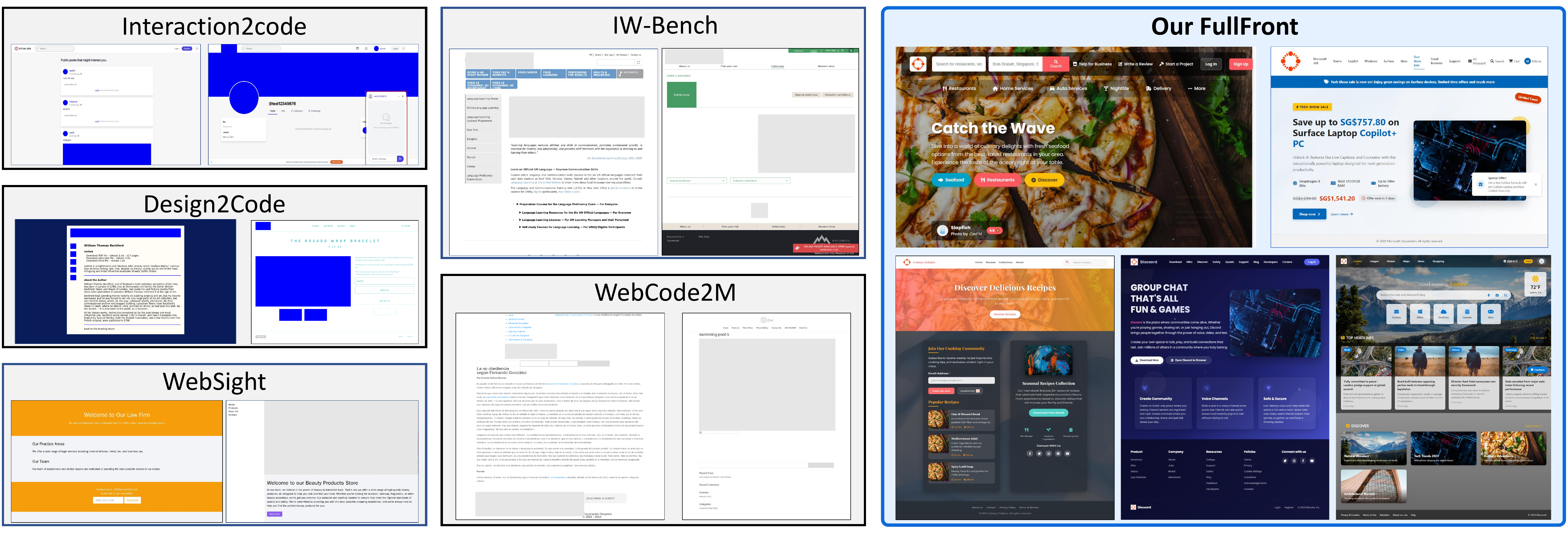}
  \caption{Comparison of the screenshots used in FullFront for webpage code generation tasks with those of other benchmarks. We are the first to not use a single image placeholder or random images.}
  \label{fig:datasets_comparison}
  \vspace{-12pt}
\end{figure*}

\section{Benchmark}

\subsection{Data Curation}

\paragraph{Webpage Design}

The Webpage Design task aims to evaluate text-to-image generation MLLMs as webpage designers. We provide 200 textual descriptions of synthetic webpages sampled from the Text to Code task dataset (see below). MLLMs are required to generate webpage design images based on these descriptions. This process tests how effectively models can transform textual requirements into visual designs, including their understanding of webpage layouts and element relationships. Since textual descriptions naturally cannot capture all visual design nuances, this task also assesses models' ability to make reasonable design decisions where specifications are incomplete.

\paragraph{Webpage Perception QA}
This task assesses MLLMs' perception of webpage elements, including their position, style, spatial relationships, and overall page layout, through three subtasks. The \textbf{Real-world QA} subtask evaluates perceptual abilities using 625 real webpage screenshots (270 manually collected, 355 from Uground~\citep{gou2024uground} and IW-Bench~\citep{DBLP:conf/acl/GuoZCGYDCHLMZ025}), resulting in 1,250 question-answer pairs. Complementing this, \textbf{Synthetic QA} assesses model performance on 400 Q/A pairs derived from 200 synthesized webpage screenshots generated via the specific methodology (detailed in the next Webpage Code Generation).
Finally, \textbf{Multi-window QA} elevates task complexity by presenting 75 samples, each combining 2-4 screenshots from the Real-world QA set (totaling 150 Q/A pairs), challenging models to accurately identify and locate the screenshot relevant to the posed question. Questions are primarily generated by GPT-4o~\citep{4o}, augmented with bounding boxes and OCR data extracted by OmniParser~\citep{lu2024omniparserpurevisionbased}. This allows GPT-4o to focus on generating challenging, high-quality multiple-choice questions based on page content and structure rather than low-level perception. All generated questions undergo rigorous manual review and modification to ensure correctness, challenge, and task validity. To mitigate ethical risks such as privacy leakage, all webpage screenshots are manually inspected and annotated to remove personal data and harmful content.

\paragraph{Webpage Code Generation}
The Webpage Code Generation task evaluates the ability to translate visual page designs into executable HTML. Existing benchmarks (e.g., WebCode2M~\citep{DBLP:conf/www/GuiL000CSC0Z00025}, Design2Code~\citep{si2024design2code}) often simplify HTML from sources like Common Crawl~\citep{commoncrawl2025} to mitigate ethical issues, remove external dependencies and redundant elements, and standardize code for comparison. Despite these benefits, simplification is inherently time-consuming and difficult to generalize across varied real-world webpages.
Meanwhile, HTML generated with LLMs from scratch (e.g., WebSight~\citep{laurenccon2024websight}) often lacks authentic complexity.
Moreover, a key limitation of existing datasets is their handling of images, such as using generic placeholders or random images, hindering the assessment of nuanced image understanding and utilization crucial for high-fidelity webpage replication. To overcome these issues, we introduce a synthesis pipeline from real-world webpages. This two-stage process (detailed in Figure \ref{fig:overview}) starts with a real-world webpage screenshot and its OmniParser-extracted element information. GPT-4o generates an initial HTML-v1, which Claude 3.7 Sonnet then refines---adjusting styles, positions, alignments, and layouts---into a higher-quality, more complex HTML-v2. This HTML-v2 and its rendered page serve as ground truth. For image handling, we utilize a category-based strategy to preserve the visual information from real-world webpage screenshots (see Appendix \ref{appendix_Category_imgs}). As shown in Figure \ref{fig:datasets_comparison}, our method generates webpages that are demonstrably superior to other benchmarks in complexity and diversity. We also demonstrate the robustness of our data construction method against pipeline-induced biases in Appendix \ref{sec:construction_bias_analysis}. Unlike traditional tasks that only involve providing a webpage screenshot for HTML code generation, we design four distinct subtasks to evaluate MLLMs' front-end code generation capabilities under various conditions: \textbf{Image to Code} (200 samples) evaluates direct HTML generation from these HTML-v2 rendered screenshots; \textbf{Text to Code} (200 samples) assesses HTML generation based solely on manually verified textual descriptions of HTML-v2 rendered pages; \textbf{Interaction Authoring} (100 samples) measures the ability to implement dynamic behaviors, requiring MLLMs to reproduce a static page (from HTML-v1 as a base) and add specified interactions based on screenshots depicting the page before and after the interaction; and \textbf{Code Refinement} (200 samples) simulates code optimization by requiring MLLMs to refine provided HTML-v1 code to match the quality and complexity of an HTML-v2 rendered screenshot. For more detailed task descriptions, see the Appendix \ref{appendix_webpage_code_generation}.

\subsection{Evaluation Metrics}
To comprehensively evaluate MLLM performance, we employ visual- and code-level metrics, detailed below and applied specifically to each core task.

\paragraph{Visual-Level Metrics}
We evaluate generative quality by comparing model outputs, including rendered HTML pages and direct design images, with ground-truth images. We use two automatic similarity metrics: \textbf{CLIP Score}~\citep{DBLP:conf/icml/RadfordKHRGASAM21} for high-level semantic consistency and \textbf{DINOv2 Score}~\citep{DBLP:journals/tmlr/OquabDMVSKFHMEA24} for finer-grained structural and visual fidelity. In addition, \textbf{Gemini Visual Score} uses Gemini 2.5 Flash to rate outputs on a 0--10 scale across ten fine-grained criteria, including alignment and spacing (Appendix~\ref{sec:appendix_gemini_visual_score}).

\paragraph{Code-Level Metrics}
To measure code similarity, we propose the \textbf{Code Score}, which compares MLLM-generated code with reference HTML. It parses both into DOM trees, extracts CSS, and aggregates structural and content-level similarities. Structural similarity is measured by the Longest Common Subsequence (LCS) ratio of DOM tag sequences. Content-level similarity is computed for text, images, and forms by matching elements and comparing their content, key styles, and critical attributes, such as image sources and form types. Each content type is adjusted by its implementation rate, reflecting the proportion of reference elements reproduced. The final Code Score combines these components with predefined weights; details are provided in Appendix~\ref{sec:appendix_code_score}. Code Score is therefore a reference-relative fidelity measure rather than a standalone test of semantic correctness; render-level metrics, executable interaction checks, and human evaluation provide complementary evidence.

For Webpage Design, we use visual-level metrics to assess generated designs. For Webpage Perception QA, we report standard multiple-choice accuracy. For Webpage Code Generation, we use both visual-level metrics and Code Score. To validate our evaluation framework, we compare automatic metrics with human preferences using Spearman's rank correlation. As shown in Appendix~\ref{sec:appendix_metric_validity} and Appendix~\ref{sec:judge_bias_analysis}, \textbf{Code Score} ($\rho=0.9152$) and \textbf{Gemini Visual Score} ($\rho=0.9364$) show strong alignment with human judgments, supporting their reliability for assessing MLLM performance.

\begin{table}[htbp]
    \vspace{-5pt}
    \centering
    \scriptsize

    \renewcommand{\arraystretch}{1.2}

    \setlength{\tabcolsep}{0pt}

    \resizebox{0.98\linewidth}{!}{
\begin{tabular*}{\linewidth}{@{\extracolsep{\fill}} l c c c c @{}}
        \toprule
        \textbf{Model} & \textbf{\makecell{Gemini\\Visual Score}} & \textbf{\makecell{CLIP\\Score}} & \textbf{\makecell{DINOv2\\Score}} & \textbf{\makecell{Human\\Score}} \\
        \midrule
        GPT-4o & \textbf{5.0450} & \textbf{0.7445} & \textbf{0.5598} & \textbf{6.9600} \\
        \makecell[l]{gemini-2.0-flash-exp-\\image-generation} & 2.0190 & 0.6901 & 0.4798 & 6.0400 \\
        \bottomrule
    \end{tabular*}
}
\caption{Evaluation results of Webpage Design task. We mark the \textbf{better results} with bold font.}
\label{table:design_score}

    \vspace{-4pt}
\end{table}

\begin{table}[t]
\vspace{-4pt}
  \centering
  \scriptsize
  \setlength{\tabcolsep}{0pt}

  \resizebox{0.98\linewidth}{!}{
\begin{tabular*}{\linewidth}{@{\extracolsep{\fill}} l ccc @{}}
    \toprule
    \textbf{Model} & \textbf{Real-world} & \textbf{Synthetic} & \textbf{Multi-window} \\
    \midrule
    Qwen2.5-VL-72B-Instruct & 0.4696 & 0.4950 & 0.4267 \\
    InternVL2.5-78B & 0.4696 & 0.5050 & 0.4267 \\
    InternVL3-78B & 0.4816 & 0.5375 & 0.4600 \\
    LLaVA-Onevision-72B & 0.3296 & 0.3275 & 0.2733 \\
    \hdashline
    Claude 3.7 Sonnet & 0.5464 & 0.5325 & 0.4533 \\
    Gemini 2.5 Flash & 0.4800 & 0.4250 & 0.3867 \\
    GPT-4o & 0.4448 & 0.4675 & 0.3733 \\
    o4-mini & 0.4976 & 0.5300 & 0.4400 \\
    GPT-4.1 & 0.4672 & 0.4650 & 0.3733 \\
    GPT-5 & \uline{0.5816} & \uline{0.5775} & \uline{0.5467} \\
    \hdashline
    o1$^*$ & 0.4350 & 0.4600 & 0.4200 \\
    Gemini 2.5 Pro$^*$ & 0.5200 & 0.5800 & 0.4800 \\
    \hdashline
    Human Expert$^*$ & \textbf{0.9700} & \textbf{0.9600} & \textbf{0.9400} \\
    \bottomrule
  \end{tabular*}
}
\caption{Evaluation results on three Webpage Perception QA tasks. Most models are evaluated on \textbf{\ourdataset}. Models marked with $^*$ (e.g., o1, Gemini 2.5 Pro, Human Expert) are evaluated on \textbf{\ourdataset-mini}.}
\label{tab:qa_performance}
  \vspace{-14pt}
\end{table}

\section{Experiments}
\subsection{Evaluation Settings}
\textbf{\ourdataset-mini Dataset} To facilitate rapid iterative evaluation of MLLMs, we constructed a \ourdataset-mini dataset. For specifics on the FullFront-mini setup, see Appendix \ref{appendix_FullFront_mini}.

\textbf{Models} We evaluate twelve state-of-the-art MLLMs across the Webpage Perception QA and Code Generation tasks, covering both open-source and proprietary models. For the Webpage Design task, which targets image-generation MLLMs, we additionally evaluate two representative image-generation models. Detailed model sources and configurations are provided in Appendix~\ref{Evaluation Settings}.

\subsection{Main Results}

\begin{table*}[t] 
    \centering
    \setlength{\tabcolsep}{2.5pt}

    \resizebox{\textwidth}{!}{
    \begin{tabular}{l *{16}{c}}
        \toprule

        \multirow{2}{*}{\textbf{Model}}
        & \multicolumn{4}{c}{\textbf{Code Score}} & \multicolumn{4}{c}{\textbf{Gemini Visual Score}} & \multicolumn{4}{c}{\textbf{CLIP Score}} & \multicolumn{4}{c}{\textbf{DINOv2 Score}} \\
        \cmidrule(lr){2-5} \cmidrule(lr){6-9} \cmidrule(lr){10-13} \cmidrule(lr){14-17}
        & \textbf{Ref} & \textbf{Img} & \textbf{Inter} & \textbf{Text} & \textbf{Ref} & \textbf{Img} & \textbf{Inter} & \textbf{Text} & \textbf{Ref} & \textbf{Img} & \textbf{Inter} & \textbf{Text} & \textbf{Ref} & \textbf{Img} & \textbf{Inter} & \textbf{Text} \\
        \midrule

        Qwen2.5-VL-72B-Instruct & 0.53 & 0.40 & 0.40 & 0.49 & 6.21 & 4.48 & 6.22 & 5.18 & 0.79 & 0.72 & 0.76 & 0.73 & 0.63 & 0.49 & 0.67 & 0.47 \\
        InternVL2.5-78B & 0.36 & 0.33 & 0.30 & 0.50 & 5.03 & 4.01 & 3.51 & 5.44 & 0.74 & 0.74 & 0.69 & 0.74 & 0.67 & 0.57 & 0.58 & 0.48 \\
        InternVL3-78B & 0.49 & 0.42 & 0.38 & 0.50 & 5.87 & 4.47 & 4.48 & 4.91 & 0.77 & 0.73 & 0.73 & 0.72 & 0.61 & 0.53 & 0.60 & 0.44 \\
        LLaVA-Onevision-72B & 0.33 & 0.14 & 0.06 & 0.41 & 4.90 & 1.89 & 0.45 & 5.13 & 0.73 & 0.65 & 0.58 & 0.74 & 0.66 & 0.42 & 0.17 & 0.50 \\
        \hdashline 
        Claude 3.7 Sonnet & 0.63 & \textbf{0.64} & \textbf{0.55} & \textbf{0.64} & 8.36 & \textbf{8.93} & \textbf{9.18} & 8.14 & \uline{0.88} & \textbf{0.89} & \textbf{0.86} & \uline{0.87} & 0.80 & \textbf{0.81} & 0.82 & 0.78 \\
        Gemini 2.5 Flash & \textbf{0.70} & \uline{0.63} & 0.52 & \uline{0.59} & 8.65 & 8.64 & 8.07 & 8.01 & \textbf{0.89} & \uline{0.88} & 0.81 & \uline{0.87} & \textbf{0.83} & \uline{0.80} & 0.79 & \uline{0.79} \\
        GPT-4o & 0.43 & 0.34 & 0.36 & 0.49 & 6.53 & 5.91 & 5.81 & 6.14 & 0.82 & 0.81 & 0.76 & 0.74 & 0.73 & 0.69 & 0.69 & 0.52 \\
        o4-mini & 0.62 & 0.57 & \textbf{0.55} & 0.51 & 8.16 & 8.47 & 8.84 & 6.87 & 0.86 & 0.87 & 0.84 & 0.82 & 0.75 & 0.76 & 0.77 & 0.68 \\
        GPT-4.1 & \uline{0.67} & 0.61 & \textbf{0.55} & 0.55 & \textbf{9.03} & \uline{8.89} & \uline{9.13} & \uline{8.36} & \textbf{0.89} & \uline{0.88} & 0.84 & \uline{0.87} & 0.80 & \uline{0.80} & \textbf{0.84} & \textbf{0.80} \\
        GPT-5 & 0.58 & 0.56 & \uline{0.53} & \uline{0.59} & \uline{8.84} & 8.58 & 9.00 & \textbf{8.48} & \uline{0.88} & \uline{0.88} & \uline{0.85} & \textbf{0.88} & \uline{0.81} & \textbf{0.81} & \uline{0.83} & \uline{0.79}  \\
        \bottomrule
    \end{tabular}
    }
\caption{Evaluation results of different models on four Webpage Code Generation tasks. Ref: Code Refinement; Img: Image to Code; Inter: Interaction Authoring; Text: Text to Code. We mark the \textbf{best results} with bold font and the \uline{second best} with underline.}
\label{tab:code_results}

    \vspace{-12pt}
\end{table*}

\paragraph{Webpage Design}
On the Webpage Design task, current text-to-image MLLMs exhibit a foundational capability in generating general layout concepts but encounter difficulties in producing high-fidelity designs that accurately reflect detailed textual descriptions. As shown in Table \ref{table:design_score}, GPT-4o outperforms gemini-2.0-flash-exp-image-generation across all metrics. Furthermore, qualitative examples in Appendix~\ref{appendix_webpage_design} illustrate that GPT-4o demonstrates superior performance in rendering overall page structure, typography, and element implementation fidelity.

\paragraph{Webpage Perception QA}
As demonstrated in Table \ref{tab:qa_performance}, MLLMs generally exhibit weak perceptual capabilities on the Webpage Perception QA task. On the FullFront-mini subset, even the top-performing model, Gemini 2.5 Pro, achieves an average accuracy barely exceeding 50\% across the three subtasks. Furthermore, LLaVA-OneVision-72B's accuracy remains below 35\% on all QA subtasks. Human experts achieve 97\%, 96\%, and 94\% on the three FullFront-mini QA subtasks. Because the full-set model rows and mini-set human row use different evaluation subsets, we do not interpret their point-estimate differences as paired accuracy gaps. Notably, this task reveals no substantial performance disparity between closed-source and open-source models; for instance, on the full FullFront benchmark, the open-source InternVL3-78B is competitive with proprietary models, surpassing Claude 3.7 Sonnet with accuracies of 53.75\% on Synthetic QA and 46.00\% on Multi-window QA. Further analysis indicates nearly identical model performance on single-page Real-world and Synthetic QA, while performance degrades considerably on the more complex Multi-window QA.

\begin{table}[htbp]
    \centering
    \scriptsize

    \setlength{\tabcolsep}{0pt}

    \resizebox{0.98\linewidth}{!}{
\begin{tabular*}{\linewidth}{@{\extracolsep{\fill}} l c c c c @{}}
        \toprule
        \textbf{Model} & \textbf{Ref} & \textbf{Image} & \textbf{Inter} & \textbf{Text} \\
        \midrule
        Qwen2.5-VL-72B-Instruct & 6.18 & 5.72 & 7.02 & 4.90 \\
        InternVL2.5-78B & 5.36 & 4.78 & 5.04 & 4.24 \\
        InternVL3-78B & 6.32 & 5.56 & 5.44 & 4.62 \\
        LLaVA-Onevision-72B & 5.64 & 2.96 & 0.58 & 4.42 \\
        \hdashline
        Claude 3.7 Sonnet & 8.00 & 8.48 & \textbf{8.80} & 8.10 \\
        Gemini 2.5 Flash & \uline{8.44} & 8.40 & 7.86 & 8.24 \\
        GPT-4o & 6.96 & 6.74 & 6.76 & 6.24 \\
        o4-mini & 8.32 & 7.88 & 7.86 & 6.84 \\
        GPT-4.1 & \textbf{8.50} & \textbf{8.88} & 8.46 & \textbf{8.42} \\
        GPT-5 & 8.38 & \uline{8.52} & \uline{8.76} & \uline{8.34} \\
        \bottomrule
    \end{tabular*}
}
\caption{Human Evaluation of MLLM-generated webpages on FullFront-mini.}
\label{tab:human_eval}

    \vspace{-14pt}
\end{table}

\paragraph{Webpage Code Generation}
\textbf{Closed-Source Model Dominance.} In the Webpage Code Generation task, closed-source models significantly outperform their open-source counterparts across all subtasks and metrics, with no open-source model securing a top-two position in any category. As detailed in Table \ref{tab:code_results}, Claude 3.7 Sonnet leads in most tasks, closely followed by other proprietary models like GPT-5, Gemini 2.5 Flash, and GPT-4.1, all demonstrating impressive, top-tier results.
For instance, in the Code Refinement task, GPT-4.1 achieves a Gemini Visual Score of 9.03, indicating near-perfect visual reproduction in most cases, whereas the best-performing open-source model, Qwen2.5-VL-72B-Instruct, scores only 6.21 under the same settings. This performance gap is more pronounced in complex scenarios; in the Interaction Authoring task, proprietary models like o4-mini maintain high Code Scores of 0.55, while open-source models struggle significantly, with LLaVA-Onevision-72B dropping to a negligible 0.06. While Qwen2.5-VL-72B-Instruct and InternVL3-78B show relatively strong performance among open-source options, their scores are generally comparable only to GPT-4o rather than the leading closed-source models. A consistent trend across models is the alignment of performance across different metrics; models excelling in one visual or code-based score typically perform similarly well in others.

\textbf{Insights on Input Modality and Task Difficulty.} Subtask analysis reveals distinct patterns regarding input efficacy. While providing partial HTML (Code Refinement) naturally yields the highest performance, a comparative analysis of input modalities offers a surprising insight: \textbf{Text to Code performance rivals Image to Code} for top-tier models. For instance, Claude 3.7 Sonnet achieves a Code Score of 0.64 on Text to Code, matching its Image to Code performance, a balanced capability also clearly evidenced by GPT-5. Furthermore, as detailed in Appendix~\ref{appendix_text_llm}, even text-only LLMs achieve highly competitive results when provided with detailed descriptions. This indicates that \textbf{precise textual specifications can effectively substitute for visual inputs}---a critical insight for designing efficient, low-latency front-end agents where visual processing overhead might be prohibitive. Conversely, Interaction Authoring remains the most challenging task, yielding lower scores despite simpler HTML-v1 targets, a difficulty underscored by interaction implementation rates (Table \ref{table:interaction_rate}) where closed-source models exceed 70\% success, far surpassing open-source models like LLaVA-Onevision-72B (16\%).

\textbf{Human Evaluation and Reliability.}
Blind human evaluation on the FullFront-mini dataset, using Gemini Visual Score criteria (Table \ref{tab:human_eval}), further confirms that closed-source models like Claude 3.7 Sonnet and GPT-4.1 are perceived as more accurate, frequently scoring above 8/10 for reproduction quality. While these models achieve high overall fidelity, examples in Appendix \ref{appendix_webpage_code_generation} reveal that even top performers exhibit minor imperfections in fine-grained details.

\begin{figure*}[t]
  \centering
  \includegraphics[width=0.8\linewidth]{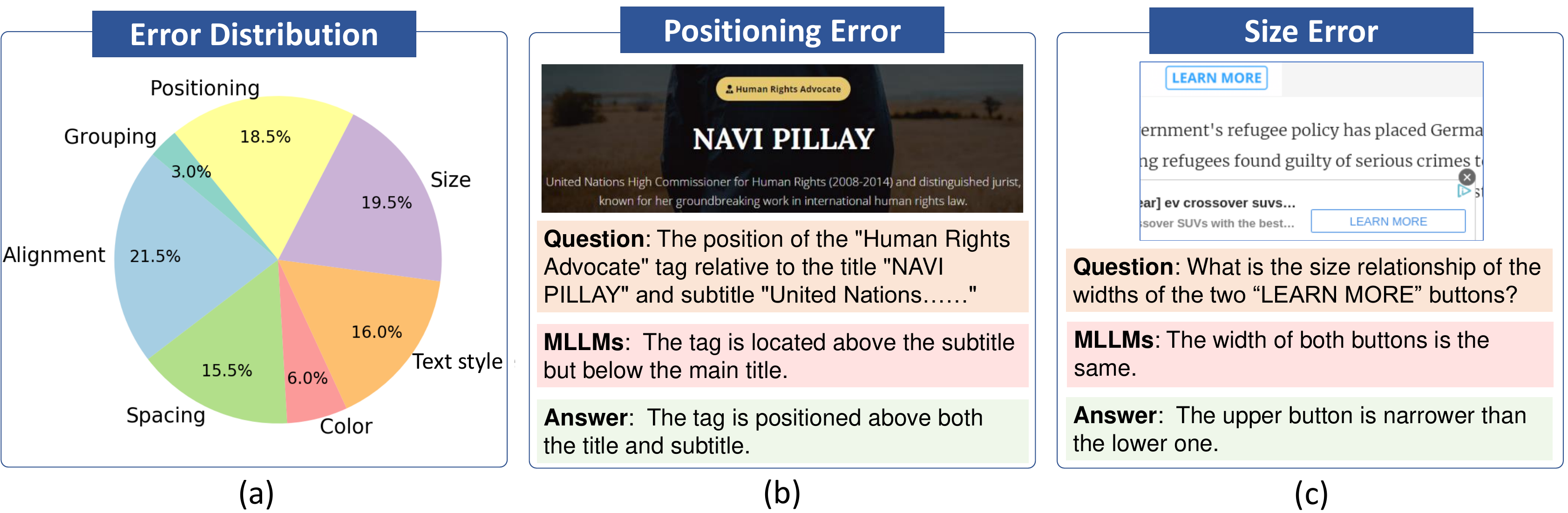}
  \caption{MLLM Errors in Webpage Perception QA. (a) Distribution of error types for 200 questions. (b) An illustrative example of a Positioning Error. (c) An illustrative example of a Size Error.}
  \label{fig:qa_errors}
    \vspace{-12pt}
\end{figure*}

\section{Discussion}

\subsection{Where Do MLLMs Struggle Most in Perception?}
By analyzing the error types of 200 questions that all MLLMs (except o1 and Gemini 2.5 Pro) fail to answer correctly, we gain insight into the primary difficulties current MLLMs face in page perception. As shown in Figure \ref{fig:qa_errors} (a), MLLMs exhibit a particular difficulty in accurately understanding the alignment (21.5\%), size (19.5\%), spacing (15.5\%), and precise positioning (18.5\%) of page elements. These factors constitute the core reasons behind perception failures. For example, Figure \ref{fig:qa_errors} (b) shows an instance where MLLMs fail to correctly identify the position of the tag labeled ``Human Rights Advocates'' relative to the main title and subtitle, while Figure \ref{fig:qa_errors} (c) demonstrates an incorrect comparison of the sizes of two buttons (More examples and analysis can be found in Appendix \ref{appendix_Webpage_Perception_QA}).

\begin{table}[t]
    \centering
    \scriptsize

    \setlength{\tabcolsep}{0pt}

    \resizebox{0.98\linewidth}{!}{
\begin{tabular*}{\linewidth}{@{\extracolsep{\fill}} l cccc @{}}
        \toprule
        \textbf{Model} & \bm{$p\checkmark c\checkmark$} & \bm{$p\checkmark c\times$} & \bm{$p\times c\checkmark$} & \bm{$p\times c\times$} \\
        \midrule
        Qwen2.5-VL-72B-Instruct & 22 & 28 & 10 & 40 \\
        InternVL3-78B & 25 & 23 & 8 & 44 \\
        GPT-4o & 32 & 13 & 21 & 34 \\
        Claude 3.7 Sonnet & 44 & 3 & 36 & 17 \\
        Gemini 2.5 Flash & 39 & 5 & 25 & 31 \\
        GPT-5 & 48 & 6 & 31 & 15 \\
        \bottomrule
    \end{tabular*}
}
\caption{Confusion matrix of Perception accuracy vs. Code Implementation accuracy on 100 samples from Synthetic QA. $p$: Perception correctness; $c$: Code implementation correctness.}
\label{tab:perception_code_correlation}
    \vspace{-12pt}
\end{table}

\subsection{Decoupling Analysis: The Relationship Between Visual Perception and Code Generation}
\label{sec:perception_code_relationship}

A counter-intuitive finding from our results (Table \ref{tab:qa_performance} and Table \ref{tab:code_results}) is that models excelling in perceptual QA tasks do not consistently outperform in code generation. For instance, InternVL3-78B demonstrates superior perceptual QA performance compared to Gemini 2.5 Flash, yet lags significantly in code generation capabilities. To investigate this causal link, we manually verified 100 samples from the Synthetic QA against their corresponding Image to Code outputs. We classified model behaviors into four quadrants based on two binary criteria: \textbf{Perception Accuracy ($p$)}, determined by whether the model correctly answered the QA question; and \textbf{Code Implementation ($c$)}, determined by whether the specific visual component mentioned in the QA was correctly implemented.

\paragraph{Divergent Failure Modes} Table \ref{tab:perception_code_correlation} presents the confusion matrix. The distribution reveals distinct failure modes across model categories. Open-source models (e.g., Qwen2.5-VL-72B-Instruct, InternVL3-78B) exhibit a high frequency of \bm{$p\checkmark c\times$} cases, indicating that even when visual perception is accurate, coding capability remains the primary bottleneck. Conversely, advanced proprietary models like Claude 3.7 Sonnet and GPT-5 show a significant number of \bm{$p\times c\checkmark$} instances. We use ``Grounding Failure'' as a behavioral label for this quadrant: the model answers the fine-grained QA incorrectly while implementing the queried relation correctly on the original page. This behavior is consistent with reliance on canonical layout priors, but the observational result alone does not uniquely distinguish such priors from memorization or distribution matching. Appendix~\ref{sec:intervention_analysis} provides controlled crop and local-edit interventions that probe these explanations directly.

\paragraph{Performance Decoupling} To further probe this decoupling effect, we analyzed the macroscopic generation scores grouped by the model's perception accuracy on the corresponding pages. As detailed in Table \ref{tab:perception_decoupling_detailed}, the generation performance metrics remain remarkably consistent regardless of whether the model correctly answered the perception questions. This statistical invariance implies that fine-grained perceptual errors do not significantly degrade the overall visual quality of the generated page. Note that Qwen2.5-VL-72B-Instruct's lower score in the ``Correct'' group is attributed to a disproportionate occurrence of fatal generation errors (e.g., ``Abnormal Image Size'') in that specific subset, reflecting generation instability rather than a negative causal link. More detailed observational examples can be found in \ref{appendix_perceptual_code} and \ref{appendix_perceptual_code_examples}; the intervention results are reported in Appendix~\ref{sec:intervention_analysis}.

\begin{table}[t]
    \centering
    \scriptsize
    \setlength{\tabcolsep}{0pt}

    \begin{tabular*}{\linewidth}{@{\extracolsep{\fill}} l l cccc @{}}
        \toprule
        \textbf{Model} & \textbf{Perc.} & \textbf{Code Score} & \textbf{G-Vis Score} & \textbf{CLIP Score} & \textbf{DINO Score} \\
        \midrule

        \multirow{2}{*}{\makecell[l]{Qwen2.5-VL\\-72B-Instruct}}
         & Cor. (46) & 0.386 & 4.430 & 0.705 & 0.409 \\
         & Wrg. (48) & 0.405 & 5.117 & 0.753 & 0.529 \\
        \midrule

        \multirow{2}{*}{InternVL3-78B}
         & Cor. (59) & 0.418 & 4.444 & 0.736 & 0.554 \\
         & Wrg. (42) & 0.409 & 4.422 & 0.715 & 0.486 \\
        \midrule

        \multirow{2}{*}{GPT-4o}
         & Cor. (43) & 0.336 & 6.007 & 0.799 & 0.672 \\
         & Wrg. (56) & 0.342 & 5.896 & 0.809 & 0.697 \\
        \midrule

        \multirow{2}{*}{\makecell[l]{Claude 3.7\\Sonnet}}
         & Cor. (57) & 0.646 & 8.826 & 0.887 & 0.805 \\
         & Wrg. (44) & 0.649 & 8.957 & 0.892 & 0.790 \\
        \midrule

        \multirow{2}{*}{\makecell[l]{Gemini 2.5\\Flash}}
         & Cor. (42) & 0.634 & 8.790 & 0.890 & 0.838 \\
         & Wrg. (72) & 0.628 & 8.596 & 0.877 & 0.796 \\
        \midrule

        \multirow{2}{*}{GPT-5}
         & Cor. (75) & 0.568 & 8.436 & 0.882 & 0.808 \\
         & Wrg. (44) & 0.535 & 8.548 & 0.878 & 0.832 \\
        \bottomrule
    \end{tabular*}
\caption{Detailed breakdown by perception accuracy. \textbf{Perc.}: Perception Group (\textbf{Cor.}: Correct; \textbf{Wrg.}: Wrong; value in parentheses denotes $N$); \textbf{G-Vis Score}: Gemini Visual Score.}
\label{tab:perception_decoupling_detailed}

    \vspace{-2pt}
\end{table}

\begin{figure*}[t]
  \centering
  \includegraphics[width=0.8\linewidth]{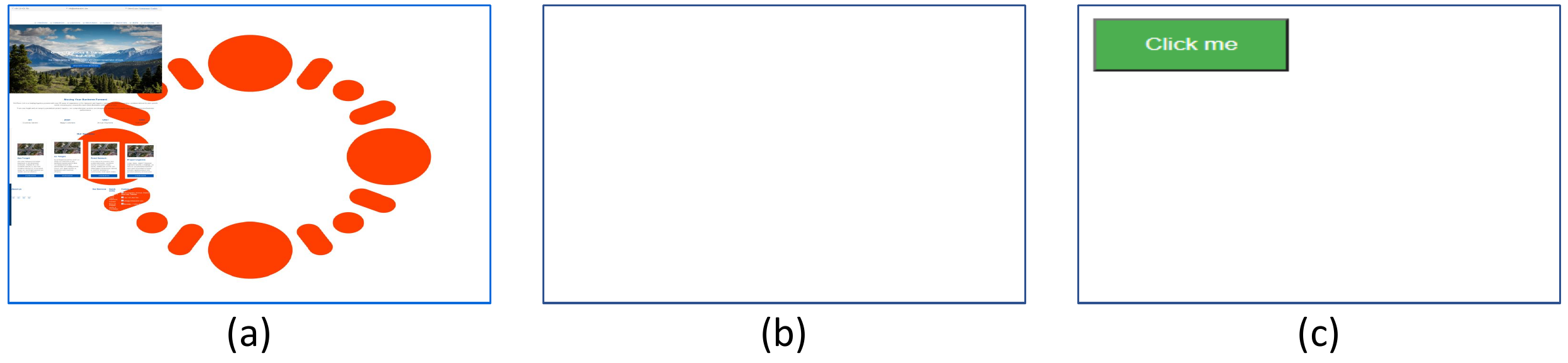}
  \caption{Three common errors in Webpage Code Generation. (a) Abnormal Image Size, where an image within the rendered page is disproportionately large. (b) Blank Page, showing an entirely blank rendered output. (c) Isolation Error, an output consisting only of an isolated interactive element.}
  \vspace{-14pt}
  \label{fig:code_errors}
\end{figure*}


\begin{table}[t]
    \centering
\resizebox{0.98\linewidth}{!}{
        \setlength{\tabcolsep}{3pt} 
        \begin{tabular}{@{}l cccc cccc c@{}}
            \toprule
            \multirow{2}{*}{\textbf{Model}} & \multicolumn{4}{c}{\textbf{Size}} & \multicolumn{4}{c}{\textbf{Blank}} & \multicolumn{1}{c}{\textbf{Isolation}} \\
            \cmidrule(lr){2-5} \cmidrule(lr){6-9} \cmidrule(lr){10-10}
            & Ref & Img & Inter & Text & Ref & Img & Inter & Text & Inter \\
            \midrule
            Qwen2.5-VL-72B-Instruct & \textbf{38} & \textbf{62} & 11 & \uline{84} & \textbf{8} & 4 & 2 & 3 & 2 \\
            InternVL2.5-78B & 5 & 20 & 2 & 76 & 2 & \uline{14} & \textbf{12} & 4 & \uline{11} \\
            InternVL3-78B & 6 & 20 & \uline{5} & \textbf{95} & \uline{5} & \uline{14} & \uline{10} & 1 & 1 \\
            LLaVA-Onevision-72B & 1 & \uline{22} & 3 & 64 & \uline{5} & \textbf{45} & 1 & \textbf{9} & \textbf{88} \\
            \hdashline
            Claude 3.7 Sonnet & 2 & 1 & 0 & 3 & 3 & 0 & 0 & 0 & 0 \\
            Gemini 2.5 Flash & 3 & 3 & 2 & 2 & 2 & 0 & 0 & 1 & 0 \\
            GPT-4o & 2 & 11 & 0 & 54 & 1 & 0 & 0 & 1 & 0 \\
            o4-mini & \uline{16} & 9 & 2 & 17 & 0 & 0 & 0 & \uline{6} & 0 \\
            GPT-4.1 & 4 & 1 & 0 & 2 & 2 & 1 & 2 & 1 & 0 \\
            GPT-5 & 1 & 1 & 0 & 1 & 0 & 1 & 0 & 0 & 0 \\
            \bottomrule
        \end{tabular}
    }
\caption{Counts of three error types in Webpage Code Generation tasks. Size: Abnormal Image Size; Blank: Blank Page; Isolation: Isolation Error.}
\label{tab:code_errors}

    \vspace{-5pt}
\end{table}

\begin{table}[t]
    \centering
    \scriptsize

    \begin{tabular*}{\linewidth}{@{\extracolsep{\fill}}lcccc@{}}
        \toprule
        \textbf{Model} & \textbf{Structure} & \textbf{Text} & \textbf{Image} & \textbf{Form} \\
        \midrule
        Qwen2.5-VL-72B-Instruct & 0.51 & 0.19 & 0.54 & 0.41 \\
        InternVL2.5-78B & 0.43 & 0.11 & 0.52 & 0.34 \\
        InternVL3-78B & 0.49 & 0.15 & 0.61 & 0.42 \\
        LLaVA-Onevision-72B & 0.29 & 0.06 & 0.38 & 0.23 \\
        \hdashline
        Claude 3.7 Sonnet & \textbf{0.72} & \uline{0.39} & 0.65 & 0.53 \\
        Gemini 2.5 Flash & \uline{0.70} & \textbf{0.40} & \textbf{0.71} & 0.51 \\
        GPT-4o & 0.42 & 0.11 & 0.53 & 0.34 \\
        o4-mini & 0.62 & 0.28 & 0.61 & \uline{0.54} \\
        GPT-4.1 & 0.63 & 0.27 & 0.64 & \textbf{0.64} \\
        GPT-5 & 0.63 & 0.24 & \uline{0.70} & 0.53 \\
        \bottomrule
    \end{tabular*}
\caption{Average code-level performance metrics across four Webpage Code Generation subtasks.}
\label{tab:code_metrics}
    \vspace{-14pt}
\end{table}

\subsection{Can MLLMs Be an Excellent Front-End Engineer?}
To determine if MLLM-generated pages are preferred by humans to real-world versions, five human experts conduct a blind evaluation on the same 100 source pages for each of eight MLLMs, comparing every generated page with its real-world counterpart. As detailed in Appendix~\ref{appendix_human_evaluation_info_aes} (Figure~\ref{fig:win_lose}), the results demonstrate that pages generated by leading models (e.g., o4-mini and Claude 3.7 Sonnet) are frequently preferred over the original real-world pages on the sampled aesthetic and information-effectiveness criteria. However, further analysis of the generated webpages reveals that MLLMs can exhibit three prevalent error categories, illustrated in Figure \ref{fig:code_errors}: Abnormal Image Size (abnormally large images disrupting layout integrity), Blank Page (entirely blank screenshots despite non-empty code), and Isolation Error (instances where only isolated interactive buttons are generated, neglecting page content). Each error type significantly degrades the effectiveness of the generated webpage. Table \ref{tab:code_errors} shows that open-source models exhibit these errors markedly more often than closed-source counterparts; this considerably diminishes their reliability and stability. Furthermore, a detailed examination of code-level performance (Table \ref{tab:code_metrics}) indicates that current MLLMs still have substantial room for improvement in text and form implementation, as similarity scores for these components do not exceed 0.65.

Overall, despite certain shortcomings in fine-grained details, MLLMs do demonstrate the capability to design generally coherent webpage interfaces from textual descriptions and can generate corresponding code from webpage screenshots. However, the overall deficiencies in their perceptual abilities, coupled with the potential for critical errors during code generation, render their current reliability and stability uncertain. We believe a promising future direction involves integrating MLLMs with specialized tools. This approach could compensate for their perceptual limitations and provide mechanisms to identify and rectify generation anomalies, thereby aiding MLLMs in their evolution towards becoming excellent front-end engineers.

\section{Conclusion}

We introduce \textbf{FullFront}, a pioneering and comprehensive Multimodal Front-end Benchmark. FullFront is designed to systematically evaluate the capabilities of Multimodal Large Language Models (MLLMs) across the entire front-end development pipeline, encompassing key aspects such as design, page perception, and code generation. By constructing high-quality, diverse synthetic data and designing a multi-layered evaluation system, FullFront serves as a powerful tool for in-depth analysis of the strengths and limitations of current MLLMs. While no single current model has yet mastered the entire workflow, FullFront provides the necessary infrastructure to evaluate future ``Generalist Front-End Agents.'' It establishes a standard for tracking the evolution of MLLMs from isolated task solvers to comprehensive engineering assistants, laying the foundation for the development of the next generation of intelligent webpage development tools.



\section*{Limitations}
FullFront provides coordinated coverage of three front-end authoring stages---conceptualization, fine-grained perception, and implementation---but evaluates them separately. It does not pass artifacts from one stage to the next or measure end-to-end error propagation. The current code tasks use standalone HTML/CSS/JavaScript in a desktop, single-page browser environment. They therefore do not directly evaluate React, Vue, Tailwind, componentization, responsive/mobile design, cross-browser compatibility, multi-page navigation, deployment, maintainability, security, asynchronous APIs, authentication, persistence, or complex state management. Interaction Authoring is limited to ten controlled click/hover micro-interactions, and Perception QA measures fixed-choice visual discrimination rather than free-form grounding, debugging, or interactive webpage use.

The HTML-v2 references are standardized executable targets reconstructed by proprietary models rather than human-authored production code. The Grok-4 reconstruction study tests stability under an alternative proprietary generator but does not provide a non-model structural anchor, so reference-distribution matching cannot be fully excluded. Code Score measures reference-relative structural and content fidelity and retains a non-zero penalty for valid render-equivalent implementations. We therefore interpret it together with render-level metrics, executable checks, and human evaluation. Although real-world screenshots are manually screened for sensitive or harmful content, the benchmark may not represent all website genres, accessibility requirements, devices, or production constraints.

\section*{Ethical Considerations}
We follow ACL/ARR ethics expectations and EMNLP 2026 paper-integrity guidance by reporting the data construction process, evaluation protocol, and model sources transparently. Real-world webpage screenshots are manually screened to remove private or harmful content, and the reconstruction pipeline is designed to avoid reproducing copyrighted source code. The benchmark is intended for evaluation and research; deployment of generated front-end code should still involve human review for privacy, accessibility, security, and reliability.

\bibliography{custom}

\appendix

\section{Appendix Overview}
This Appendix is organized as follows:

\begin{itemize}[leftmargin=15pt]

    \item \textbf{Section \textcolor{red}{\ref{appendix_FullFront_mini}}} provides a detailed description of the FullFront-mini dataset, a condensed subset of the main benchmark designed for rapid evaluation.

    \item \textbf{Section \textcolor{red}{\ref{appendix_webpage_design}}} offers additional qualitative examples for the Webpage Design task, providing a clearer view of the capabilities and limitations of current text-to-image models.

    \item \textbf{Section \textcolor{red}{\ref{appendix_Webpage_Perception_QA}}} presents illustrative case studies for each subtask within the Webpage Perception QA task, highlighting common model failures and their underlying cause

    \item \textbf{Section \textcolor{red}{\ref{appendix_webpage_code_generation}}} elaborates on the four Webpage Code Generation subtasks, detailing the data curation process and presenting qualitative results that compare model outputs against ground-truth designs.

    \item \textbf{Section \textcolor{red}{\ref{appendix_evaluation}}} outlines our comprehensive evaluation framework, including the formulation and validation of the Gemini Visual Score and Code Score, MLLM experimental settings, and the human evaluation protocol.

    \item \textbf{Section \textcolor{red}{\ref{sec:construction_bias_analysis}}} validates the robustness of our benchmark against data-pipeline artifacts by introducing an independent dataset constructed via Grok-4, confirming that model rankings and metric reliability remain consistent across different data sources.

    \item \textbf{Section \textcolor{red}{\ref{sec:judge_bias_analysis}}} examines the impartiality of the Gemini 2.5 Flash evaluator, presenting a cross-judge validation that rules out ``same-family bias'' and demonstrates high alignment with human expert rankings.

    \item \textbf{Section \textcolor{red}{\ref{sec:code_score_sensitivity}}} analyzes the sensitivity of the Code Score metric, verifying its ability to tolerate valid syntactic variations (e.g., alternative layout techniques) while effectively penalizing semantic errors.

    \item \textbf{Section \textcolor{red}{\ref{sec:intervention_analysis}}} reports crop-assisted and counterfactual local-edit interventions for the Perception--Implementation Gap, separating the observed behavior from its possible mechanisms.

    \item \textbf{Section \textcolor{red}{\ref{sec:frontier_model_analysis}}} evaluates four newer frontier models on all Perception QA, Image-to-Code, and Text-to-Code items and quantifies the remaining benchmark headroom.

    \item \textbf{Section \textcolor{red}{\ref{sec:release_scope}}} documents the planned release contents, licenses, resource locations, and task-coverage accounting.

    \item \textbf{Section \textcolor{red}{\ref{appendix_prompts}}} contains the complete, verbatim prompts used for both data curation and model evaluation across all tasks in the FullFront benchmark.
\end{itemize}

\section{FullFront-mini}
\label{appendix_FullFront_mini}
\textbf{FullFront-mini Dataset} To facilitate rapid iterative evaluation of MLLMs and initial exploration of the benchmark by researchers, we constructed a FullFront-mini dataset. This subset is a condensed version of the full FullFront dataset, with the following specific composition.
Webpage Perception QA: Includes 200 Real-world QA, 100 Synthetic QA, and 50 Multi-window QA data samples. Webpage Code Generation: Comprises 50 Image to Code, 50 Text to Code, 50 Interaction Authoring, and 50 Code Refinement data samples. Webpage Design: Consists of 50 Webpage Design task data samples.

\section{Webpage Design}
\label{appendix_webpage_design}
\begin{figure*}[t]
  \centering
  \includegraphics[width=0.98\textwidth]{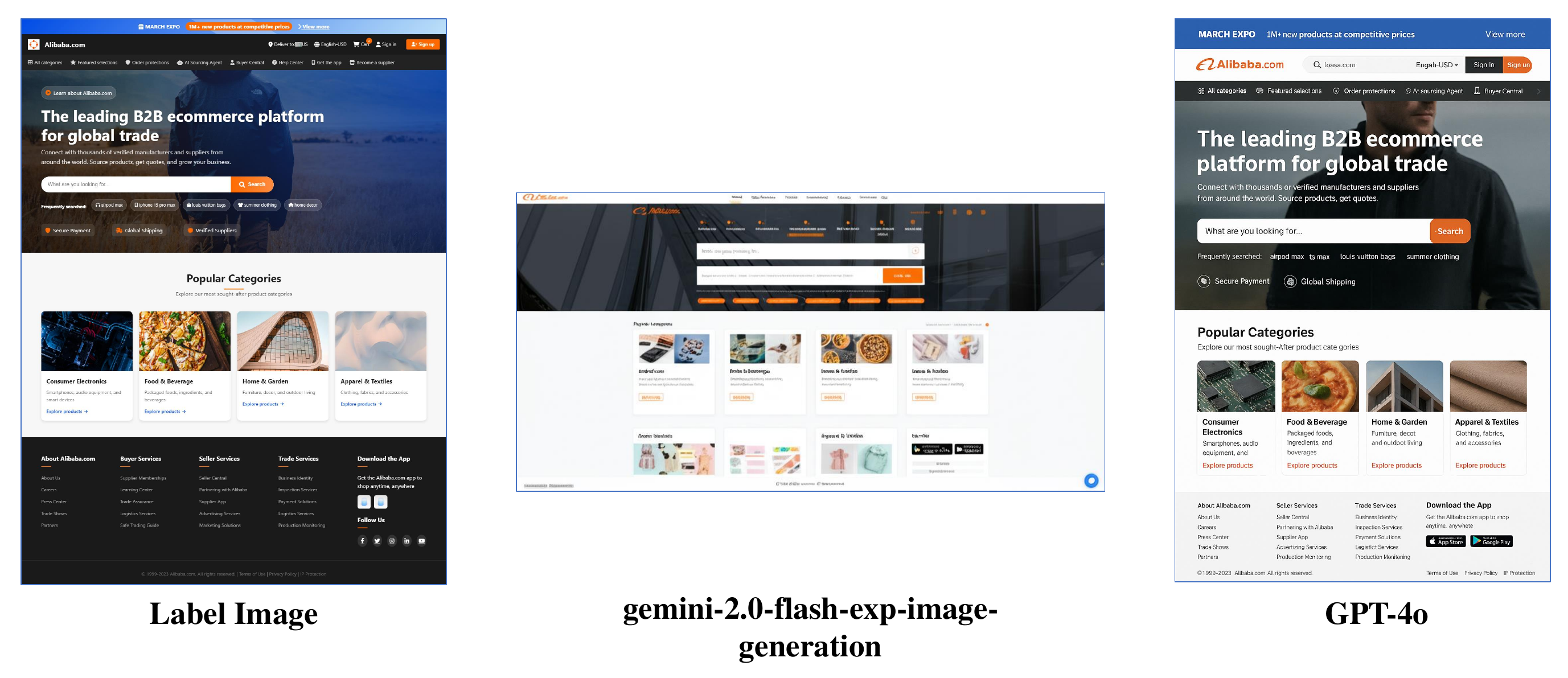}
  \caption{Comparative Webpage Designs: Ground Truth (``Label Image'') vs. gemini-2.0-flash-exp-image-generation and GPT-4o.}
  \label{fig:webpage_design}
\end{figure*}

The Webpage Design task evaluates an MLLM's ability to generate a visual webpage design based on a textual description, assessing its capacity for conceptualization within the front-end workflow.
Figure \ref{fig:webpage_design} illustrates the outputs from two evaluated text-to-image MLLMs, GPT-4o and gemini-2.0-flash-exp-image-generation, alongside the target ``Label Image'' (ground truth) for a representative example.

As observed in Figure \ref{fig:webpage_design}, the design generated by GPT-4o (right) demonstrates a notably closer resemblance to the ``Label Image'' (left) compared to the output from gemini-2.0-flash-exp-image-generation (middle).
Specifically:
\begin{itemize}
    \item \textbf{Layout and Structure:} GPT-4o more successfully replicates the overall page structure, including the header, hero section, ``Popular Categories'' grid, and footer arrangement. The placement and relative sizing of these major components are more aligned with the ground truth. In contrast, the gemini-2.0-flash-exp-image-generation produces a layout that, while containing some similar thematic elements (like a search bar and category-like items), deviates more significantly in its structural organization and visual hierarchy.
    \item \textbf{Element Completeness and Typography:} GPT-4o tends to generate a design with a higher degree of element completeness. For example, the navigation links in the header, the search bar within the hero section, and the individual category cards appear more fully formed and are stylistically closer to the target. The typography choices in GPT-4o's output also generally exhibit greater fidelity.
    \item \textbf{Detail Discrepancies:} Despite its superior overall performance, the GPT-4o design still exhibits discrepancies in fine-grained details. For instance, the footer section in the GPT-4o output uses a light background, contrasting with the dark background of the ``Label Image'' footer.
\end{itemize}
\begin{figure*}[t]
  \centering
  \includegraphics[width=0.98\textwidth]{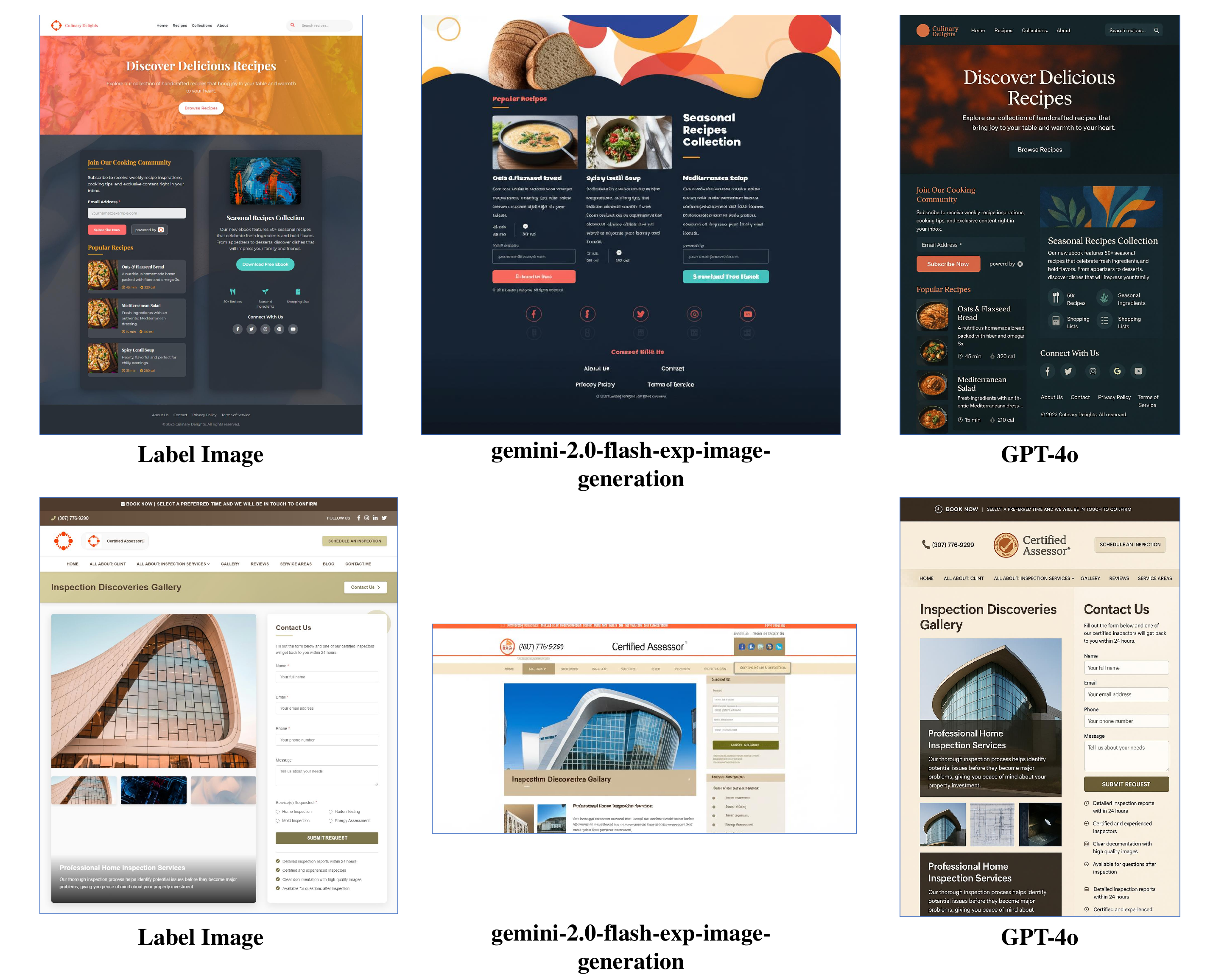}
  \caption{Results of GPT-4o and gemini-2.0-flash-exp-image-generation on the Webpage Design task.}
  \label{fig:webpage_design1}
\end{figure*}

Additional qualitative results are provided in Figure~\ref{fig:webpage_design1}. To summarize, the qualitative  examples indicate that while text-to-image MLLMs such as GPT-4o are capable of producing webpage designs that align with the primary layout and components of a textual prompt, they still lack the fine-grained control necessary to precisely render all visual attributes. Details like exact color codes, specific text, and subtle styling elements present a significant opportunity for future research. The models excel at converting abstract concepts into concrete webpage structures, but their capacity to follow detailed, nuanced instructions needs to be enhanced.

\section{Webpage Perception QA}
\label{appendix_Webpage_Perception_QA}
In this section, we provide a detailed breakdown of the three subtasks that constitute the Webpage Perception QA benchmark: Real-world QA, Synthetic QA, and Multi-window QA. For each subtask, we present an illustrative case study that highlights common model failures. Through an analysis of these specific examples, we aim to shed light on the root causes of these errors and pinpoint the underlying perceptual limitations of current MLLMs.

\subsection{Real-world QA}

The Real-world QA subtask is designed to rigorously evaluate an MLLM's ability to comprehend visual information from authentic webpage screenshots. This subtask consists of 1,250 question-answer pairs derived from 625 real-world webpages. These screenshots are sourced from a combination of manual collection and established academic benchmarks, including Uground~\citep{gou2024uground} and IW-Bench~\citep{DBLP:conf/acl/GuoZCGYDCHLMZ025}. The primary challenge of this subtask lies in the complexity and subtlety of the questions, which demand a high degree of perceptual acuity to correctly identify fine-grained details regarding element positioning, styling, and spatial relationships.

\begin{figure*}[t]
  \centering
  \includegraphics[width=0.98\textwidth]{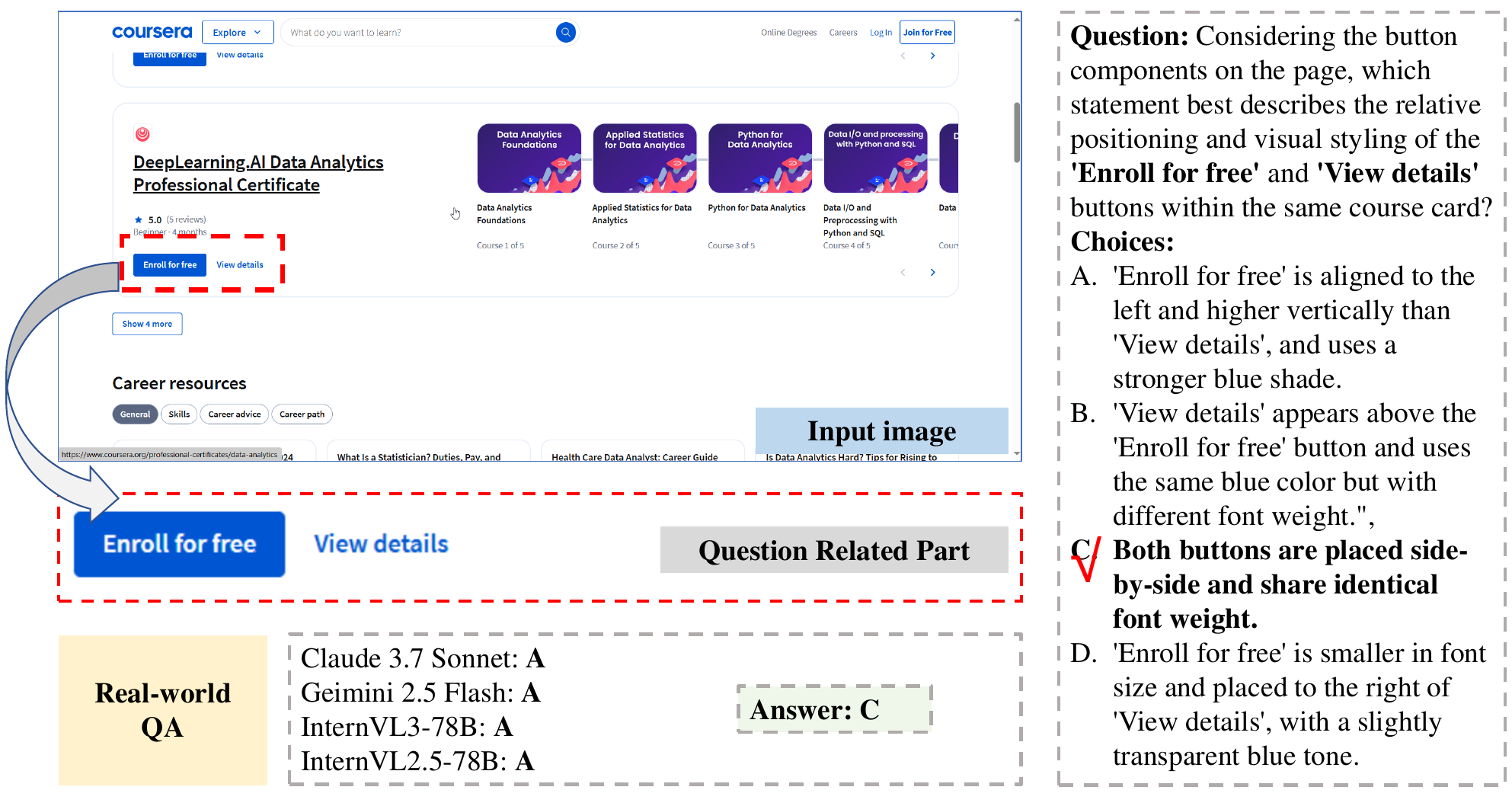}
  \caption{A sample case in Real-world QA.}
  \label{fig:Real-world QA}
\end{figure*}

Figure~\ref{fig:Real-world QA} presents a representative example, where MLLMs are tasked with assessing the positional and stylistic relationship between the ``Enroll for free'' and ``View details'' buttons. In this instance, all evaluated MLLMs converged on an incorrect answer (A). Their choice indicates that they successfully perceived coarse-grained information: that the ``Enroll for free'' button is located to the left of ``View details'' and features a solid blue background. However, their analysis is flawed, as they incorrectly inferred that it was positioned vertically higher than the other button.

Crucially, the models failed to identify the subtle but correct detail required for answer C: that both buttons share an identical font weight. This case study starkly highlights a critical limitation in current MLLMs. While they can often process general layout and prominent visual cues, they struggle with the precise perception of element alignment and nuanced typographic styles, which are fundamental for high-fidelity front-end development.

\subsection{Synthetic QA}
To complement the Real-world QA, the Synthetic QA subtask assesses model performance in a more controlled setting. This subtask consists of 400 question-answer pairs generated from 200 unique webpage screenshots. These screenshots are not sourced from the wild; instead, they are products of the synthesis pipeline detailed in the main paper. This methodology allows for the creation of webpages that, while visually diverse and realistic, can feature subtle and deliberate variations from standard design conventions. The key challenge, therefore, is to test whether MLLMs are truly perceiving the provided visual evidence or are relying on strong priors about common webpage layouts.

\begin{figure*}[t]
  \centering
  \includegraphics[width=0.98\textwidth]{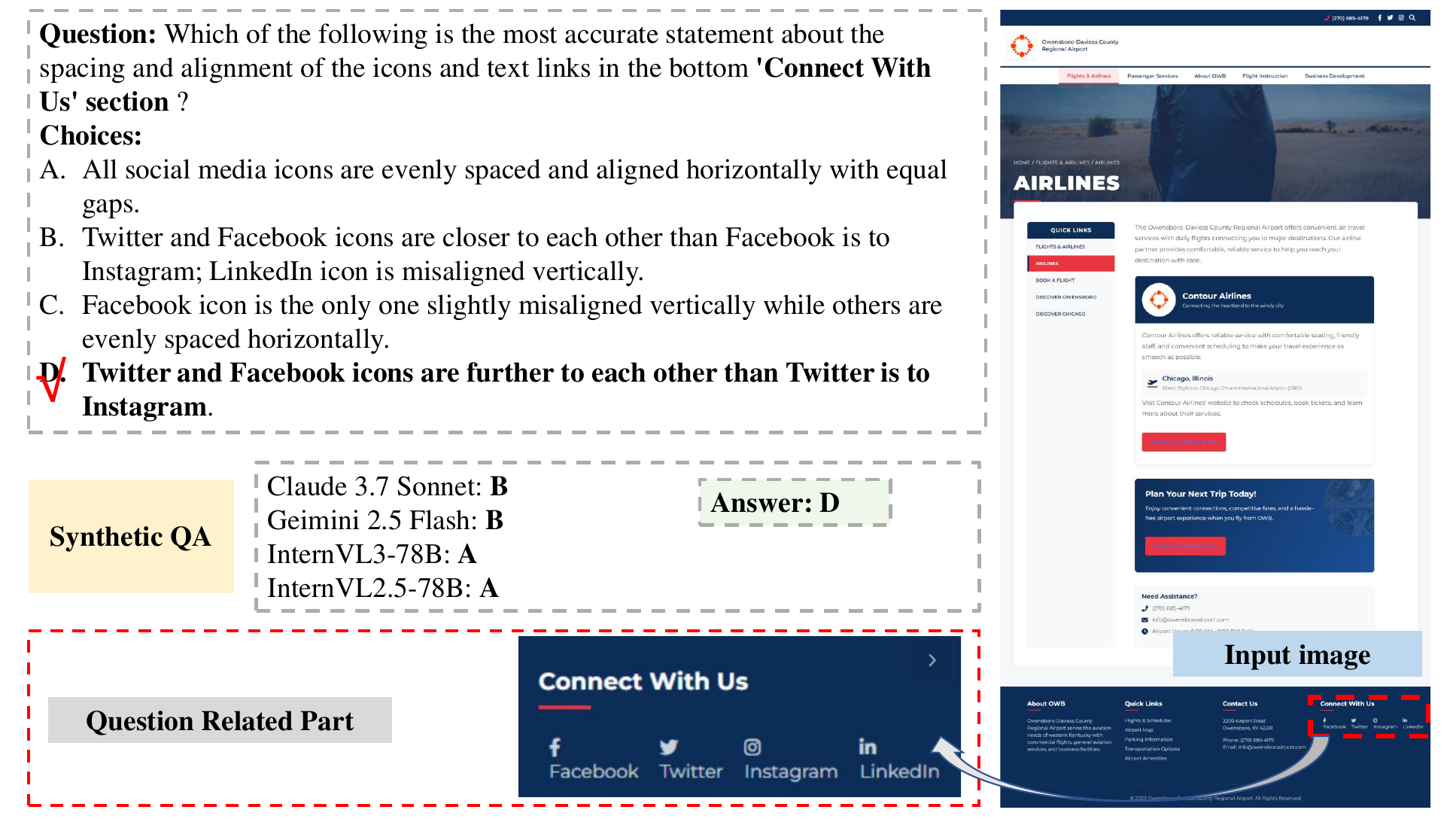}
  \caption{A sample case in Synthetic QA.}
  \label{fig:Synthetic QA}
\end{figure*}

The example in Figure~\ref{fig:Synthetic QA} effectively demonstrates this challenge, requiring models to compare the relative spacing and alignment of four social media icons. The analysis reveals two distinct modes of failure among the models:
\begin{itemize}
    \item \textbf{InternVL2.5-78B and InternVL3-78B} incorrectly concluded that the icons are evenly spaced (Option A). This error likely stems from a powerful prior, as equidistant spacing is a highly common design pattern. The models appear to have defaulted to this assumption---a form of model hallucination---instead of precisely analyzing the visual data.
    \item In contrast, \textbf{Claude 3.7 Sonnet and Gemini 2.5 Flash} exhibited a more nuanced partial understanding, correctly identifying the varied spacing among the first three icons. However, they also failed by incorrectly perceiving a vertical misalignment of the LinkedIn icon (Option B), leading them away from the correct answer, D.
\end{itemize}

This case study illustrates a critical pattern: even when MLLMs can successfully localize the specific region relevant to a question, their ability to perform a meticulous analysis of fine-grained spatial details within that region remains unreliable.

\subsection{Multi-window QA}
The Multi-window QA subtask significantly elevates the complexity to better simulate real-world scenarios where relevant information must be found across different contexts. This subtask is composed of 150 question-answer pairs distributed across 75 samples, where each sample presents a model with 2 to 4 distinct screenshots drawn from our Real-world QA set. The difficulty is twofold and distinguishes it from the previous subtasks: a model must first correctly localize the specific webpage screenshot relevant to the question, and only then perform the fine-grained perceptual analysis required to answer it. This tests both search and comprehension capabilities.

\begin{figure*}[t]
  \centering
  \includegraphics[width=0.98\textwidth]{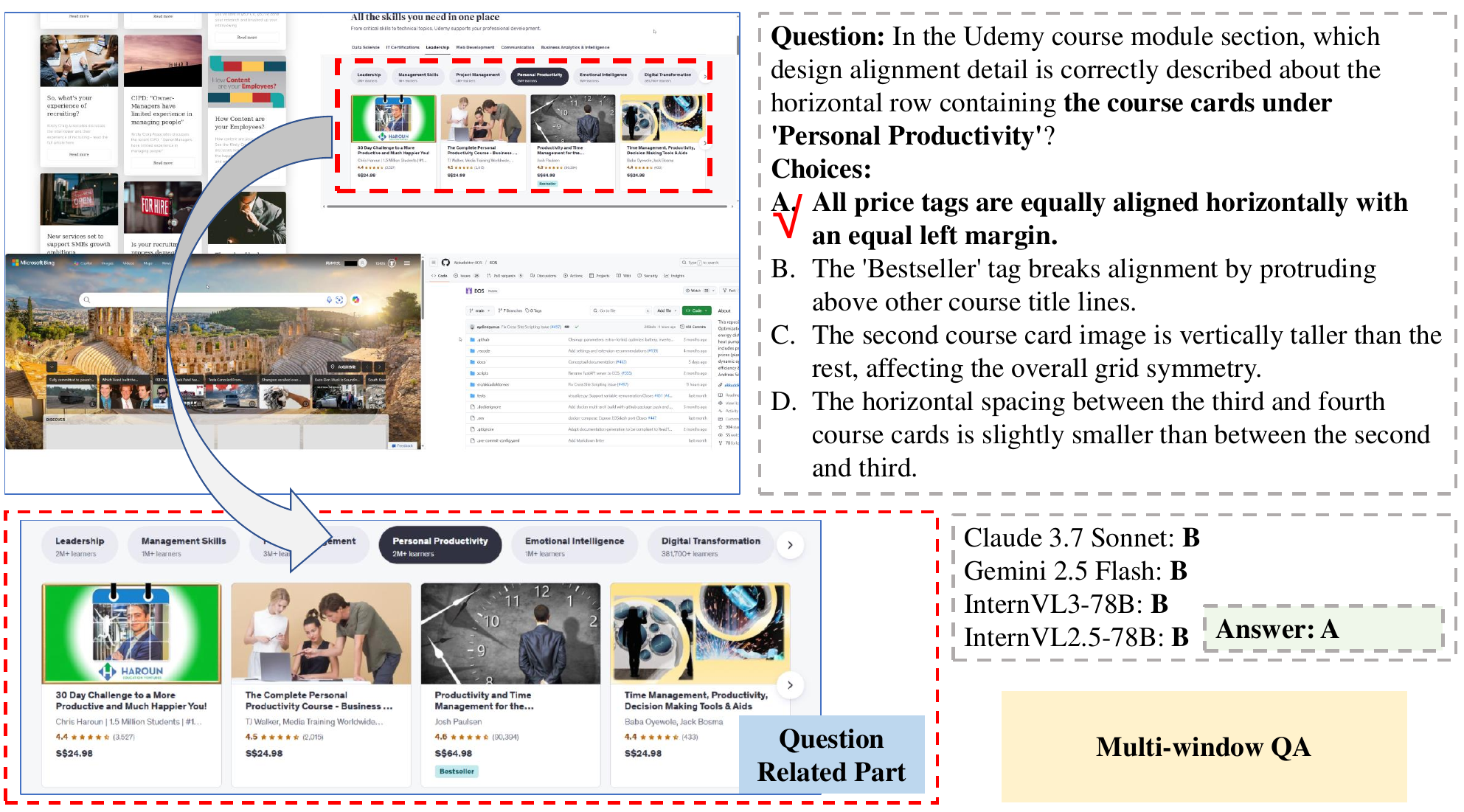}
  \caption{A sample case in Multi-window QA.}
  \label{fig:Multi-window QA}
\end{figure*}

The case presented in Figure~\ref{fig:Multi-window QA} highlights the compounded difficulty of this task. To answer a question about the layout of ``Personal Productivity'' course cards, the MLLMs must first identify the correct screenshot among the options. The results show a uniform failure, with all models incorrectly selecting option B. This choice reveals a significant perceptual error: the models incorrectly concluded that the ``Bestseller'' tag is positioned above the title lines, when in fact, the tag is located at the very bottom of the course card.

Simultaneously, this fixation on a misunderstood element caused them to overlook the correct statement in option A, which accurately describes that all price tags are equally aligned horizontally. This dual failure is revealing. It indicates that MLLMs not only face challenges in disambiguating and focusing on the correct visual context but also remain highly susceptible to misinterpreting spatial relationships even after localizing the right section. Their inability to perform systematic, holistic visual checks leads to errors in judgment.

\section{Webpage Code Generation}
\label{appendix_webpage_code_generation}

\subsection{OmniParser}
\begin{figure*}[t]
  \centering
  \includegraphics[width=0.98\textwidth]{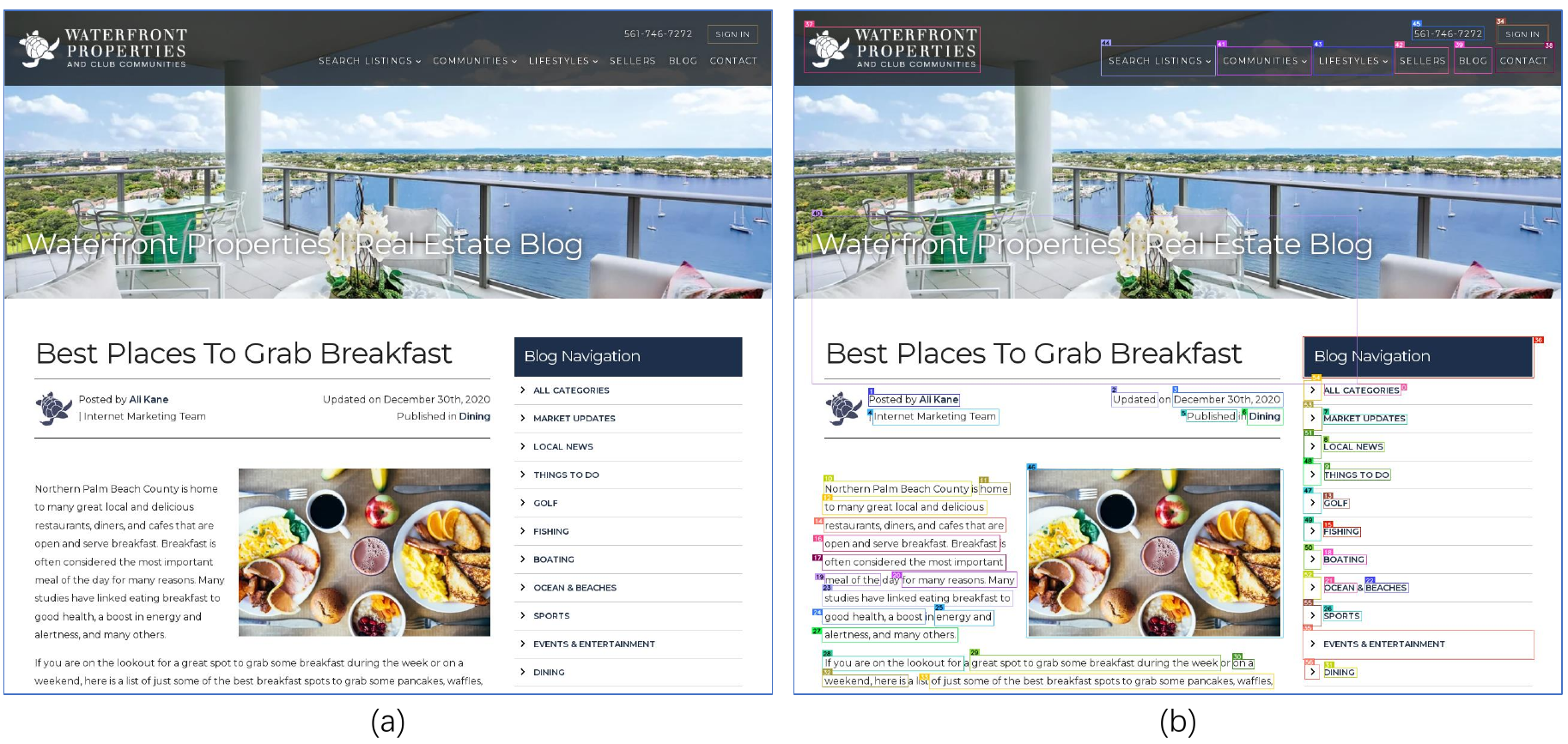}
  \caption{OmniParser extracting UI elements from a webpage screenshot. (a) Original webpage screenshot. (b) Webpage screenshot with UI elements detected and bounded by OmniParser.}
  \label{fig:omniparser}
\end{figure*}

OmniParser, a screen parsing tool developed by Microsoft, is engineered to deconstruct User Interface (UI) screenshots into structured and readily comprehensible elemental data. This capability significantly enhances the proficiency of visual models, such as GPT-4o, in generating accurate operational instructions within a given interface. As illustrated in Figure~\ref{fig:omniparser}, OmniParser adeptly detects a wide array of fine-grained UI elements from screenshots, including diminutive icons, buttons, and text boxes.

The output from OmniParser typically includes the bounding box coordinates and the recognized textual content for each detected element on the original webpage. An example of such extracted information is provided below:

\begin{tcolorbox}[
  breakable, 
  colback=gray!5, 
  colframe=black, 
  boxrule=0.8pt, 
  arc=2mm, 
]

{"bbox": [0.69, 0.546, 0.79, 0.562], "content": "ALL CATEGORIES"} \\
{"bbox": [0.096, 0.561, 0.215, 0.578], "content": "Posted by Ali Kane"} \\
{"bbox": [0.412, 0.558, 0.473, 0.579], "content": "Updated"} \\
{"bbox": [0.492, 0.558, 0.636, 0.579], "content": "December 30th 2020 2020"} \\
{"bbox": [0.102, 0.582, 0.266, 0.605], "content": "Internet Marketing Team"} \\
{"bbox": [0.51, 0.583, 0.576, 0.601], "content": "Published"} \\
{"bbox": [0.589, 0.582, 0.636, 0.605], "content": "Dining"} \\
{"bbox": [0.688, 0.59, 0.798, 0.604], "content": "MARKET UPDATES"} \\
{"bbox": [0.688, 0.631, 0.768, 0.645], "content": "LOCAL NEWS"} \\
{"bbox": [0.69, 0.671, 0.776, 0.687], "content": "THINGS To Do"}
\end{tcolorbox}

Within the FullFront benchmark's data curation pipeline, OmniParser serves as an auxiliary tool to facilitate the generation of high-quality training and evaluation instances. Specifically, it extracts elemental information (bounding boxes and content) from webpage screenshots. This extracted data is then leveraged by GPT-4o, which is involved in the dataset creation process, for data generation.

More precisely, for the Webpage Perception QA Task Data Generation, GPT-4o is provided with the original webpage screenshot, the same screenshot overlaid with OmniParser's detected bounding boxes, and the structured data from OmniParser (which includes the coordinates of these boxes and their corresponding textual content). This comprehensive input is designed to alleviate the perceptual load on GPT-4o, allowing it to concentrate on formulating challenging Question-Answer (QA) pairs.

Similarly, for the Webpage Code Generation Task Data Generation, this same set of information (original screenshot, screenshot with bounding boxes, and OmniParser's structured box/text data) is supplied to GPT-4o. This assists GPT-4o in generating an initial HTML representation (HTML-v1) as a foundational step in reconstructing the webpage.

\subsection{Category-based Utilization Strategy for Images}
\label{appendix_Category_imgs}
\begin{figure*}[t]
  \centering
  \includegraphics[width=0.98\textwidth]{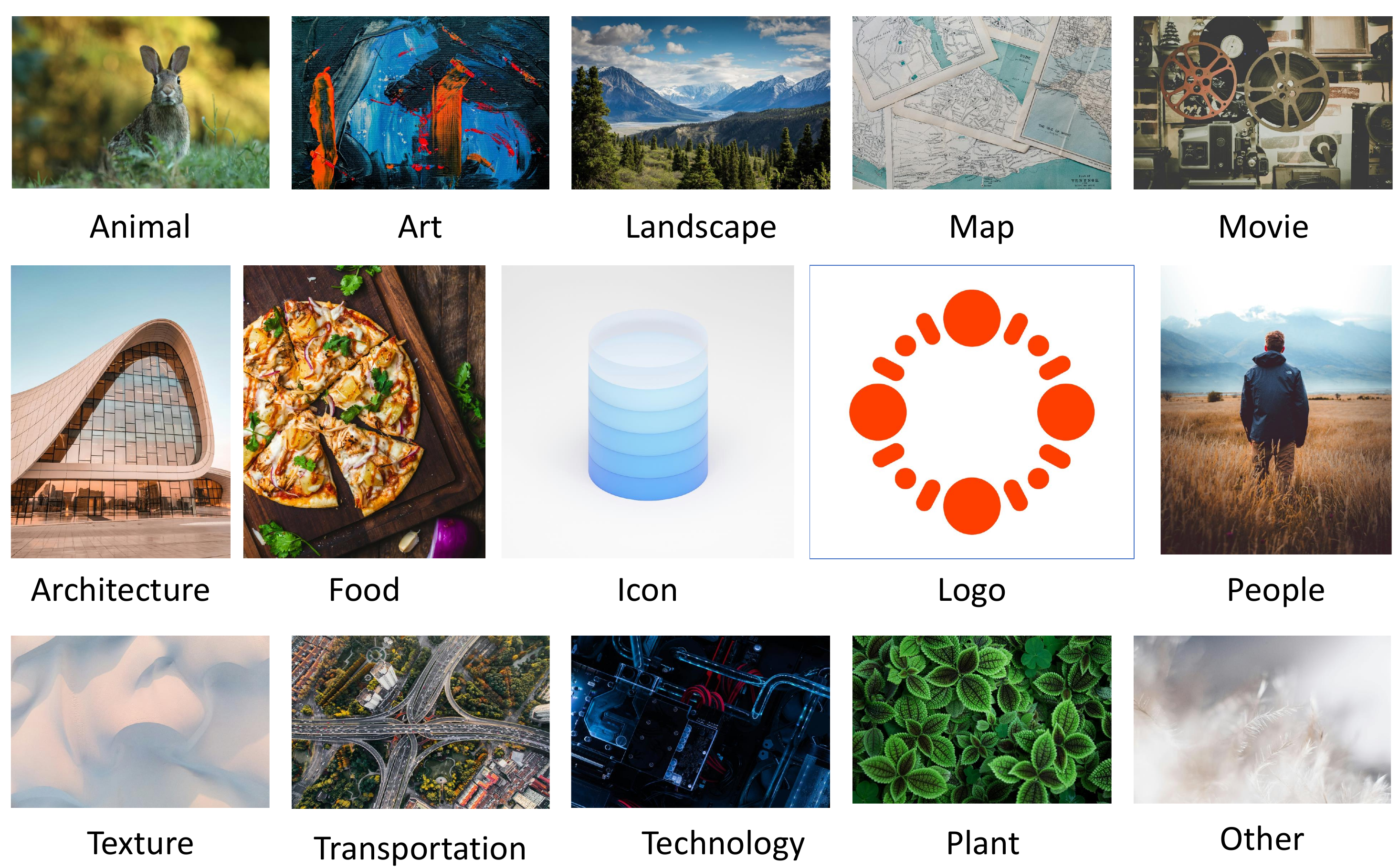}
  \caption{The 15 predefined image categories used in FullFront for standardized image representation.}
  \label{fig:category_imgs}
\end{figure*}

Regarding images, instead of using simple placeholders, we employ a category-based utilization strategy. We classify common real-world image content into 15 predefined categories: People, Animal, Food, Plant, Landscape, Icon, Logo, Architecture, Technology, Transportation, Map, Texture, Art, Movie, and Other (visualized in Figure \ref{fig:category_imgs}). Each category is linked to a fixed, non-copyrighted image URL following a standardized format, such as ``https://fixed\_part/\{Category\}.jpg''.
During the ground truth generation for code tasks, GPT-4o and Claude 3.7 Sonnet are instructed to select an appropriate category for any required image and use its corresponding URL. For evaluation, when an MLLM is tasked with generating webpage code, it must understand the image content from the provided screenshot, classify it into one of these 15 categories, and then generate HTML using the correct category-specific URL. Furthermore, because the intrinsic dimensions of these repository images are unknown, the MLLM is explicitly required to manually set the image sizes (width and height) and position within the HTML code to ensure the rendered output matches the layout depicted in the screenshot.
This approach assesses MLLMs' capabilities in image perception, categorization, and appropriate styling. It also ensures visual consistency for subsequent evaluation steps. Crucially, by deriving visual designs from real-world screenshots, our method generates webpages with greater diversity compared to ``from scratch'' techniques. This strategy ingeniously bypasses the laborious simplification of real-world code while still achieving simplification's primary goals---such as removing sensitive information and external dependencies---and preserving as much original visual information as possible through categorized representation. The use of category-specific image URLs also facilitates straightforward dataset expansion with new image types in the future.

\subsection{Image to Code}
The Image to Code subtask serves as the foundational evaluation for webpage implementation. It is designed to assess an MLLM's core ability to translate a static visual design into functional HTML code. The subtask comprises 200 distinct samples. During the evaluation, each MLLM is provided with a single webpage screenshot, which has been rendered from the high-fidelity HTML-v2 ground truth. The objective is to generate code that accurately replicates the visual appearance of the given image. This is the most direct and fundamental screenshot-to-code challenge within our benchmark, establishing a baseline for model performance.

\begin{figure*}[t]
  \centering
  \includegraphics[width=0.98\textwidth]{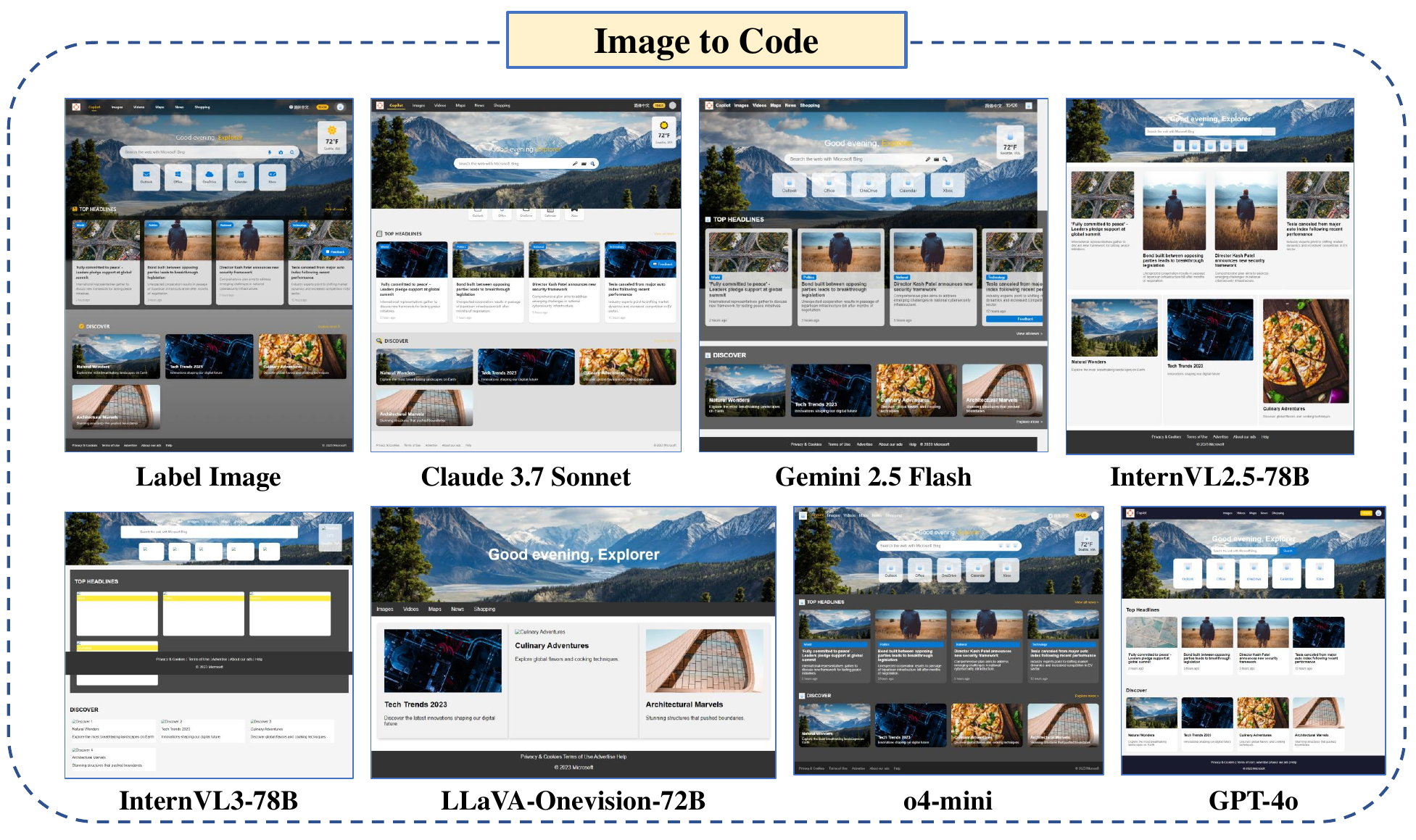}
  \caption{A sample case in Image to Code.}
  \label{fig:image-to-code}
\end{figure*}

\begin{figure*}[t]
  \centering
  \includegraphics[width=0.98\textwidth]{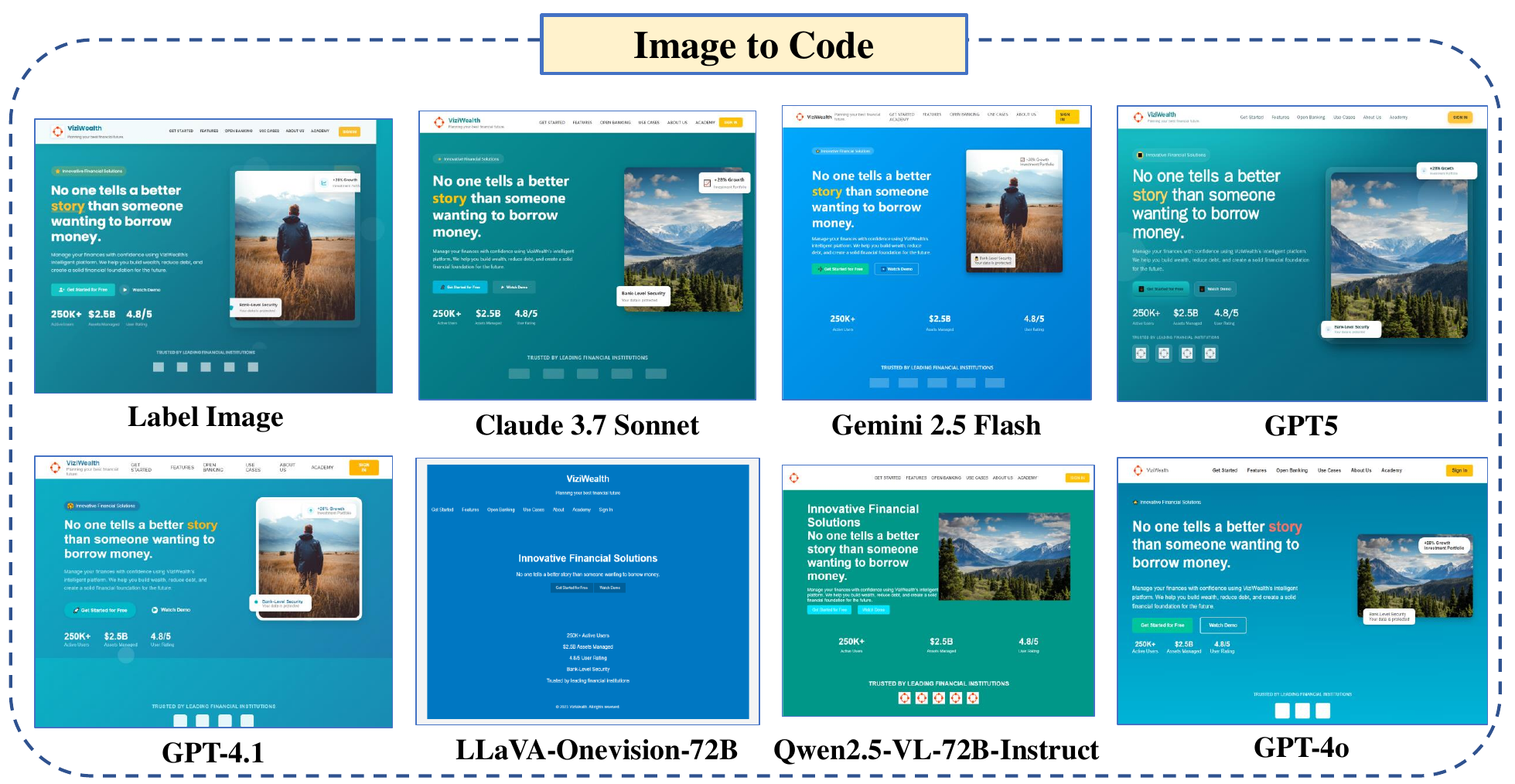}
  \caption{A sample case in Image to Code.}
  \label{fig:image-to-code1}
\end{figure*}

Figure~\ref{fig:image-to-code} and Figure~\ref{fig:image-to-code1} provide qualitative examples of various MLLM outputs on this task. The ``Label Image'' in each figure is the ground-truth screenshot provided as input to the models. A visual inspection reveals a clear performance gap between proprietary and open-source models.

Leading models like Claude 3.7 Sonnet, Gemini 2.5 Flash, and GPT-5 demonstrate a strong capability to grasp and reproduce the overall page layout, color schemes, and general style. However, they are not without flaws and often fail on fine-grained details. For example, as shown in Figure\ref{fig:image-to-code1}, Claude 3.7 Sonnet incorrectly identifies the image category, while Gemini 2.5 Flash fails to apply the correct background color to the page.

In contrast, the performance of the open-source models is generally less consistent. As illustrated across both figures by models like LLaVA-Onevision-72B and InternVL3-78B, there is substantial room for improvement in fundamental areas such as overall layout integrity, the correct implementation of headers and footers, and the accurate rendering of text and image elements.

\subsection{Code Refinement}
The Code Refinement subtask is designed to simulate a realistic and critical developer workflow: enhancing and optimizing existing code. Comprising 200 samples, this task evaluates an MLLM's ability to modify and augment existing webpage code in order to achieve a target visual design. In this scenario, the model is provided with the HTML-v1 along with the Label Image rendered from the corresponding HTML-v2. The model's objective is to modify and augment the provided HTML-v1 code to match the superior visual quality and complexity of the Label Image.

\begin{figure*}[t]
  \centering
  \includegraphics[width=0.98\textwidth]{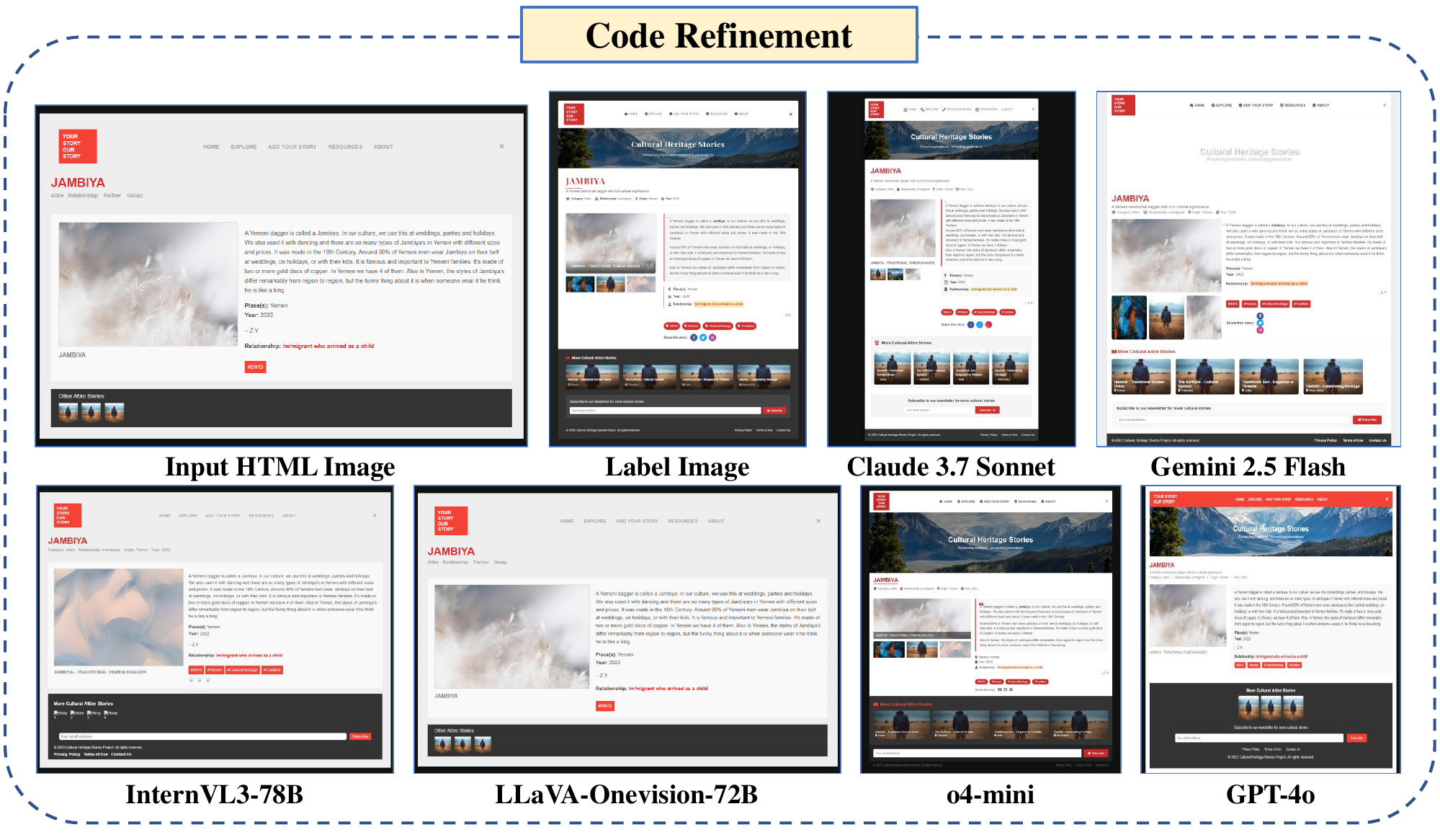}
  \caption{A sample case in Code Refinement.}
  \label{fig:code_refinement}
\end{figure*}

\begin{figure*}[t]
  \centering
  \includegraphics[width=0.98\textwidth]{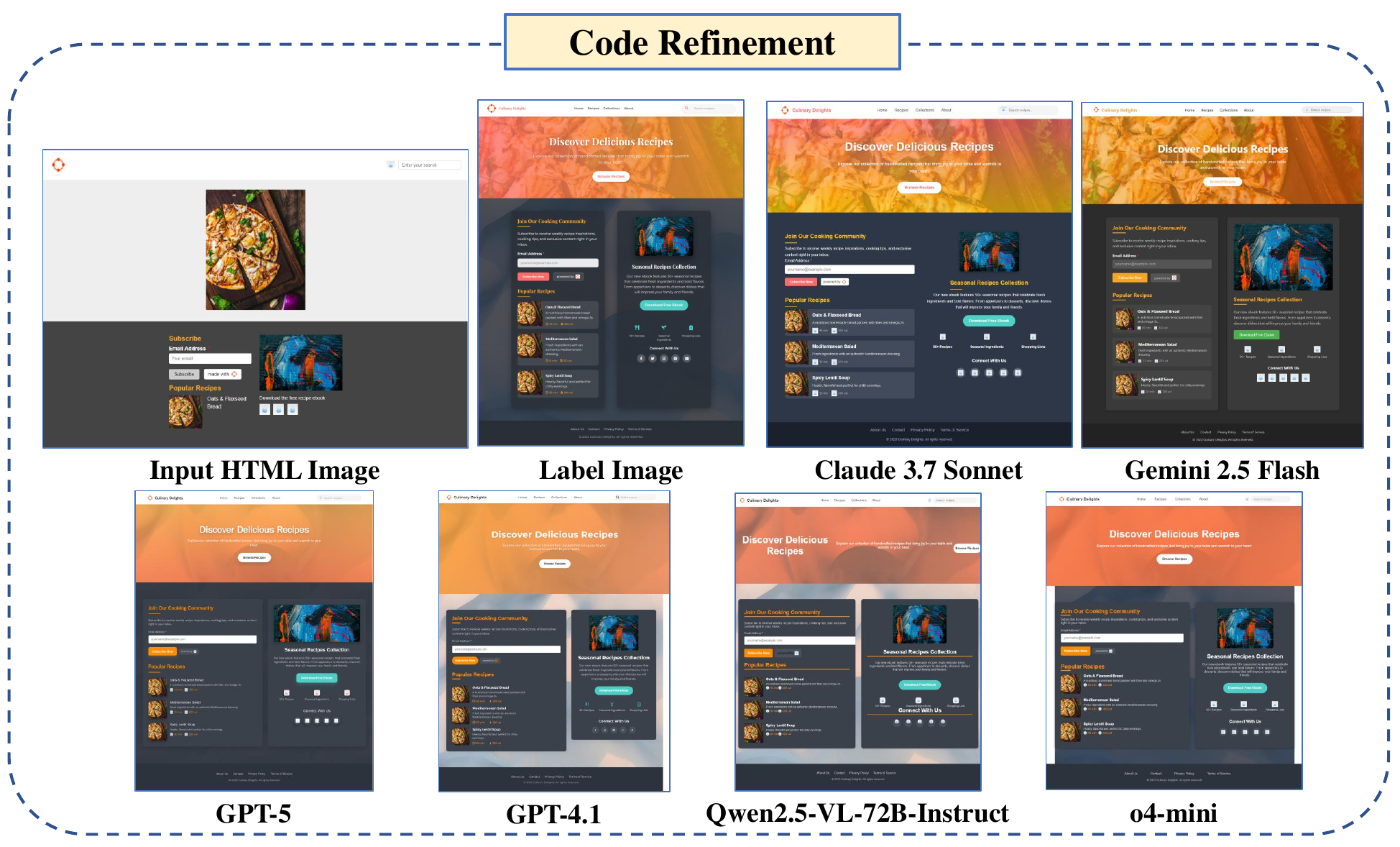}
  \caption{A sample case in Code Refinement.}
  \label{fig:code_refinement1}
\end{figure*}

As illustrated in the examples in Figure~\ref{fig:code_refinement} and Figure~\ref{fig:code_refinement1}, the results highlight a significant performance disparity. Advanced closed-source models, notably Claude 3.7 Sonnet, o4-mini, and GPT-5, demonstrate a strong capability to perform this complex task. They successfully identify the discrepancies between the input and target visuals and apply appropriate changes to the base code to closely reproduce the Label Image. While their outputs are highly accurate, there remains some room for improvement in perfecting fine-grained details related to headers, footers, text content, and element proportions.

In stark contrast, many open-source models exhibit a more fundamental difficulty with the task's core instruction. Rather than refining the provided HTML-v1, models such as InternVL3-78B and LLaVA-Onevision-72B often disregard the modification requirement entirely. Their outputs tend to be a near-verbatim copy of the initial code provided, showing minimal or no changes based on the target Label Image. This indicates a significant challenge not only in their code generation abilities but also in their capacity to comprehend and execute complex instructions that involve editing, rather than simply generating, code.

\subsection{Text to Code}
The Text to Code subtask evaluates an MLLM's ability to function as both a designer and a developer by generating a complete webpage based solely on a natural language description. This subtask includes 200 samples. For each, a detailed textual description was generated by Claude 3.7 Sonnet based on a high-fidelity HTML-v2 rendered page, followed by a rigorous manual review to ensure accuracy and quality. During evaluation, models are provided only with this text, making this a ``blind'' generation task where no visual reference is available. The goal is to translate the abstract requirements into a concrete and visually accurate webpage.

\begin{figure*}[t]
  \centering
  \includegraphics[width=0.98\textwidth]{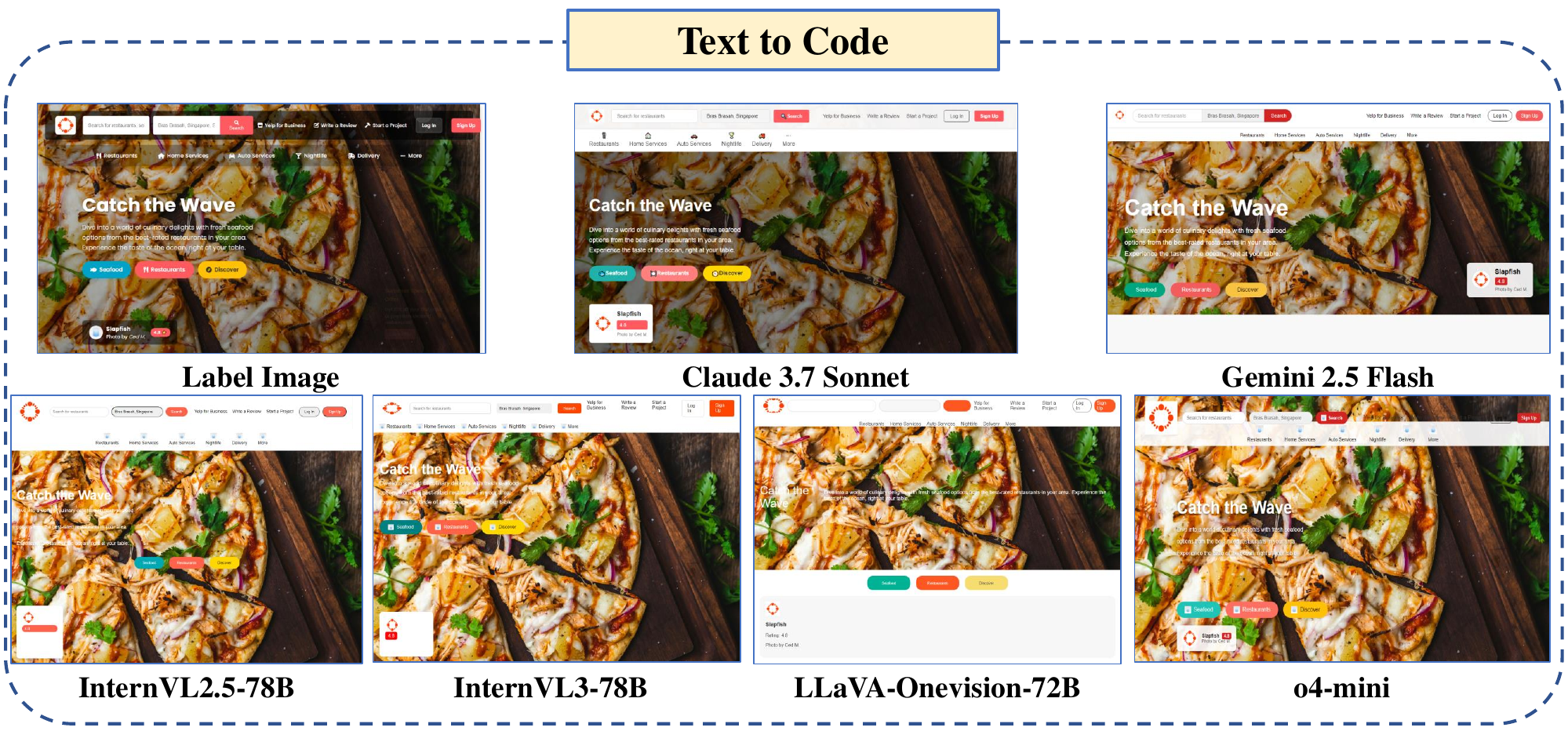}
  \caption{A sample case in Text to Code.}
  \label{fig:Text_to_Code}
\end{figure*}

\begin{figure*}[t]
  \centering
  \includegraphics[width=0.98\textwidth]{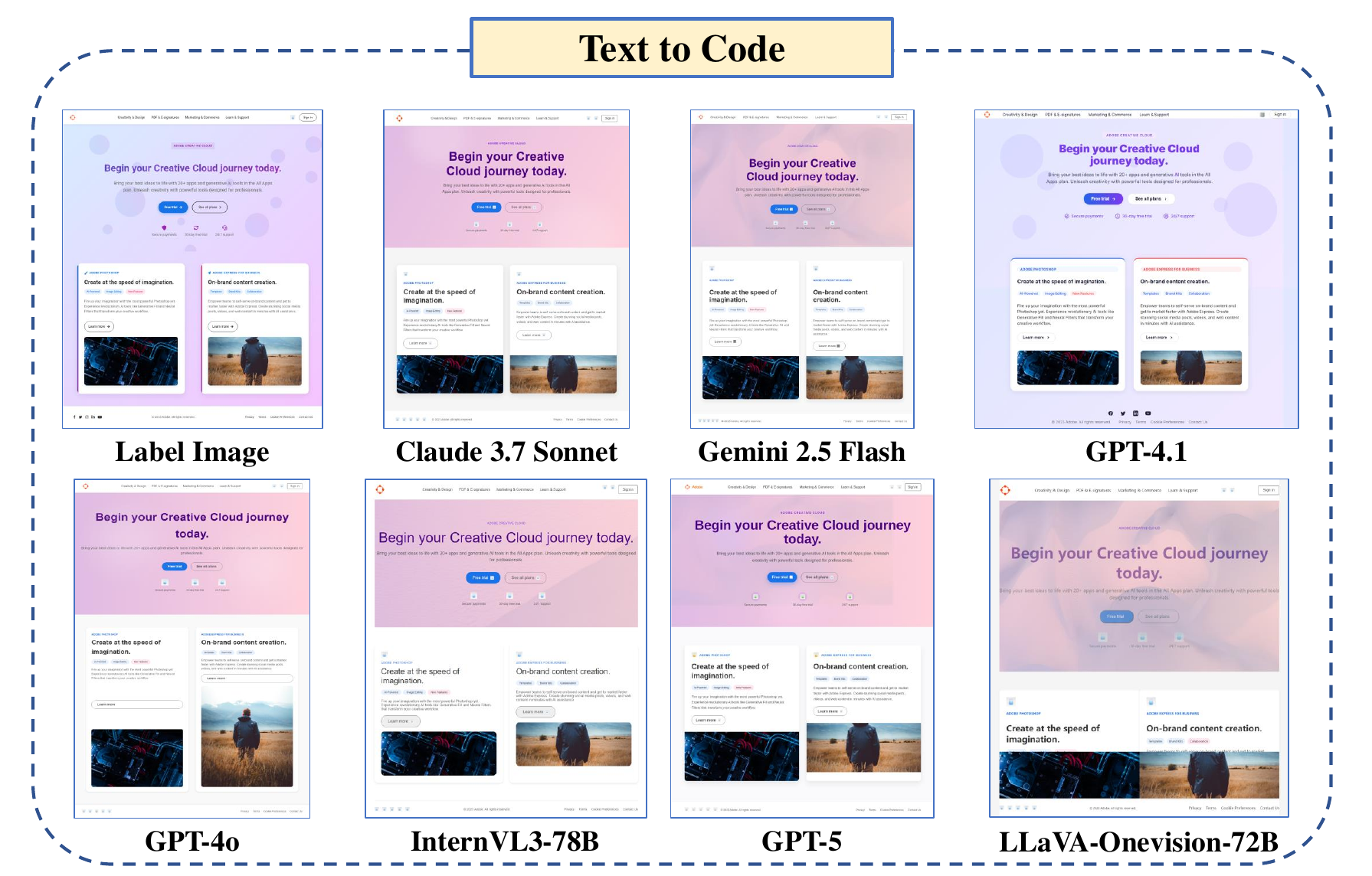}
  \caption{A sample case in Text to Code.}
  \label{fig:Text_to_Code1}
\end{figure*}

The qualitative results are presented in Figure~\ref{fig:Text_to_Code} and Figure~\ref{fig:Text_to_Code1}. It is important to note that the ``Label Image'' is shown here only for reference; it represents the original webpage that was described but was not provided to the models during the test.

The examples reveal two key findings. First, when faced with a more unique design, such as the one in Figure~\ref{fig:Text_to_Code}, the models successfully capture the main theme---a page with a large, food-themed background. However, they share a common, subtle error: rendering the header with a solid white background instead of the correct transparent one. This suggests that while high-level concepts are well-understood, models may default to common design patterns for details not explicitly or strongly specified. Second, for webpages with more conventional, standardized layouts, as seen in Figure~\ref{fig:Text_to_Code1}, the models perform remarkably well. Most are able to accurately reproduce the page's structure, layout, and style from the text alone. This demonstrates their significant potential in the front-end development workflow, particularly for users without professional design or coding expertise. The ability to generate high-quality prototypes directly from a textual brief can dramatically lower the barrier to web creation and accelerate the initial stages of development.

\subsection{Interaction Authoring}
Moving beyond static page generation, this task evaluates MLLMs' ability to implement dynamic, interactive webpages. Inspired by Interaction2Code~\citep{xiao2024interaction2code}, we define ten common interaction types, categorized under click and hover events. For data construction, 100 samples derived from the static HTML-v1 (allowing models to focus primarily on interaction logic) are augmented with interaction code by Claude 3.7 Sonnet, followed by manual verification.

During testing, MLLMs receive ``before'' and ``after'' interaction screenshots and must reproduce the static page while implementing the depicted interactive behavior. To facilitate automated interaction detection, MLLMs are instructed to assign the ID ``\texttt{\#InteractionPart}'' to the primary HTML element involved. The ten defined interaction types, with specific implementation requirements for each, are:
\begin{enumerate}
    \item \textbf{Click to Display Dropdown (\texttt{Interaction\_click\_1}):} An element, when clicked, reveals a dropdown menu whose content, position, and style are contextually adapted. Requires \texttt{aria-expanded} attribute toggling and specific dropdown selectors.
    \item \textbf{Click to Toggle Checkbox (\texttt{Interaction\_click\_2}):} A clickable checkbox that toggles its checked/unchecked state. Must use an \texttt{<input type="checkbox">} or an element with \texttt{role='checkbox'}, displaying a checked state after interaction.
    \item \textbf{Click to Change Background Color (\texttt{Interaction\_click\_3}):} An element significantly changes its background color upon being clicked, with the new color being distinct and detectable.
    \item \textbf{Click to Display Modal/Dialog (\texttt{Interaction\_click\_4}):} Clicking an element triggers a modal window or dialog box with contextually generated content and styling. The modal must match specific selectors like \texttt{.modal} or \texttt{[aria-modal='true']}.
    \item \textbf{Click to Display Tooltip (\texttt{Interaction\_click\_5}):} An element, when clicked, displays a tooltip providing additional information. The tooltip must adhere to specified class names or attributes (e.g., \texttt{.tooltip}, \texttt{[role='tooltip']}).
    \item \textbf{Click to Display Text Input (\texttt{Interaction\_click\_6}):} Clicking an element reveals a text box or input area for user entry, appropriately sized and positioned.
    \item \textbf{Hover to Display Dropdown (\texttt{Interaction\_hover\_1}):} A dropdown menu appears when the mouse hovers over an element, with adaptive content and styling.
    \item \textbf{Hover to Bold Text (\texttt{Interaction\_hover\_2}):} Text within an element becomes bold (fontWeight $\geq$600 or `bold`/`bolder`) upon mouse hover.
    \item \textbf{Hover to Underline Text (\texttt{Interaction\_hover\_3}):} Text within an element gains an underline when hovered over, with the computed \texttt{textDecoration} including ``underline''.
    \item \textbf{Hover to Display Tooltip (\texttt{Interaction\_hover\_4}):} A tooltip with additional information appears when the mouse hovers over an element, conforming to specific class or attribute requirements.
\end{enumerate}
The models must determine the correct interaction type from the visual cues and implement it according to these detailed specifications, providing the complete HTML, CSS, and JavaScript in a single file.

\textbf{Model Performance Analysis.} As shown in \textbf{Table \ref{table:interaction_rate}}, there is a significant performance gap between closed-source and open-source models in interaction implementation. Top-tier models like o4-mini and GPT-5 achieve over 90\% success rates, demonstrating robust capabilities in dynamic front-end logic. In contrast, open-source models struggle with this task, with success rates generally falling below 60\%.

\begin{table*}[htbp]
    \centering
    \setlength{\tabcolsep}{0pt}
    \begin{tabular*}{\textwidth}{@{\extracolsep{\fill}} l cc @{}}
        \toprule
        \textbf{Model} & \textbf{Interaction Success Rate} & \textbf{Interaction Success Rate (mini)} \\
        \midrule
        Qwen2.5-VL-72B-Instruct & 57.00 & 64.00 \\
        InternVL2.5-78B & 47.00 & 54.00 \\
        InternVL3-78B & 48.00 & 56.00 \\
        LLaVA-Onevision-72B & 16.00 & 24.00 \\
        \hdashline
        Claude 3.7 Sonnet & 78.00 & 86.00 \\
        Gemini 2.5 Flash & 70.00 & 72.00 \\
        GPT-4o & 80.00 & 74.00 \\
        o4-mini & \textbf{93.00} & \textbf{94.00} \\
        GPT-4.1 & 78.00 & 80.00 \\
        GPT-5 & \uline{92.00} & \uline{92.00} \\
        \bottomrule
    \end{tabular*}
\caption{Interaction success rate (\%). ``(mini)'' indicates the results on the FullFront-mini.}
\label{table:interaction_rate}

    \vspace{-5pt}
\end{table*}

\begin{table*}[htbp]
\centering
\resizebox{\textwidth}{!}{
\begin{tabular}{lcccccccccc}
\toprule
\textbf{Model} & \textbf{C1} & \textbf{C2}  & \textbf{C3} & \textbf{C4} & \textbf{C5} & \textbf{C6} & \textbf{H1} & \textbf{H2} & \textbf{H3} & \textbf{H4} \\
\midrule
    Qwen2.5-VL-72B-Instruct & 9 & 7 & 6 & 7 & 5 & 8 & 3 & 2 & 2 & 8 \\
    InternVL2.5-78B & 7 & 6 & 6 & 8 & 4 & 6 & 0 & 2 & 4 & 4 \\
    InternVL3-78B & 7 & 3 & 5 & 7 & 6 & 4 & 1 & 7 & 1 & 7 \\
    LLaVA-Onevision-72B & 8 & 0 & 7 & 0 & 0 & 0 & 0 & 0 & 0 & 1 \\
    \hdashline
    Claude 3.7 Sonnet & 9 & 8 & 5 & 10 & 9 & 9 & 6 & 3 & 9 & 10 \\
    Gemini 2.5 Flash & 7 & 9 & 6 & 9 & 8 & 9 & 6 & 1 & 8 & 7 \\
    GPT-4o & 10 & 8 & 8 & 10 & 8 & 7 & 9 & 6 & 5 & 9\\
    o4-mini & 9 & 10 & 10 & 10 & 8 & 10 & 9 & 8 & 10 & 9 \\
    GPT-4.1 & 10 & 5 & 8 & 10 & 8 & 9 & 7 & 8 & 3 & 10\\
    GPT-5 & 8 & 9 & 8 & 10 & 10 & 10 & 8 & 9 & 10 & 10\\
    \hdashline
    Average & 8.4 & 6.5 & 6.9 & 8.1 & 6.6 & 7.2 & 4.9 & 4.6 & 5.2 & 7.5 \\
\bottomrule
\end{tabular}
}
\setlength{\tabcolsep}{6pt}
\caption{Performance of MLLMs across ten distinct interaction categories (C1-C6 for click-based, H1-H4 for hover-based) in the Interaction Authoring subtask. Each category comprises 10 test cases.}
\label{table:interaction_analysis}
\end{table*}

\textbf{Interaction Type Analysis.} Beyond overall model performance, our analysis of success rates across different interaction types (\textbf{Table \ref{table:interaction_analysis}}) revealed varying capabilities among the tested models. The MLLMs demonstrated the highest proficiency in implementing \texttt{Interaction\_click\_1} (\textbf{Click to Display Dropdown}), where, on average, models successfully realized this functionality in 8.4 out of the 10 test instances. Conversely, the most challenging categories proved to be \texttt{Interaction\_hover\_2} (\textbf{Hover to Bold Text}) and \texttt{Interaction\_hover\_1} (\textbf{Hover to Display Dropdown}). Notably, for these two hover-activated interaction types, no single model managed to successfully implement all 10 test cases.

\begin{figure*}[t]
  \centering
  \includegraphics[width=0.98\textwidth]{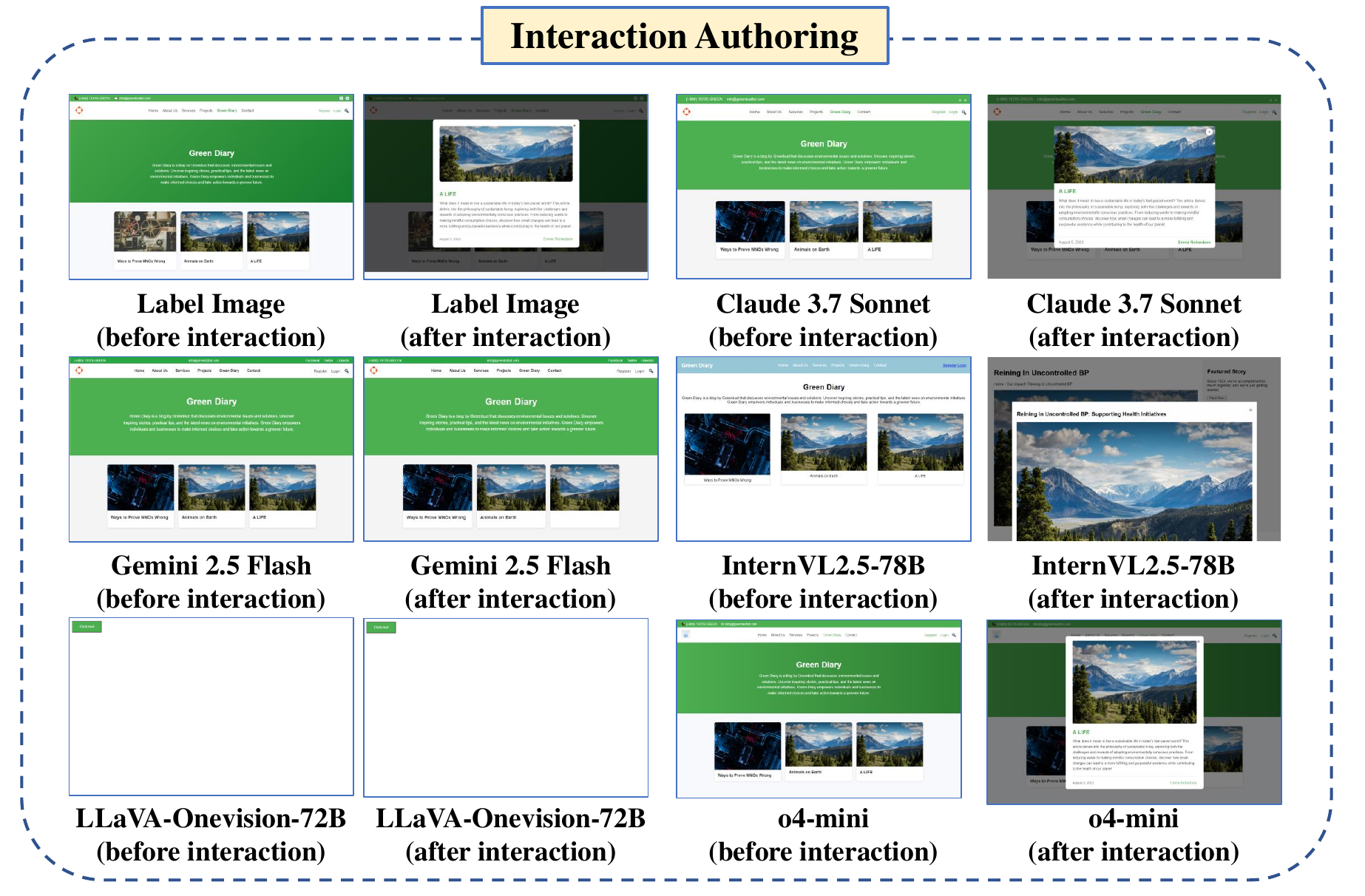}
  \caption{A sample case in Interaction Authoring.}
  \label{fig:Interaction_Authoring}
\end{figure*}

\begin{figure*}[t]
  \centering
  \includegraphics[width=0.98\textwidth]{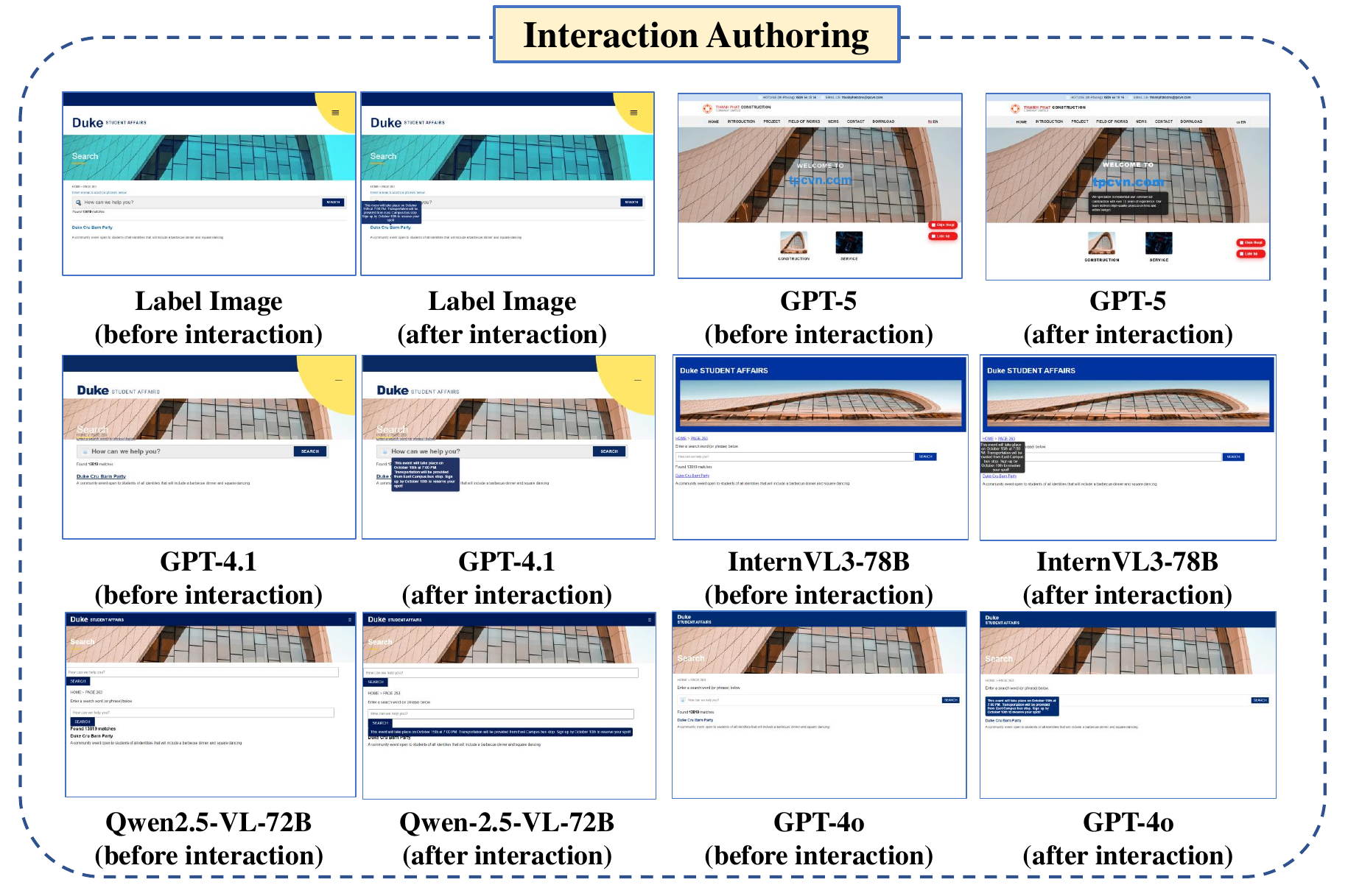}
  \caption{A sample case in Interaction Authoring.}
  \label{fig:Interaction_Authoring1}
\end{figure*}

The difficulty of this subtask is underscored by the qualitative results shown in Figure~\ref{fig:Interaction_Authoring} and Figure~\ref{fig:Interaction_Authoring1}. A critical observation is that the added complexity of implementing interaction often degrades the model's performance in rendering the static base page. Even though the static portion is based on the simpler HTML-v1, the visual fidelity of the replication is frequently lower than in static-only tasks. This suggests the cognitive load of generating interaction logic detracts from the model's ability to achieve precise visual accuracy. Furthermore, even when models successfully implement the correct interactive behavior (e.g., displaying a modal on click or a tooltip on hover), the final visual appearance of the ``after interaction'' state often shows significant discrepancies when compared to the ground-truth screenshot. These combined failures---in both static replication and dynamic fidelity---demonstrate a key deficiency in current MLLMs when it comes to creating robust, interactive web experiences.


\subsection{Three Common errors}

\begin{figure*}[t]
  \centering
  \includegraphics[width=0.98\textwidth]{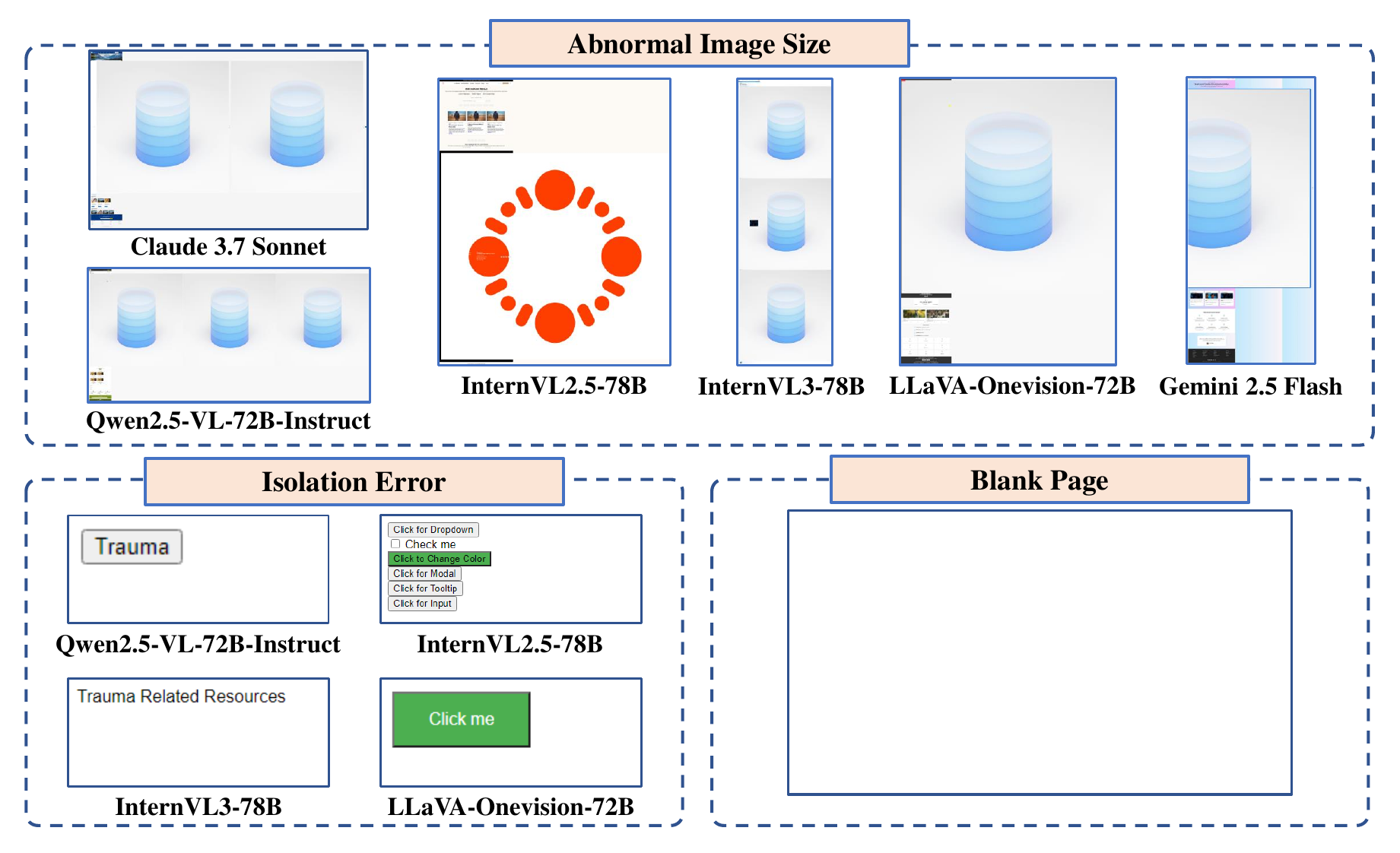}
  \caption{Three types of code errors.}
  \label{fig:three_code_errors}
\end{figure*}

Across the Webpage Code Generation subtasks, we identified three recurring and critical failure modes that affect both open- and closed-source models: Abnormal Image Size, Blank Page, and Isolation Error. These errors are illustrated in Figure~\ref{fig:three_code_errors}.

\begin{itemize}
    \item \textbf{Abnormal Image Size:} This error occurs when rendered images, often several at once, are displayed at disproportionately large sizes, severely compromising the webpage's layout and aesthetics. Our analysis reveals this issue is most prevalent with images categorized as ``Icon'' and ``Logo''. We hypothesize this is because these image types are numerous, typically small, and can be easily overlooked by the model when applying explicit size constraints, unlike more prominent content images. Interestingly, models would sometimes correctly size several icons on a page but fail on a single one, indicating a lack of consistent style application.
    \item \textbf{Isolation Error:} This failure mode is unique to the Interaction Authoring subtask. It manifests as an output where the model generates only the code for the requested interactive element, completely omitting the surrounding static content of the webpage. This error points to a model's limited ability to handle complex, multi-layered instructions; it appears to fixate on the primary directive (implement the interaction) while ignoring the implicit but essential requirement to reproduce the entire page.
    \item \textbf{Blank Page:} This error results in a completely empty page upon rendering, even when the generated code is not empty. The root cause typically traces back to a common failure mode in Large Language Models: repetitive generation. The model enters a loop, outputting the same content or code block repeatedly, which produces malformed HTML that a web browser cannot successfully parse and render.
\end{itemize}

In summary, the consistent occurrence of these three error types demonstrates that current MLLMs, regardless of their origin, have yet to achieve the stability and reliability required of a professional front-end engineer. Overcoming these fundamental failure modes is a critical step toward their practical application in autonomous web development.

\subsection{Text-Only LLM Results on Text to Code}
\label{appendix_text_llm}
To explore the potential of powerful text-only LLMs in front-end development, we conducted a supplementary experiment on our \textbf{Text to Code} subtask, benchmarking models like the DeepSeek series~\citep{DBLP:journals/corr/abs-2501-12948} and Kimi K2~\citep{DBLP:journals/corr/abs-2507-20534} against the MLLMs from the main evaluation.

\begin{table*}[htbp]
\centering
\resizebox{\textwidth}{!}{
\begin{tabular}{l cccc}
\toprule
\textbf{Model} & \textbf{Gemini Visual Score $\uparrow$} & \textbf{Clip Score $\uparrow$} & \textbf{Code Score $\uparrow$} & \textbf{DINOv2 Score $\uparrow$}\\
\hline
Qwen2.5-VL-72B-Instruct & 5.1750 & 0.7313 & 0.4854 & 0.4665 \\
InternVL2.5-78B & 5.4444 & 0.7366 & 0.4989 & 0.4824 \\
InternVL3-78B & 4.9144 & 0.7204 & 0.4969 & 0.4405 \\
LLaVA-Onevision-72B & 5.1273 & 0.7365 & 0.4129 & 0.5006 \\
Claude 3.7 Sonnet & 8.1358 & 0.8655 & \textbf{0.6436} & 0.7790 \\
Gemini 2.5 Flash & 8.0111 & 0.8660 & 0.5927 & 0.7929 \\
GPT-4o & 6.1351 & 0.7364 & 0.4857 & 0.5176 \\
o4-mini & 6.8728& 0.8197 & 0.5052 & 0.6807 \\
GPT-4.1 & \uline{8.3555} & 0.8685 & 0.5509 & \uline{0.7960} \\
GPT-5 & \textbf{8.4780} & \textbf{0.8759} & 0.5886 & 0.7928 \\
\hdashline 
R1-Distill-Llama-8B & 2.9540 & 0.6991 & 0.4189 & 0.4014 \\
DeepSeek-V3 & 7.9985 & 0.8617 & 0.5885 & 0.7828 \\
DeepSeek-R1 & 7.9381 & 0.8607 & \uline{0.6090} & 0.7487 \\
Kimi K2 & 7.7675 & 0.8431 & 0.5422 & 0.7323 \\
DeepSeek-V3.1 & 7.8775 & \uline{0.8719} & 0.6050 & \textbf{0.7980} \\
\bottomrule
\end{tabular}
}
\caption{Performance Comparison of MLLMs and Text-Only LLMs on the \textbf{Text to Code} Subtask.}
\label{tab:text_to_code_text_LLM}

\end{table*}

\begin{figure*}[t]
  \centering
  \includegraphics[width=0.98\textwidth]{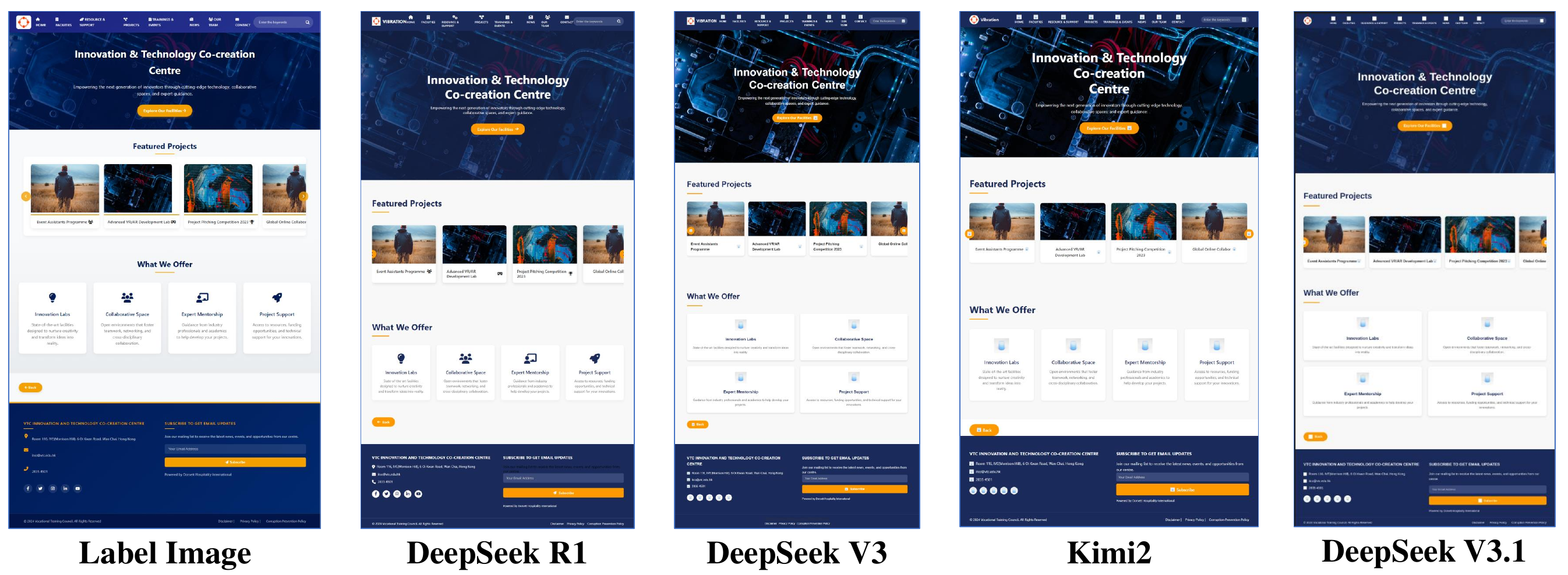}
  \caption{A qualitative comparison of webpages generated by leading text-only LLMs on the \textbf{Text to Code} subtask.}
  \label{fig:text_llm_sample1}
\end{figure*}

\begin{figure*}[t]
  \centering
  \includegraphics[width=0.98\textwidth]{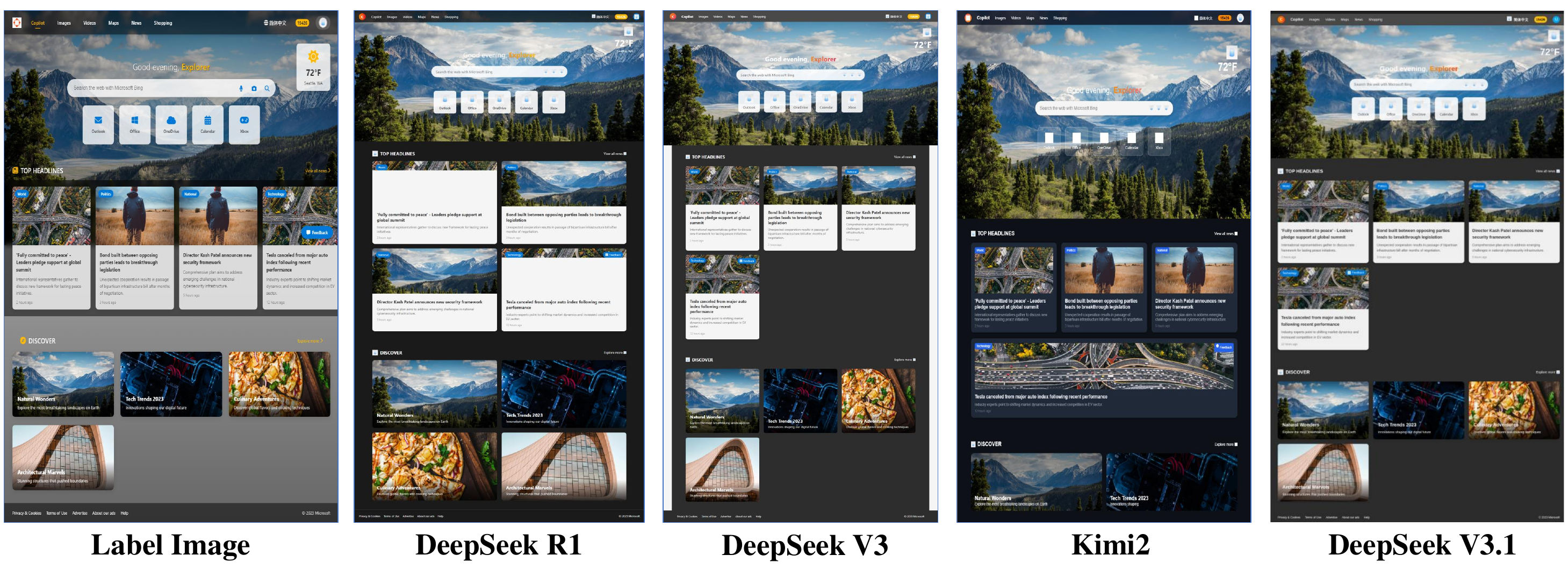}
  \caption{A qualitative comparison of webpages generated by leading text-only LLMs on the \textbf{Text to Code} subtask.}
  \label{fig:text_llm_sample2}
\end{figure*}

The results in Table~\ref{tab:text_to_code_text_LLM} are compelling. Despite lacking visual input, leading text-only models like \textbf{DeepSeek-R1} and \textbf{DeepSeek-V3.1} demonstrate highly competitive performance, achieving \textbf{Code Scores} (0.6090 and 0.6050 respectively) that not only surpass all open-source MLLMs but also rival top-tier closed-source MLLMs such as `Claude 3.7 Sonnet` (0.6436). Their strong performance is consistent across other visual and code-based metrics. This finding strongly suggests that with sufficiently detailed textual specifications, the advanced comprehension and coding abilities of text-only models can effectively compensate for the absence of a visual modality. As illustrated in Figure~\ref{fig:text_llm_sample1} and Figure~\ref{fig:text_llm_sample2}, this highlights their significant potential in text-driven front-end development workflows, where they can successfully reproduce a webpage's layout, element positioning, and overall style.

\subsection{Correct Code Implementation Despite Perceptual Errors}
\label{appendix_perceptual_code}
\begin{figure*}[t]
  \centering
  \includegraphics[width=0.98\textwidth]{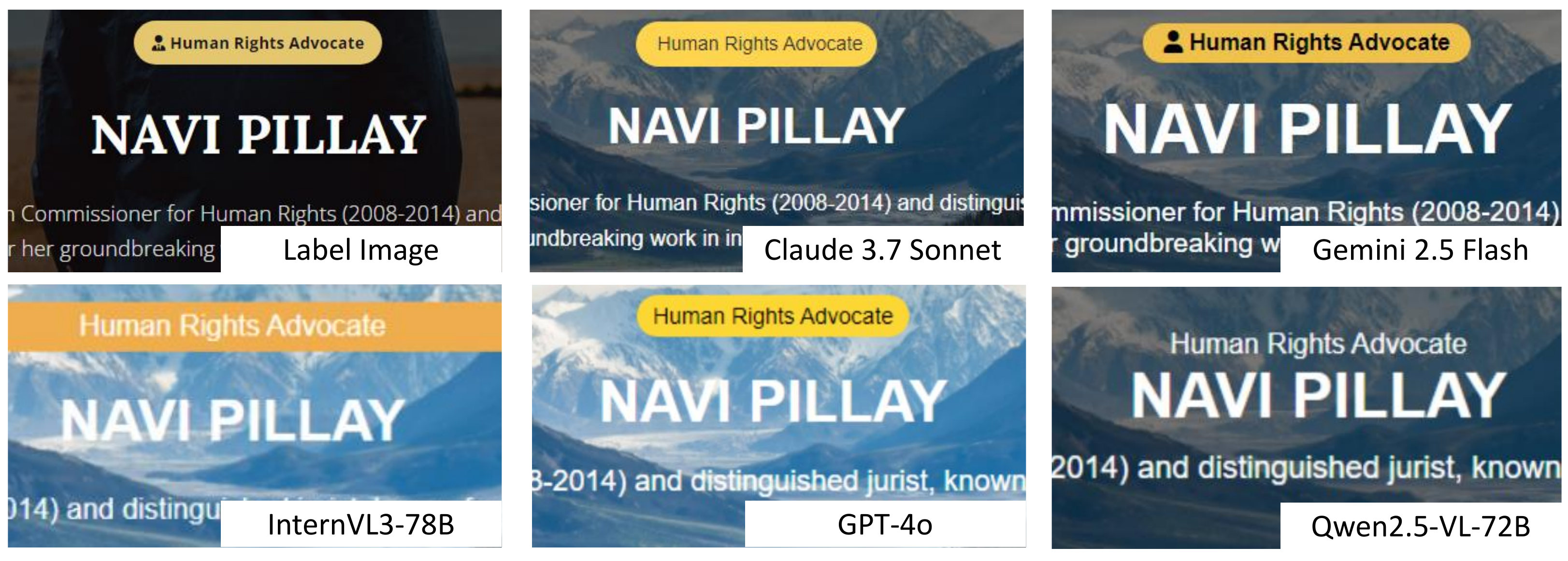}
  \caption{Rendered outputs from various MLLMs for the ``NAVI PILLAY'' webpage section. Despite failing the perceptual QA task regarding the tag's position relative to the title, all these MLLMs correctly implement the ``Human Rights Advocate'' tag above the main ``NAVI PILLAY'' title in their generated code.}
  \label{fig:error_analysis}
\end{figure*}

This part provides a visual illustration supporting the discussion in Section \ref{sec:perception_code_relationship}, which highlights an intriguing discrepancy between MLLM performance on perceptual QA tasks and their ability to generate visually accurate code. Specifically, as detailed in Figure \ref{fig:qa_errors} (b) of the main paper, all evaluated MLLMs incorrectly identify the positioning of the ``Human Rights Advocate'' tag relative to the main title (``NAVI PILLAY'') and subtitle in the Webpage Perception QA task.
However, when these same MLLMs are tasked with generating the webpage code, they often demonstrate correct implementation of this very element's placement. Figure \ref{fig:error_analysis} presents the rendered outputs for the ``NAVI PILLAY'' webpage section from several MLLMs benchmarked in FullFront. These outputs are derived from the code generation tasks where models are asked to reproduce the webpage.

As can be observed in Figure \ref{fig:error_analysis}, despite their prior failure in the specific perceptual QA question regarding the tag's position, all depicted MLLM outputs correctly place the ``Human Rights Advocate'' tag directly above the main ``NAVI PILLAY'' title. This placement is consistent with the ground-truth webpage.

\subsection{Detailed Visual Analysis of Perception-Code Quadrants}
\label{appendix_perceptual_code_examples}

To provide concrete evidence for the decoupling phenomenon discussed in Section~\ref{sec:perception_code_relationship}, we present representative case studies for each of the four behavior quadrants. These examples illustrate how models succeed or fail in bridging the gap between visual perception (QA accuracy) and implementation (code generation). For clarity, we display cropped sections of the full webpages relevant to the specific QA pairs.

\begin{figure*}[h]
    \centering
    \includegraphics[width=0.98\textwidth]{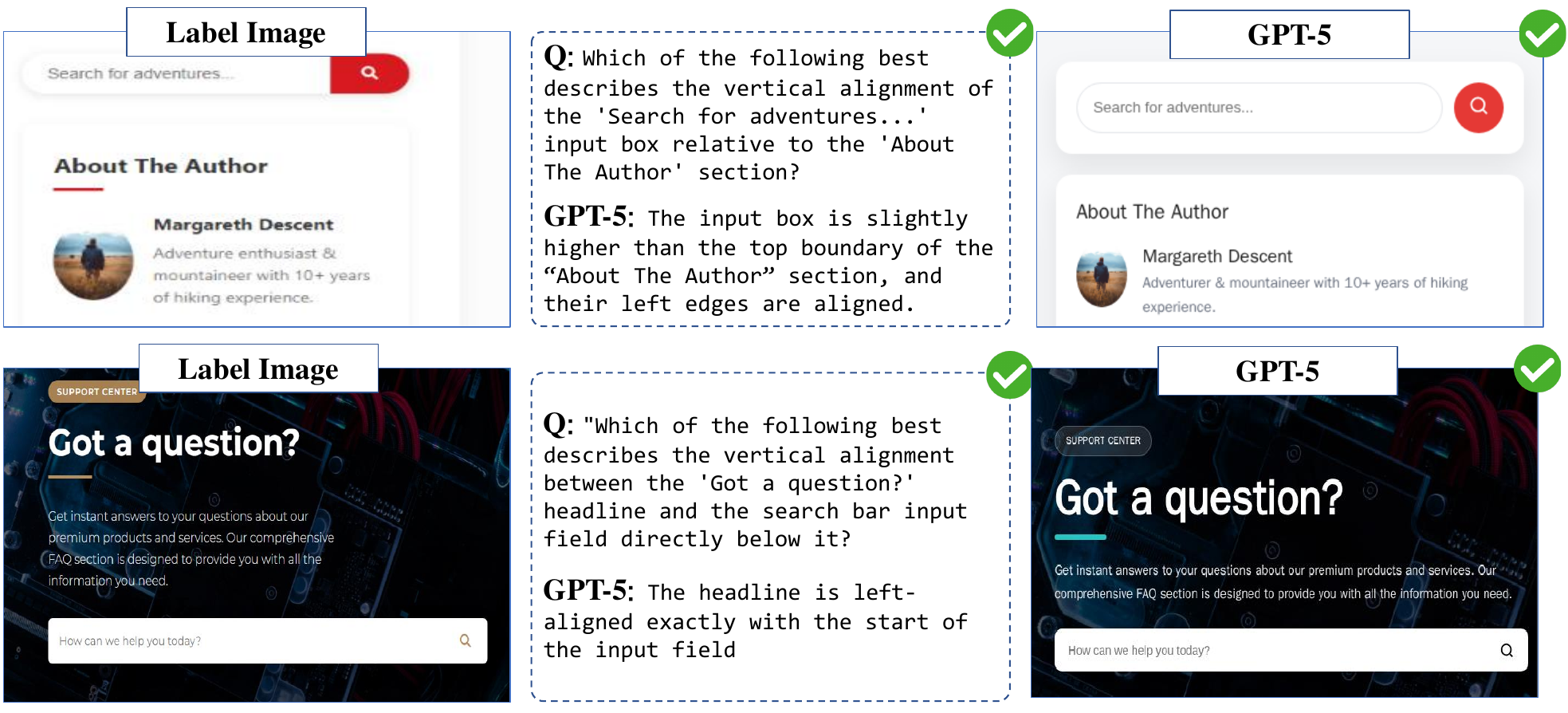}
    \caption{Quadrant 1 ($p\checkmark c\checkmark$): Examples where GPT-5 correctly perceives spatial relationships (top: vertical alignment; bottom: left alignment) and successfully implements them in code.}
    \label{fig:quadrant_yesyes}
\end{figure*}

\textbf{Quadrant 1: Correct Perception, Correct Implementation ($p\checkmark c\checkmark$).}
Figure~\ref{fig:quadrant_yesyes} demonstrates the ideal scenario where models successfully ground their visual understanding into code. In the first example, GPT-5 accurately perceives the vertical alignment between the search input box and the ``About The Author'' section, and faithfully reproduces this layout. Similarly, in the second example, GPT-5 correctly identifies and implements the left-alignment of the ``Got a question?'' headline relative to the search bar.

\begin{figure*}[h]
    \centering
    \includegraphics[width=0.98\textwidth]{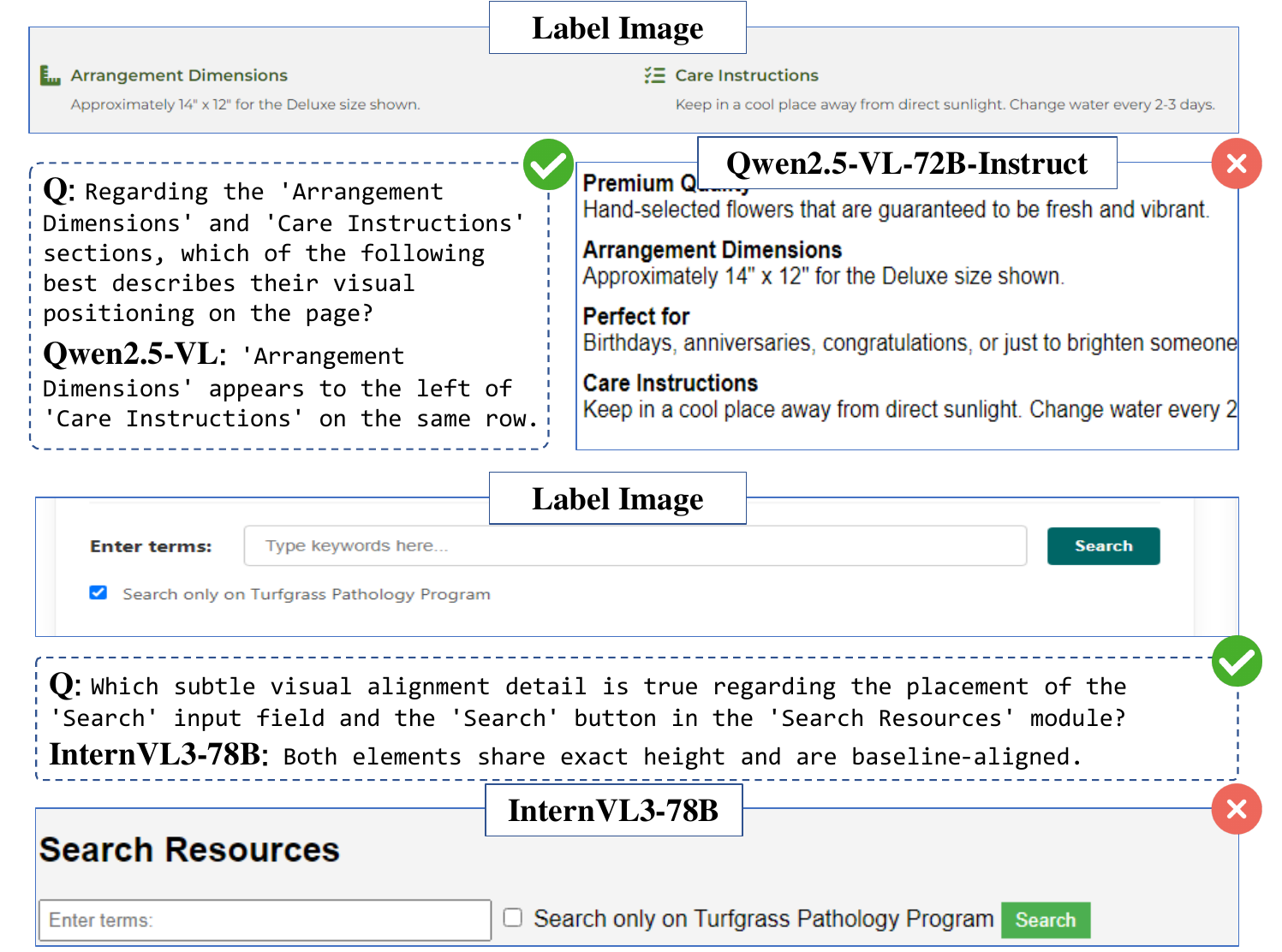}
    \caption{Quadrant 2 ($p\checkmark c\times$): Examples of Coding Bottlenecks. Qwen2.5-VL (top) and InternVL3 (bottom) correctly answer the perceptual questions but fail to translate this understanding into the correct layout or element arrangement.}
    \label{fig:quadrant_yesno}
\end{figure*}

\textbf{Quadrant 2: Correct Perception, Incorrect Implementation ($p\checkmark c\times$).}
Figure~\ref{fig:quadrant_yesno} highlights the ``Coding Bottleneck'' typical of open-source models. In the top example, Qwen2.5-VL-72B-Instruct correctly answers that ``Arrangement Dimensions'' is positioned to the left of ``Care Instructions'' on the same row. However, its generated code fails to achieve this layout, stacking the elements vertically instead. In the bottom example, InternVL3-78B correctly perceives the baseline alignment and height match of the search elements but erroneously inserts a checkbox between the input field and the button during implementation, breaking the intended visual structure.

\begin{figure*}[h]
    \centering
    \includegraphics[width=0.98\textwidth]{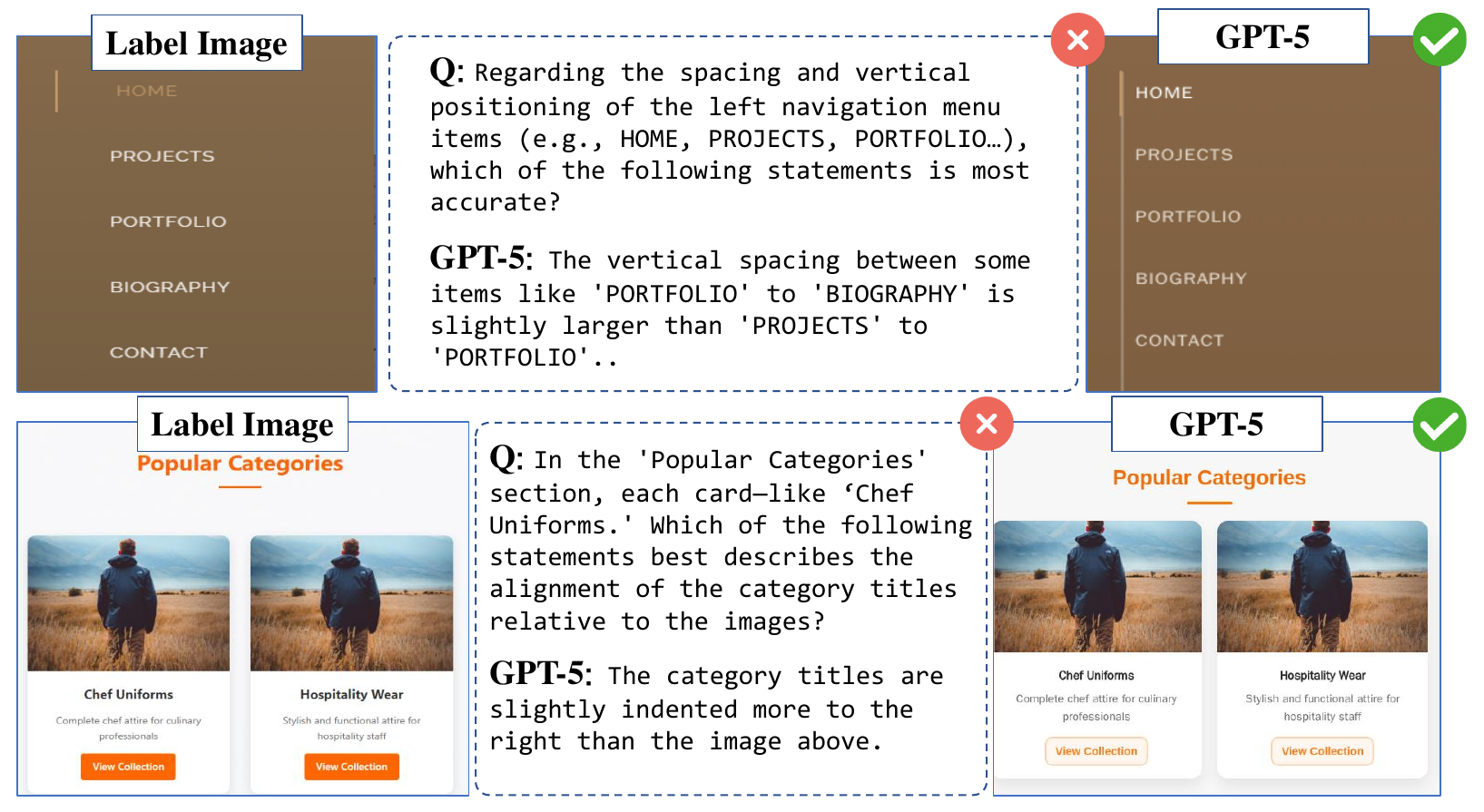}
    \caption{Quadrant 3 ($p\times c\checkmark$): Examples of Grounding Failure / Implicit Correction. GPT-5 hallucinates non-existent spacing (top) or alignment issues (bottom) in the QA phase but relies on strong priors to generate the correct, standard design patterns in the final code.}
    \label{fig:quadrant_noyes}
\end{figure*}

\textbf{Quadrant 3: Incorrect Perception, Correct Implementation ($p\times c\checkmark$).}
Figure~\ref{fig:quadrant_noyes} illustrates the ``Grounding Failure'' phenomenon prevalent in advanced proprietary models. In the first case, GPT-5 hallucinates a vertical spacing discrepancy in the navigation menu that does not exist (claiming ``PORTFOLIO'' to ``BIOGRAPHY'' is larger), yet its generated code produces perfectly even spacing. In the second case, GPT-5 incorrectly claims the category titles are indented to the right of the images. The ground truth (and the correct design principle) is center alignment, which the model successfully implements in its code despite the perceptual error. This strongly suggests reliance on internal design priors over direct visual grounding.

\begin{figure*}[h]
    \centering
    \includegraphics[width=0.98\textwidth]{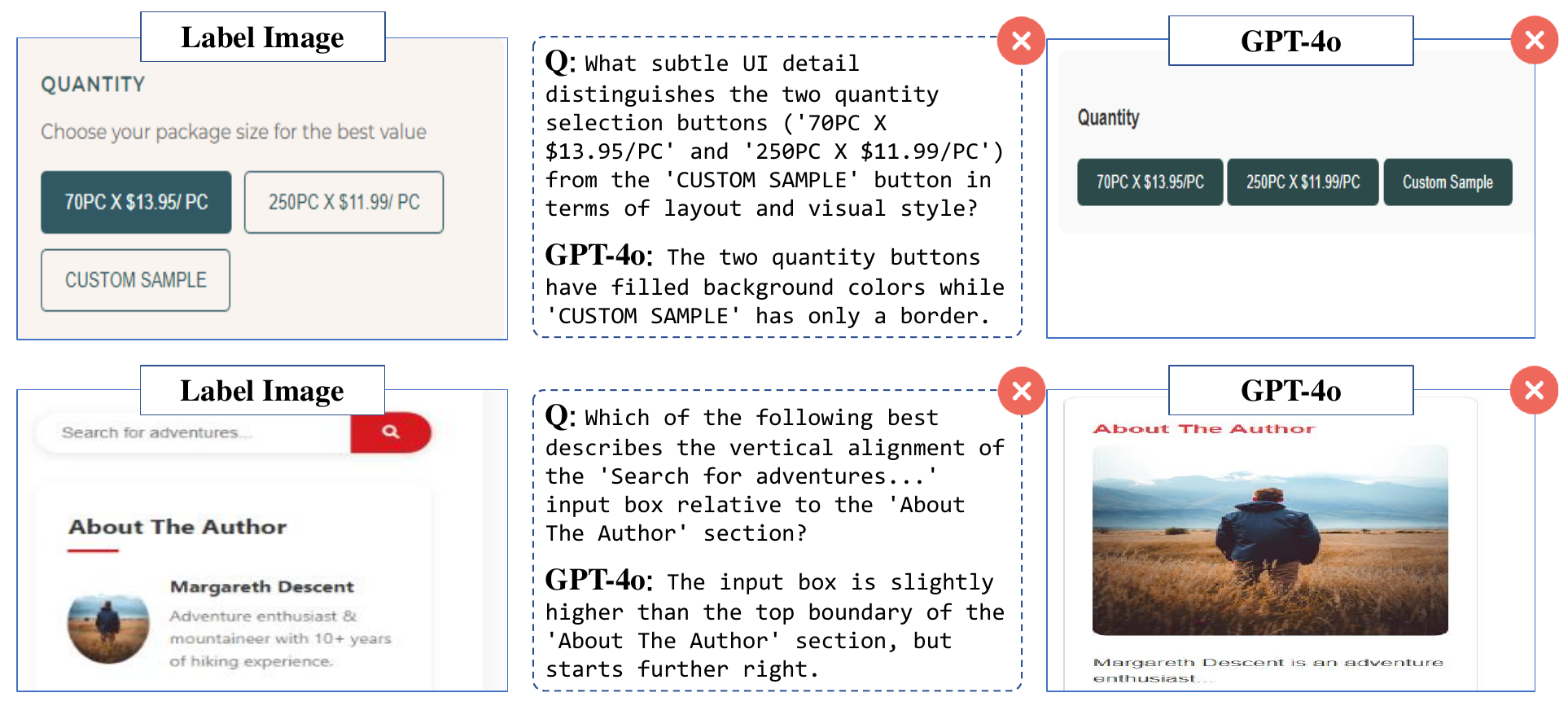}
    \caption{Quadrant 4 ($p\times c\times$): Examples of total failure. GPT-4o fails to perceive button styles (top) and element positioning (bottom), leading to code that either applies incorrect styles or omits key elements entirely.}
    \label{fig:quadrant_nono}
\end{figure*}

\textbf{Quadrant 4: Incorrect Perception, Incorrect Implementation ($p\times c\times$).}
Figure~\ref{fig:quadrant_nono} shows cases of complete failure. In the top example, GPT-4o misidentifies the styling of the ``CUSTOM SAMPLE'' button (claiming it has a border only), and proceeds to generate code where all buttons have filled backgrounds, contradicting the visual design. In the bottom example, GPT-4o not only fails to perceive the correct horizontal alignment of the search box relative to the author section but also fails to generate the search box entirely in the code, leading to a significant visual omission.

\begin{figure*}[p]
\centering
\begin{wideprompt}
\begin{promptbox}{Prompt for Gemini Visual Score}
Your task is to assess two webpage images and output a score between 0 and 10 for each of the following 10 questions, reflecting the degree of similarity between the webpages. A score of 10 indicates perfect similarity (identical in every aspect), while a score of 0 indicates no similarity. For partial similarities, assign a score between 1 and 9, where higher scores reflect greater similarity. Only output a comma-separated list of 10 numbers enclosed in square brackets, e.g., [10,8,6,4,2,0,0,0,0,0]. Do not assign a score of 10 unless the two images are identical in the respective category.

1. \textbf{Element Reproduction (Score: 0-10):} Are key elements (text, images, buttons) fully present and styled identically to the original design? (e.g., 10 for identical elements, 5 for missing or slightly altered elements, 0 for completely different elements.) \\
2. \textbf{Proportion and Size Consistency (Score: 0-10):} Do the sizes and proportions of elements (text, images, buttons) match the original design, maintaining visual harmony? (e.g., 10 for exact proportions, 6 for minor size differences, 0 for significant discrepancies.) \\
3. \textbf{Layout and Typography Fidelity (Score: 0-10):} Does the overall layout (headers, footers, navigation bars, sidebars) faithfully replicate the original design's typography and structure? (e.g., 10 for identical layouts, 5 for similar but not exact placements, 0 for entirely different layouts.) \\
4. \textbf{Alignment and Spacing Accuracy (Score: 0-10):} Are elements aligned and spaced (margins, padding) as in the original design? (e.g., 10 for perfect alignment and spacing, 6 for minor misalignments, 0 for major misalignments.) \\
5. \textbf{Visual Hierarchy Clarity (Score: 0-10):} Does the webpage maintain the same visual hierarchy as the original, allowing users to quickly identify key information? (e.g., 10 for identical hierarchy, 5 for slightly altered emphasis, 0 for unclear or different hierarchy.) \\
6. \textbf{Color Consistency (Score: 0-10):} Do the overall color scheme, hues, and tones match the original design? (e.g., 10 for identical colors, 6 for similar palette with minor variations, 0 for completely different colors.) \\
7. \textbf{Style Consistency (Score: 0-10):} Does the webpage's overall aesthetic style (e.g., modern, minimalistic) align with the original design? (e.g., 10 for identical style, 4 for similar but distinguishable style, 0 for entirely different style.) \\
8. \textbf{Text Style Consistency (Score: 0-10):} Are text attributes (font type, size, line spacing, paragraph spacing, alignment) consistent with the original design? (e.g., 10 for identical text styles, 5 for similar fonts with spacing issues, 0 for completely different text styles.) \\
9. \textbf{Text Content Accuracy (Score: 0-10):} Does the webpage accurately reproduce the main textual content of the original design? (e.g., 10 for identical text, 5 for partial matches, 0 for entirely different text.) \\
10. \textbf{Overall Content Representation (Score: 0-10):} Does the webpage convey the same content information and intent as the original design? (e.g., 10 for identical content representation, 6 for similar but incomplete content, 0 for entirely different content.) \\

\textbf{Output Format:} [score1,score2,score3,score4,score5,score6,score7,score8,score9,score10]
\end{promptbox}
    \end{wideprompt}
    \caption{Gemini Visual Score Prompt.}
    \label{fig:gemini_evaluate_prompt} 
\end{figure*}

\section{Evaluation}
\label{appendix_evaluation}

\subsection{Gemini Visual Score: Criteria and Rubric}
\label{sec:appendix_gemini_visual_score}
To facilitate a fine-grained and human-aligned visual assessment of MLLM-generated webpages, we employ the \textbf{Gemini 2.5 Flash} model as a sophisticated visual evaluator. This model is tasked with comparing a rendered webpage image (generated by an MLLM) against its corresponding ground-truth image. To operationalize this evaluation, we designed a comprehensive rubric consisting of ten distinct visual dimensions. For each pair of images, the model provides a quantitative score on a scale of 0 to 10 for each dimension, where a score of 10 signifies perfect identity and a score of 0 indicates no discernible similarity.

The ten evaluation criteria are defined as follows:

\begin{description}
    \item[1. Element Reproduction:] Assesses whether all key visual elements (e.g., text, images, buttons, icons) are fully present and styled identically to the original.
    \item[2. Proportion and Size Consistency:] Evaluates if the relative and absolute sizes of all elements match the original design, maintaining the intended visual harmony.
    \item[3. Layout and Typography Fidelity:] Examines the faithful replication of the overall page structure (e.g., headers, footers, content grids) and the consistent application of typography (font families, weights).
    \item[4. Alignment and Spacing Accuracy:] Measures the precision of element alignment (e.g., left, center, relative to other elements) and the consistency of spacing (margins, padding) compared to the ground-truth design.
    \item[5. Visual Hierarchy Clarity:] Assesses if the generated webpage successfully maintains the same visual hierarchy as the original, guiding the user's attention to key information and calls to action similarly.
    \item[6. Color Consistency:] Evaluates the match of the overall color scheme, including primary, secondary, and accent colors, as well as specific hues and saturation levels.
    \item[7. Style Consistency:] Judges whether the overall aesthetic style of the generated webpage (e.g., modern, minimalistic, brutalist) aligns with the intended ``look and feel'' of the original design.
    \item[8. Text Style Consistency:] Focuses specifically on typographic attributes such as font family, size, weight, color, line height, and text alignment.
    \item[9. Text Content Accuracy:] Evaluates if the primary textual content (headings, body text, labels) accurately reproduces the text from the original design without omissions or significant alterations.
    \item[10. Overall Content Representation:] A holistic measure of whether the generated page effectively conveys the same core information, message, and user intent as the original.
\end{description}

To apply these criteria and collect the scores, the prompt shown in Figure~\ref{fig:gemini_evaluate_prompt} was provided to the Gemini 2.5 Flash model.

\subsection{Code Score: Formulation and Components}
\label{sec:appendix_code_score}
Our Code Score evaluates the similarity between an MLLM-generated HTML document ($H_{gen}$) and a reference HTML document ($H_{ref}$). The process involves parsing both documents into Document Object Model (DOM) trees, extracting associated CSS, and then performing a weighted aggregation of several similarity aspects.

\paragraph{1. Structural Similarity}
Both $H_{gen}$ and $H_{ref}$ are first parsed into DOM trees. We then extract sequences of HTML tags, $S_{gen}$ and $S_{ref}$ respectively, representing the structural hierarchy of the documents. The structural similarity ($Sim_{struct}$) is quantified by the ratio of the length of the Longest Common Subsequence (LCS) of these tag sequences to the length of the reference sequence, $S_{ref}$. A threshold ($\theta_{match} = 0.9$ in our implementation) is used within the LCS calculation to determine if two tags are considered similar enough to be part of a common subsequence.
\begin{equation}
Sim_{struct} = \frac{\text{LCS\_Length}_{\theta_{match}}(S_{gen}, S_{ref})}{\text{Length}(S_{ref})}
\end{equation}
If Length($S_{ref}$) is zero, $Sim_{struct}$ is defined as 1.0.

\paragraph{2. Content-Type Similarity}
This assesses similarity for three key content types: text, images, and forms. For each type, corresponding elements are identified and compared.

\subparagraph{Element Matching}
For each content type $c \in \{\text{text, image, form}\}$, we extract all elements of that type from $H_{gen}$ (denoted $E_{gen,c}$) and $H_{ref}$ (denoted $E_{ref,c}$). A matching algorithm identifies optimal corresponding pairs $(e_{gen}, e_{ref})$ between $E_{gen,c}$ and $E_{ref,c}$ based on type-specific similarity measures (detailed below) and a matching threshold ($\theta_{match} = 0.9$). This process yields a set of matched pairs $M_c$.

\subparagraph{Implementation Rate}
For each content type $c$, an implementation rate ($Rate_c$) is calculated. This reflects the proportion of reference elements found and successfully matched in the generated HTML:
\begin{equation}
Rate_c = \frac{|M_c|}{|E_{ref,c}|}
\end{equation}
If $|E_{ref,c}|$ is zero, $Rate_c$ is 1.0.

\subparagraph{Similarity Scores for Matched Elements}
For each matched pair $(e_{gen}, e_{ref}) \in M_c$:
\begin{itemize}
    \item \textbf{Text Elements ($c=\text{text}$):}
    Similarity is assessed based on:
    \begin{enumerate}
        \item Content Similarity ($Sim_{text\_content}(e_{gen}, e_{ref})$): Calculated using Python's SequenceMatcher on the textual content extracted.
        \item Style Similarity ($Sim_{text\_style}(e_{gen}, e_{ref})$): Computed by comparing key CSS properties (e.g., color, font-size, font-weight, background-color, etc.). Each property $p$ has a weight $w_p$. The style similarity is a weighted average of individual property similarities. Numerical properties (e.g., sizes) are compared using a ratio, while string properties use SequenceMatcher.
    \end{enumerate}
    The average $Sim_{text\_content}$ and $Sim_{text\_style}$ are calculated over all matched text elements.

    \item \textbf{Image Elements ($c=\text{image}$):}
    Similarity ($Sim_{image}(e_{gen}, e_{ref})$) is a weighted combination of:
    \begin{enumerate}
        \item URL Similarity (0.6 weight): Based on comparing extracted category information from the image `src` attribute (e.g., ``Animal'' from ``.../Animal.jpg'') or filenames if the category pattern doesn't match.
        \item Style Similarity (0.3 weight): Calculated similarly to text styles, using image-specific CSS properties (e.g., width, height, border-radius, as per self.style\_weights['image']).
        \item Alt Text Similarity (0.1 weight): Comparing the `alt` attributes using SequenceMatcher.
    \end{enumerate}
    The average $Sim_{image}$ is calculated over all matched image elements.

    \item \textbf{Form Elements ($c=\text{form}$):}
    Similarity ($Sim_{form}(e_{gen}, e_{ref})$) depends on the specific form element type (e.g., `input`, `button`). It's generally a weighted combination of:
    \begin{enumerate}
        \item Attribute Similarity: Compares critical HTML attributes specific to the form element type (e.g., `type`, `name`, `value`, `placeholder` for `input` elements) using SequenceMatcher.
        \item Style Similarity: Calculated using form-specific CSS properties (e.g., width, height, background-color).
        \item Text Content Similarity (only for elements like `button`, `label`, `option`): Compares textual content using SequenceMatcher.
    \end{enumerate}
    The average $Sim_{form}$ is calculated over all matched form elements.
\end{itemize}

\subparagraph{Adjusted Similarity Scores}
The average similarity scores for each content aspect are then adjusted by their respective implementation rates to penalize incompleteness:
\begin{align}
Sim'_{text\_content} &= \overline{Sim_{text\_content}} \times Rate_{text} \\
Sim'_{text\_style} &= \overline{Sim_{text\_style}} \times Rate_{text} \\
Sim'_{image} &= \overline{Sim_{image}} \times Rate_{image} \\
Sim'_{form} &= \overline{Sim_{form}} \times Rate_{form}
\end{align}
where $\overline{Sim}$ denotes the average similarity for matched elements of that type.

\paragraph{3. Final Code Score Aggregation}
The final Code Score ($Score_{code}$) is a weighted sum of the structural similarity and the adjusted content-type similarities. To ensure rigorous reproducibility, we explicitly specify the weight values used in our benchmark:
\begin{itemize}
    \item \textbf{Structure:} $W_{struct} = 0.25$
    \item \textbf{Text:} $W_{text\_content} = 0.20$, $W_{text\_style} = 0.10$
    \item \textbf{Visual Elements:} $W_{image} = 0.20$, $W_{form} = 0.25$
\end{itemize}

These weights are substituted into the aggregation formula as follows:
\begingroup\small
\begin{equation}
\begin{gathered}
Score_{code} = 0.25\, Sim_{struct} \\
{}+ 0.20\, Sim'_{text\_content} \\
{}+ 0.10\, Sim'_{text\_style} \\
{}+ 0.20\, Sim'_{image} \\
{}+ 0.25\, Sim'_{form}.
\end{gathered}
\end{equation}
\endgroup

This multi-faceted Code Score provides a nuanced evaluation of the generated HTML, considering its structural integrity, content accuracy, stylistic fidelity, and overall completeness across different element types. These parameters are also strictly defined in our open-sourced evaluation scripts.

\begin{table*}[htbp]
\centering
\begin{tabular}{l ccccc}
\toprule
\textbf{Metric} & \textbf{Ref} & \textbf{Img} & \textbf{Inter} & \textbf{Text} & \textbf{Average}\\
\hdashline
Gemini Visual Score & \textbf{0.9394} & \textbf{0.9273} & 0.9394 & \textbf{0.9394} & \textbf{0.9364} \\
Code Score & 0.8909 & \uline{0.9152} & \uline{0.9515} & \uline{0.9030} & \uline{0.9152} \\
Clip Score & \textbf{0.9394} & \uline{0.9152} & \textbf{0.9879} & 0.7333 & 0.8939 \\
DINOv2 Score & \uline{0.9030} & 0.9030 & 0.9394 & 0.8061 & 0.8879 \\
\bottomrule
\end{tabular}
\caption{Spearman's Correlation Coefficient ($\rho$) between Automatic Metrics and Human Preference Rankings. A higher value indicates a stronger alignment with human judgment. The best-performing metric in each column is marked in \textbf{bold}, and the second-best is \uline{underlined}.}
\label{tab:metric_validity}
\end{table*}

\subsection{Metric Validity}
\label{sec:appendix_metric_validity}

To rigorously assess the reliability of our proposed automatic evaluation metrics, we conducted a comprehensive validation against human judgments using \textbf{Spearman's rank correlation analysis}. This non-parametric statistical method is ideal for our purpose as it measures the strength and direction of the monotonic relationship between two sets of ranked data, without assuming a linear relationship between them. A high positive correlation coefficient ($\rho$) would signify that a given automatic metric ranks the performance of different Multimodal Large Language Models (MLLMs) in a manner highly consistent with human evaluators' preferences.

The validation process was executed as follows:
\begin{enumerate}
    \item \textbf{Data Ranking:} For each of the four subtasks within the Webpage Code Generation benchmark (Code Refinement, Image to Code, Interaction Authoring, and Text to Code), we established two sets of rankings for all evaluated MLLMs. The first set was derived from the human evaluation scores presented in Table \ref{tab:human_eval}. The second set was generated by ordering the models based on their performance under each automatic metric (Code Score, Gemini Visual Score, Clip Score, and DINOv2 Score), as detailed in Table \ref{tab:code_results}.
    \item \textbf{Correlation Calculation:} We then computed the Spearman's rank correlation coefficient ($\rho$) between the human preference ranking and each automatic metric's ranking. This calculation was performed independently for each of the four subtasks to analyze the metrics' performance in different contexts.
    \item \textbf{Average Correlation:} Finally, to determine the overall reliability of each metric across the entire benchmark, we calculated the average correlation coefficient across all four subtasks.
\end{enumerate}

The results of this analysis are summarized in Table \ref{tab:metric_validity}. The findings unequivocally demonstrate that our proposed metrics serve as excellent proxies for human assessment.

As presented in Table \ref{tab:metric_validity}, our proposed \textbf{Gemini Visual Score} achieves the highest average correlation with human preferences across all tasks, with an impressive $\rho = 0.9364$. This indicates an exceptionally strong monotonic relationship, confirming that the Gemini Visual Score is a highly robust and reliable indicator of generation quality as perceived by humans. It performs particularly well on the Ref, Img, and Text subtasks, achieving the best or tied-best correlation scores of $\rho = 0.9394$, $\rho = 0.9273$, and $\rho = 0.9394$, respectively. These results demonstrate its strong capability to capture nuanced visual and semantic attributes that are critical to human evaluators.

Similarly, our custom-designed \textbf{Code Score} also demonstrates outstanding reliability, achieving the second-highest average correlation of $\rho = 0.9152$. It performs particularly well on the Interaction Authoring subtask, where it achieves a strong correlation of $\rho = 0.9515$, and also maintains consistently high correlations on the Img and Text subtasks, with $\rho = 0.9152$ and $\rho = 0.9030$, respectively. This highlights its effectiveness in evaluating both structural fidelity and functional implementation. Concurrently, the established \textbf{Clip Score} and \textbf{DINOv2 Score} also prove their merit as valuable supplementary evaluation tools, achieving strong average correlations of $\rho = 0.8939$ and $\rho = 0.8879$, respectively. These strong positive correlations indicate that they can provide complementary perspectives to our primary scores. Overall, this robust statistical validation underscores the credibility and effectiveness of the FullFront benchmark's evaluation framework.
\begin{table*}[htbp]
    \centering
{\scriptsize
    \begin{tabular}{@{}p{0.18\textwidth} p{0.18\textwidth} p{0.20\textwidth} p{0.34\textwidth}@{}}
        \toprule
        \textbf{Model} & \textbf{Parameter Setting} & \textbf{Source} & \textbf{URL} \\
        \midrule
        Qwen2.5-VL-72B-Instruct & temperature = 0.0 & local checkpoint & \url{https://huggingface.co/Qwen/Qwen2.5-VL-72B-Instruct} \\
        \midrule
        InternVL2.5-78B & temperature = 0.0 & local checkpoint & \url{https://huggingface.co/OpenGVLab/InternVL2_5-78B} \\
        \midrule
        InternVL3-78B & temperature = 0.0 & local checkpoint & \url{https://huggingface.co/OpenGVLab/InternVL3-78B}\\
        \midrule
        LLaVA-Onevision-72B & temperature = 0.0 & local checkpoint & \url{https://huggingface.co/llava-hf/llava-onevision-qwen2-72b-ov-hf} \\
        \midrule
        R1-Distill-Llama-8b & temperature = 0.0 & local checkpoint & \url{https://huggingface.co/deepseek-ai/DeepSeek-R1-Distill-Llama-8B} \\
        \midrule
        Claude 3.7 Sonnet & temperature = 0.0 & claude-3-7-sonnet-20250219 & \url{https://www.anthropic.com/} \\
        \midrule
        Gemini 2.5 Flash & \begin{tabular}[t]{@{}l@{}}temperature = 0.0 \\ thinking\_budget = 0\end{tabular} & gemini-2.5-flash-preview-04-17 & \url{https://ai.google.dev/} \\
        \midrule
        GPT-4o & temperature = 0.0 & gpt-4o-2024-11-20 & \url{https://platform.openai.com}\\
        \midrule
        o4-mini & \makecell[l]{reasoning\_effort=medium\\(default)} & o4-mini-2025-04-16 & \url{https://platform.openai.com} \\
        \midrule
        GPT-4.1 & temperature = 0.0 & gpt-4.1-2025-04-14 & \url{https://platform.openai.com}\\
        \midrule
        GPT-5 & default & gpt-5-2025-08-07 & \url{https://platform.openai.com}\\
        \midrule
        Gemini 2.5 Pro & \begin{tabular}[t]{@{}l@{}}temperature = 0.0 \\ thinking\_budget = 1024\end{tabular} & gemini-2.5-pro-preview-05-06 & \url{https://ai.google.dev/} \\
        \midrule
        o1 & \makecell[l]{reasoning\_effort=medium\\(default)} & o1 & \url{https://platform.openai.com} \\
        \midrule
        \begin{tabular}[t]{@{}l@{}}gemini-2.0-flash-exp-image \\ -generation\end{tabular}
        & - &
        \begin{tabular}[t]{@{}l@{}}gemini-2.0-flash-exp-image \\ -generation\end{tabular}
        & \url{https://ai.google.dev/} \\
        \midrule
        DeepSeek-V3 & temperature=0.0 & deepseek-v3-250324 & \url{https://www.deepseek.com} \\
        \midrule
        DeepSeek-R1 & temperature=0.0 & deepseek-r1-250528 & \url{https://www.deepseek.com} \\
        \midrule
        DeepSeek-V3.1 & temperature=0.0 & deepseek-v3-1-250821 & \url{https://www.deepseek.com} \\
        \midrule
        Kimi-K2 & temperature=0.0 & kimi-k2-0711-preview & \url{https://api.moonshot.cn/v1} \\
        \midrule
    \end{tabular}
    }
\caption{The sources of models used in the experiments and the hyperparameters configuration.}
\label{tab:model_parameters_my_table}

\end{table*}

\subsection{MLLM Evaluation Settings}
\label{Evaluation Settings}
The experimental evaluation of the four open-source Multimodal Large Language Models (MLLMs) featured in this study---specifically, Qwen2.5-VL-72B-Instruct, InternVL2.5-78B, InternVL3-78B, and LLaVA-Onevision-72B---is conducted on a high-performance computing infrastructure. For the inference phase, these models are benchmarked utilizing a configuration of 48 CPU cores and 4 NVIDIA A100-SXM4-80GB GPUs. Our development and evaluation environment is built upon PyTorch version 2.6.0+cu118, leveraging CUDA version 11.8 for GPU acceleration. In line with our commitment to transparency and to facilitate further research within the community, all code developed for the FullFront benchmark, including model evaluation pipelines and custom metric implementations, has been made publicly available. The official sources for all evaluated models are comprehensively listed in Table \ref{tab:model_parameters_my_table}. In adherence to best practices for reproducible research, hyperparameters were typically maintained at their default settings during model evaluation. Detailed hyperparameter configurations for each model can be found in Table~\ref{tab:model_parameters_my_table}.

\subsection{Human Evaluation}
This section details the methodologies for the human evaluations referenced in the main paper. We first outline the standardized rating interface and protocol used for scoring the visual fidelity of generated webpages against their ground-truth counterparts. Following this, we describe a supplementary comparative study designed to provide a more nuanced assessment by pitting MLLM-generated webpages against their original, real-world versions across multiple qualitative dimensions.
\subsubsection{Evaluation Methodology and Rating Interface}
To ensure a consistent and reproducible scoring process, human evaluators were provided with a standardized rating environment.

\begin{figure*}[t]
  \centering
  \includegraphics[width=0.98\textwidth]{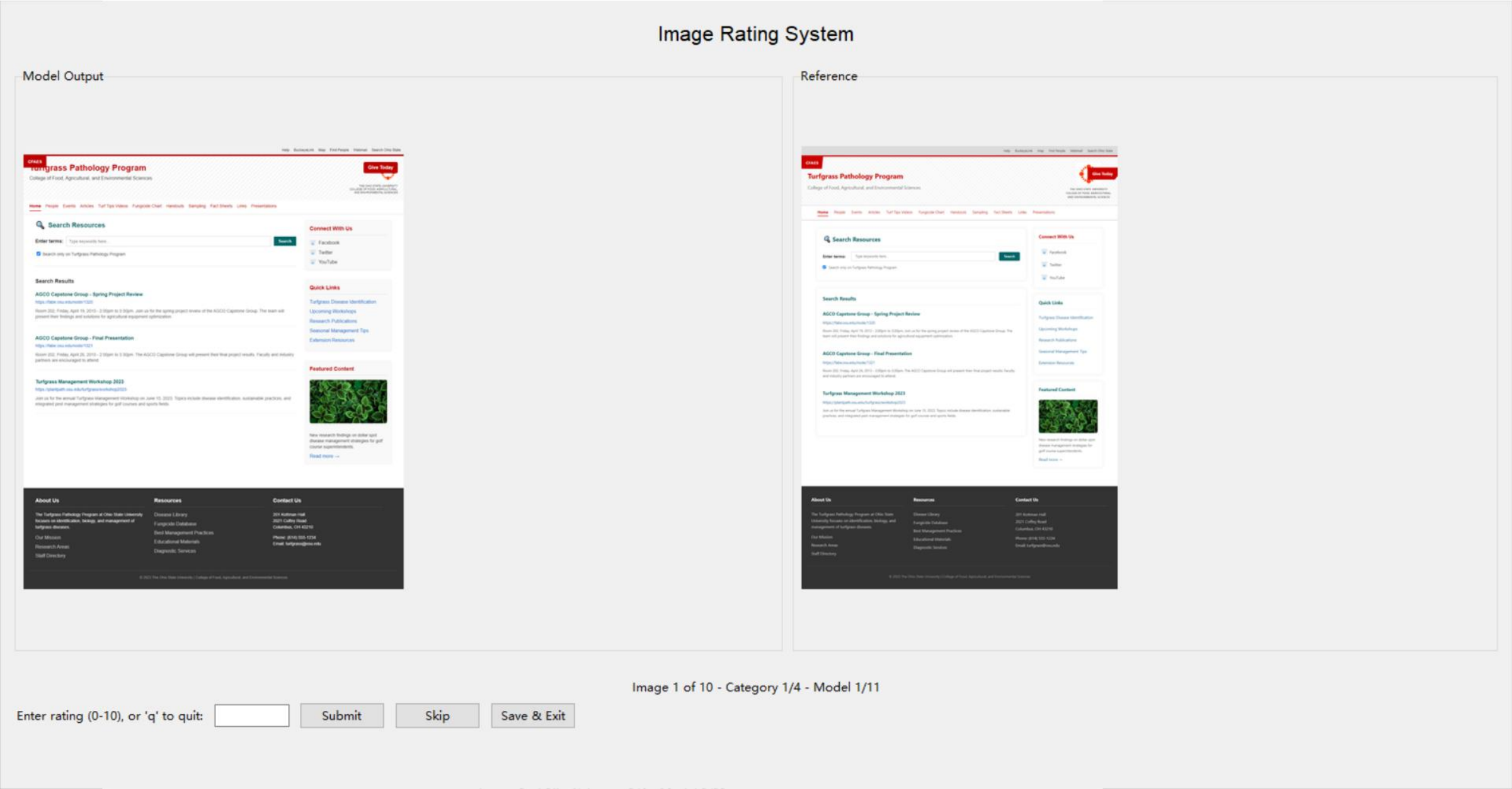}
  \caption{The human evaluation interface for image rating.}
  \label{fig:rating_interface_appendix}
\end{figure*}

The rating tool is a self-contained application developed in Python, with a graphical user interface (GUI) built using Tkinter. Upon launch, the interface presents two images side-by-side, as illustrated in Figure~\ref{fig:rating_interface_appendix}. The left image displays the MLLM-generated output, while the right image serves as the ground-truth reference.

\textbf{Key features of the rating process:}
\begin{enumerate}
    \item \textbf{Blind Evaluation:} Crucially, evaluators are not informed of the specific model that generated the output they are assessing at any given time, thereby ensuring a blind evaluation process to mitigate potential biases.
    \item \textbf{Scoring Scale:} A numerical rating scale from 0 to 10 (inclusive) is used, where 0 indicates complete inconsistency with the ground truth or specified criteria, and 10 signifies perfect adherence. Decimal scores are permitted to allow for finer-grained assessments.
    \item \textbf{Workflow:} Evaluators submit scores via the ``Enter'' key or a ``Submit'' button. Progress is saved automatically, allowing evaluators to pause and resume their work seamlessly. All ratings are stored in a JSON file for subsequent analysis.
\end{enumerate}

The evaluators' assessments are guided by the same detailed rubric used for the Gemini Visual Score (as outlined in Appendix~\ref{sec:appendix_gemini_visual_score}). The final 0-10 score for each generated image is a holistic rating based on ten criteria grouped into three categories: Visual Structure, Color and Aesthetics, and Textual Content.

\subsubsection{Comparative Evaluation Against Real-World Webpages}
\label{appendix_human_evaluation_info_aes}

\begin{figure*}[t]
  \centering
  \includegraphics[width=0.98\textwidth]{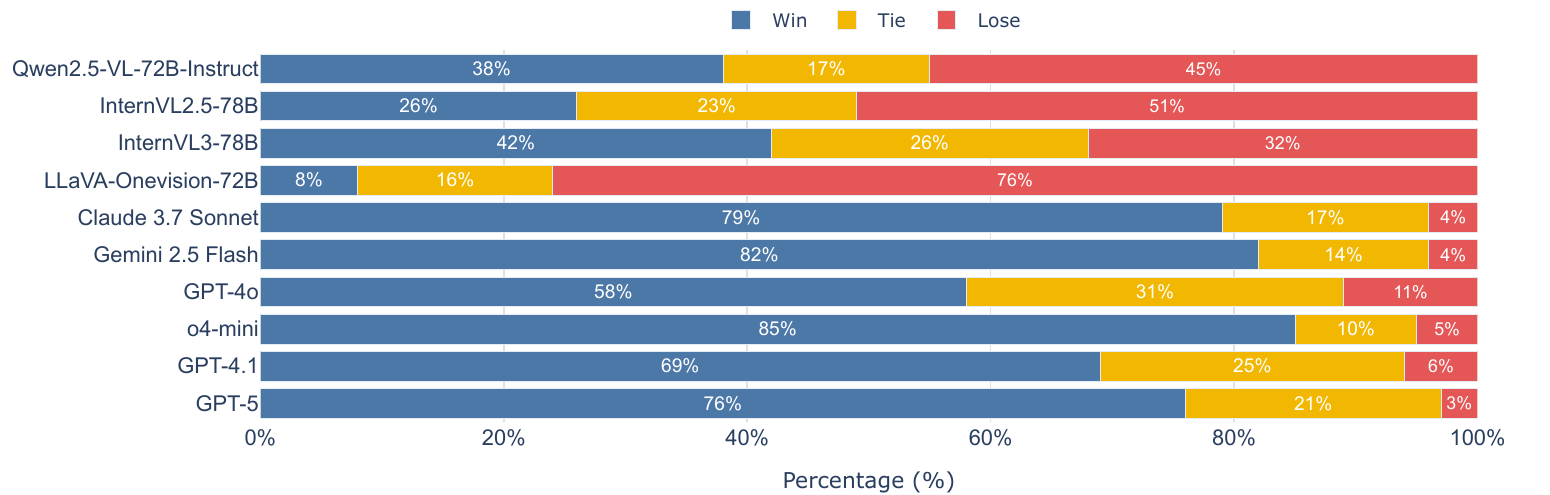}
  \caption{Human evaluation comparing MLLM-generated and Real-World webpages.}
  \label{fig:win_lose}
\end{figure*} 

Figure~\ref{fig:win_lose} illustrates a direct comparison of human preference between MLLM-generated webpages and their real-world counterparts. However, a simple assertion of ``superiority'' over real-world examples can be imprecise and subject to bias; for instance, evaluators might prefer the clean, generic templates produced by MLLMs over cluttered, complex source websites. To investigate this with greater nuance, we conducted a supplementary human evaluation designed to separate preference across multiple dimensions.

We randomly sampled 100 of the 200 Image-to-Code source pages and used this same page set for all eight models in Table~\ref{tab:human_evaluation_info_aes}. We recruited five evaluators from diverse professional backgrounds. For every model, each evaluator blindly compared all 100 generated pages with their corresponding original real-world pages, yielding 500 judgments per model per criterion and 4,000 judgments per criterion across the eight models. The study therefore evaluates rendered-page quality on the Image-to-Code subset; it does not pool the other Code Generation subtasks into its denominator. Each pair received a ``win--tie--loss'' judgment along two distinct axes:
\begin{enumerate}
    \item \textbf{Aesthetic Appeal:} Which page is more visually attractive and demonstrates a higher quality of design?
    \item \textbf{Information Effectiveness:} Which page conveys its core message more clearly and efficiently guides the user toward potential tasks?
\end{enumerate}

\begin{table*}[htbp]
\centering
\begin{tabular}{l cccccc}
\toprule
\textbf{Model} & \textbf{Aes(w)} & \textbf{Aes(t)}  & \textbf{Aes(l)} & \textbf{Inf(w)} & \textbf{Inf(t)} & \textbf{Inf(l)}\\
\hdashline 
Qwen2.5-VL-72B-Instruct & 15.2 & 34.4 & 50.4 & 13.8 & 38.4 & 47.8 \\
InternVL2.5-78B & 18.6 & 28.8 & 52.6 & 29.8 & 38.2 & 32.0 \\
InternVL3-78B & 23.6 & 33.4 & 43.0 & 34.6 & 34.0 & 31.4 \\
LLaVA-Onevision-72B & 6.0 & 10.8 & 83.2 & 7.6 & 25.0 & 67.4 \\
\hdashline
Claude 3.7 Sonnet & 80.0 & 15.6 & 4.4 & 62.8 & 29.0 & 8.2 \\
Gemini 2.5 Flash & 75.4 & 16.6 & 8.0 & 63.2 & 27.8 & 9.0 \\
GPT-4o & 38.6 & 43.6 & 17.8 & 29.0 & 54.6 & 16.4 \\
o4-mini & 69.0 & 21.6 & 9.4 & 77.6 & 14.0 & 8.4 \\
\bottomrule
\end{tabular}
\caption{Human Preference for MLLM vs. Real-World Webpages on Aesthetic and Informational Criteria.}
\label{tab:human_evaluation_info_aes}

\end{table*}

The aggregated results, presented in Table~\ref{tab:human_evaluation_info_aes}, offer a more granular perspective. On this sampled set, leading MLLMs like Claude 3.7 Sonnet (80.0\% win rate) are strongly preferred for Aesthetic Appeal. This is consistent with evaluators favoring the clean, modern, and often minimalistic designs generated by these models.

However, the advantage is less pronounced in Information Effectiveness. While top models still outperform, the gap narrows considerably. This suggests that while MLLM-generated designs are visually pleasing, they may not always match the purpose-driven, information-rich layouts of real-world websites, which have often been iteratively optimized for user experience. This multi-dimensional analysis reframes the discussion from simple ``superiority'' to a more nuanced understanding of the specific strengths (aesthetics) and potential trade-offs (information architecture) of current MLLM-based front-end generation.

\section{Investigation of Data-Pipeline Bias}
\label{sec:construction_bias_analysis}

A valid concern in evaluating Multimodal Large Language Models (MLLMs) is the potential for data-pipeline bias. Since our primary benchmark dataset utilizes GPT-4o for initial generation and Claude 3.7 Sonnet for refinement, there is a risk that the evaluation might implicitly favor these models due to stylistic overfitting or shared latent patterns in the data distribution.

\subsection{Experimental Setup}
To scrutinize this potential bias and verify that FullFront evaluates fundamental front-end engineering capabilities rather than pipeline-specific artifacts, we conducted a controlled experiment using an independent data generation engine. We employed \textbf{Grok-4}, a model entirely distinct from those used in our main pipeline, to construct a new evaluation dataset. This ``Grok-4 Set'' comprises 300 samples, evenly distributed across three core subtasks: Code Refinement (\textbf{Ref}), Image to Code (\textbf{Img}), and Text to Code (\textbf{Text}). We then evaluated six representative models on this unseen dataset.

\subsection{Results and Analysis}
The detailed evaluation results are presented in Table~\ref{tab:grok_detailed_results}. Our analysis focuses on two key dimensions: the consistency of model performance hierarchies and the validity of our automated metrics on this new data distribution.

\textbf{Consistency of Model Rankings:}
Despite the shift in the data generation source to Grok-4, the performance hierarchy of the evaluated models remains consistent with our main benchmark results.
As shown in the ``Human Score'' columns, \textbf{GPT-5} and \textbf{Claude 3.7 Sonnet} occupy the top two positions across the three tested subtasks (e.g., GPT-5 achieves a Human Score of 8.52 on Ref and 6.78 on Text). \textbf{Gemini 2.5 Flash} follows, while the evaluated open-source models score lower, particularly on the visual-fidelity tasks (Ref and Img). This result shows ranking stability under one alternative proprietary reconstruction pipeline. Because Grok-4 is also a proprietary model and the study omits Interaction Authoring, it mitigates but does not eliminate concerns about model-generated reference distributions or establish equivalence to human-authored production code.

\textbf{Metric Validity and Alignment with Human Judgment:}
To verify the reliability of our automated metrics on this independent dataset, we calculated the Spearman rank correlation coefficient ($\rho$) between the model rankings produced by each automated metric and those produced by human experts for each subtask (labeled as IRR in Table~\ref{tab:grok_detailed_results}).
The results demonstrate that our proposed metrics maintain high robustness:
\begin{itemize}
    \item \textbf{Gemini Visual Score:} Exhibits the strongest alignment with human judgment, achieving Spearman correlations of \textbf{0.89}, \textbf{0.94}, and \textbf{0.83} across the Code Refinement, Image to Code, and Text to Code tasks, respectively. This confirms its utility as a reliable proxy for human visual assessment even on unseen data distributions.
    \item \textbf{Code Score:} Also shows strong correlation, particularly in the Refinement ($\rho=0.89$) and Image to Code ($\rho=0.89$) tasks, validating its effectiveness in measuring structural and content fidelity.
    \item \textbf{Limitations of Pure Visual Embeddings:} Interestingly, while CLIP and DINOv2 scores correlate well in image-based tasks, their correlation drops significantly in the \textit{Text to Code} task ($\rho=0.31$ and $0.37$). This further highlights the necessity of our specialized Code and Gemini scores for a comprehensive evaluation of front-end generation.
\end{itemize}

The replicated ranking pattern and metric correlations on the Grok-4 set provide evidence that the main results are not unique to the original GPT-4o--Claude 3.7 Sonnet construction pipeline. We interpret this as a partial robustness check under an alternative proprietary generator, not as a non-model reference anchor or proof that all pipeline-specific bias has been removed.

\begin{table*}[htbp]
    \vspace{-2pt}
    \centering
\resizebox{\textwidth}{!}{
    \begin{tabular}{l *{15}{c}}
        \toprule
        \multirow{2}{*}{\textbf{Model}}
        & \multicolumn{3}{c}{\textbf{Code Score}} & \multicolumn{3}{c}{\textbf{Gemini Visual Score}} & \multicolumn{3}{c}{\textbf{CLIP Score}} & \multicolumn{3}{c}{\textbf{DINOv2 Score}} & \multicolumn{3}{c}{\textbf{Human Score}} \\
        \cmidrule(lr){2-4} \cmidrule(lr){5-7} \cmidrule(lr){8-10} \cmidrule(lr){11-13} \cmidrule(lr){14-16}
        & \textbf{Ref} & \textbf{Img} & \textbf{Text} & \textbf{Ref} & \textbf{Img} & \textbf{Text} & \textbf{Ref} & \textbf{Img} & \textbf{Text} & \textbf{Ref} & \textbf{Img} & \textbf{Text} & \textbf{Ref} & \textbf{Img} & \textbf{Text} \\
        \midrule
        Qwen2.5-VL-72B-Instruct & 0.27 & 0.35 & 0.34 & 4.05 & 4.85 & 5.83 & 0.69 & 0.69 & 0.76 & 0.48 & 0.47 & 0.57 & 6.16 & 4.16 & 5.62 \\
        InternVL3-78B & 0.27 & 0.34 & 0.26 & 5.93 & 5.96 & 5.08 & 0.72 & 0.72 & 0.73 & 0.45 & 0.51 & 0.54 & 5.97 & 5.03 & 5.75 \\
        Claude 3.7 Sonnet & \uline{0.54} & \uline{0.51} & \textbf{0.41} & \textbf{8.79} & \uline{8.60} & \uline{6.24} & \uline{0.80} & \uline{0.79} & 0.74 & 0.68 & \uline{0.67} & 0.54 & \uline{8.30} & \textbf{8.52} & \uline{6.46} \\
        Gemini 2.5 Flash & 0.48 & 0.48 & 0.38 & 8.07 & 8.27 & 6.20 & \uline{0.80} & \uline{0.79} & \uline{0.78} & \uline{0.69} & \uline{0.67} & \uline{0.62} & 7.87 & 7.62 & 6.15 \\
        GPT-4o & 0.53 & 0.47 & 0.33 & 7.17 & 6.78 & 6.19 & 0.76 & 0.74 & 0.75 & 0.53 & 0.52 & 0.59 & 6.38 & 5.76 & 5.69 \\
        GPT-5 & \textbf{0.63} & \textbf{0.60} & \textbf{0.41} & \uline{8.45} & \textbf{8.73} & \textbf{6.51} & \textbf{0.83} & \textbf{0.86} & \textbf{0.79} & \textbf{0.70} & \textbf{0.74} & \textbf{0.64} & \textbf{8.52} & \uline{8.47} & \textbf{6.78} \\
        \hline
        \textbf{Spearman $\rho$ (IRR)} & 0.89 & 0.89 & 0.77 & 0.89 & 0.94 & 0.83 & 0.89 & 0.83 & 0.31 & 0.94 & 0.94 & 0.37 & 1.00 & 1.00 & 1.00 \\
        \bottomrule
    \end{tabular}
    }
\caption{Performance of six representative models on the new \textbf{Grok-4 constructed dataset} (300 samples). We report detailed scores across three subtasks: Code Refinement (\textbf{Ref}), Image to Code (\textbf{Img}), and Text to Code (\textbf{Text}). The last row, \textbf{Spearman $\rho$ (IRR)}, represents the rank correlation between the specific metric and Human Score across the 6 models for that subtask.}
\label{tab:grok_detailed_results}

    \vspace{-10pt}
\end{table*}

\section{Analysis of Model-as-Judge Bias}
\label{sec:judge_bias_analysis}

The deployment of Multimodal Large Language Models (MLLMs) as evaluators introduces the specific challenge of ``same-family bias,'' wherein a judge may disproportionately favor models from the same developer. To preemptively quantify and rule out such artifacts within our framework, we conducted a cross-judge validation experiment to verify the robustness and impartiality of our chosen evaluator, Gemini 2.5 Flash.

\subsection{Experimental Setup}
We curated a stratified subset of 200 samples from the FullFront benchmark, selecting 50 samples from each of the four Code Generation subtasks to ensure diverse coverage. We compared the scoring behaviors of Human Experts against four automated judges: our primary judge \textbf{Gemini 2.5 Flash}, two other proprietary models (\textbf{Claude 3.7 Sonnet}, \textbf{Grok 4}), and a state-of-the-art open-source model, \textbf{Qwen3-VL-235B-A22B-Instruct}.

We assessed the judges based on two primary dimensions:
\begin{itemize}
    \item \textbf{Inter-rater Reliability (IRR):} We employed Spearman's rank correlation coefficient ($\rho$) and Intraclass Correlation Coefficient (ICC) to measure sample-level agreement with human scores. Additionally, we calculated the average score deviation (Bias) to identify systematic scaling differences.
    \item \textbf{Ranking Consistency:} We used Kendall's $\tau$ to evaluate the alignment of the aggregated model leaderboards produced by each judge against the human-derived leaderboard.
    \item \textbf{Stability:} We performed three independent inference runs (with temperature=0 where applicable) to measure test-retest reliability via standard deviation ($\sigma$) and inter-run correlation.
\end{itemize}

\subsection{Results and Analysis}

\textbf{Absence of Self-Preference Bias.}
The results presented in Table~\ref{tab:merged_scores} provide compelling evidence against self-preference bias in the Gemini 2.5 Flash judge.
First, the model demonstrates strict objectivity by ranking itself consistently below GPT-5 and Claude 3.7 Sonnet, achieving a perfect alignment with human ranking (Kendall's $\tau = 1.0$).
Second, while Gemini 2.5 Flash tends to assign higher absolute scores than human evaluators, this inflation is \textit{systematic} rather than \textit{selective}. As detailed in Table~\ref{tab:merged_scores}, the bias toward its own outputs ($+0.38$) is effectively identical to the bias exhibited toward its primary competitors, GPT-5 ($+0.37$) and Claude 3.7 ($+0.39$).

Crucially, this pattern of \textit{uniform deviation} is corroborated by other automated judges, confirming that such offsets reflect a judge's inherent strictness rather than preferential treatment. For instance, the Claude 3.7 Sonnet judge exhibits a uniform strictness, systematically subtracting approximately $0.84$ points across the top three models (GPT-5: $-0.83$, Claude: $-0.84$, Gemini: $-0.86$). Conversely, the Qwen3-VL judge displays a lenient baseline similar to Gemini's, adding approximately $0.45$ points uniformly (GPT-5: $+0.42$, Claude: $+0.49$, Gemini: $+0.46$). The consistency of these relative gaps across diverse judges confirms that Gemini's scoring reflects a stable, objective standard---merely shifted by a constant calibration factor---rather than an algorithmic attempt to artificially boost its relative standing.

\textbf{Robustness of the Scoring Rubric.}
It is notable that all four automated judges achieved a perfect ranking alignment with human experts (Kendall's $\tau = 1.0$). This unanimity suggests that our fine-grained scoring rubric (detailed in Appendix G.1) successfully guides diverse models to replicate human evaluation logic, reducing the dependence on model-specific priors and ensuring that the evaluation criteria are model-agnostic.

\textbf{Viability of Open-Source Judges.}
Our experiment identifies \textbf{Qwen3-VL-235B-A22B-Instruct} as a highly capable open-source alternative for automated evaluation. It demonstrates reliability metrics (ICC=0.851, Spearman $\rho$=0.764) that are comparable to, and in some cases competitive with, proprietary models. This finding offers a reproducible and accessible evaluation pathway for the research community.

\textbf{Judge Stability.}
Table~\ref{tab:stability} summarizes the test-retest reliability. Gemini 2.5 Flash exhibits exceptional stability with a negligible standard deviation ($\sigma < 0.1$) and a high inter-run correlation ($0.965$). Similarly, Qwen3-VL-235B-A22B-Instruct demonstrates remarkable consistency ($0.972$). In contrast, Grok 4 shows significant variance ($\sigma \approx 1.10$), indicating that despite its high ranking capability, its sample-level scoring is less deterministic.

In conclusion, our cross-validation confirms that the use of Gemini 2.5 Flash as a judge does not introduce same-family bias regarding model rankings. Furthermore, the systematic nature of its scoring behavior, combined with high stability, validates its suitability for the \textsc{FullFront} benchmark.

\begin{table*}[htbp]
    \centering
\resizebox{\textwidth}{!}{
    \begin{tabular}{lccccc}
    \toprule
    \textbf{Evaluated Model} & \textbf{Human Judge} & \textbf{Gemini 2.5 Flash Judge} & \textbf{Claude 3.7 Sonnet Judge} & \textbf{Grok 4 Judge} & \textbf{Qwen3-VL Judge} \\
    \midrule
    GPT-5 & 8.50 (R1) & 8.87 (+0.37, R1) & 7.67 (-0.83, R1) & 9.02 (+0.52, R1) & 8.92 (+0.42, R1) \\
    Claude 3.7 Sonnet & 8.35 (R2) & 8.73 (+0.39, R2) & 7.51 (-0.84, R2) & 8.97 (+0.63, R2) & 8.83 (+0.49, R2) \\
    Gemini 2.5 Flash & 8.23 (R3) & 8.61 (+0.38, R3) & 7.38 (-0.86, R3) & 8.92 (+0.68, R3) & 8.70 (+0.46, R3) \\
    GPT-4o & 6.67 (R4) & 6.72 (+0.05, R4) & 5.93 (-0.74, R4) & 8.19 (+1.52, R4) & 7.34 (+0.67, R4) \\
    Qwen2.5-VL-72B & 5.96 (R5) & 5.98 (+0.02, R5) & 5.31 (-0.65, R5) & 8.13 (+2.17, R5) & 6.51 (+0.55, R5) \\
    InternVL3-78B & 5.49 (R6) & 5.34 (-0.15, R6) & 4.96 (-0.53, R6) & 7.98 (+2.50, R6) & 6.13 (+0.65, R6) \\
    \midrule
    \textbf{Spearman $\rho$ (IRR)} & 1.00 & \textbf{0.769} & \textbf{0.805} & 0.361 & \textbf{0.764} \\
    \textbf{ICC (Consistency)} & 1.00 & \textbf{0.846} & \textbf{0.873} & 0.457 & \textbf{0.851} \\
    \textbf{Kendall's $\tau$ (Rank)} & 1.00 & \textbf{1.000} & \textbf{1.000} & \textbf{1.000} & \textbf{1.000} \\
    \bottomrule
    \end{tabular}
    }
\caption{\textbf{Cross-Judge Validation Results.} Aggregated performance across 200 representative samples. The table compares human expert scores against four automated judges. Format: \textbf{Score (Bias vs. Human, Rank)}. \textit{Bias} indicates the average deviation from human scores. \textit{Inter-rater Reliability (IRR)} is measured by Spearman $\rho$ and Intraclass Correlation (ICC) against human ground truth. Kendall's $\tau$ measures the alignment of the resulting model leaderboard with the human-derived leaderboard.}
\label{tab:merged_scores}

\end{table*}

\begin{table*}[htbp]
    \centering
\resizebox{0.8\textwidth}{!}{
    \begin{tabular}{lcc}
    \toprule
    \textbf{Judge Model} & \textbf{Avg. Score $\pm$ Std Dev} & \textbf{Inter-Run Correlation} \\
    \midrule
    Gemini 2.5 Flash & 7.38 $\pm$ 0.080 & 0.965 \\
    Claude 3.7 Sonnet & 6.46 $\pm$ 0.300 & 0.876 \\
    Grok 4 & 8.53 $\pm$ 1.097 & 0.410 \\
    Qwen3-VL-235B-A22B-Instruct & 7.74 $\pm$ 0.079 & 0.972 \\
    \bottomrule
    \end{tabular}
    }
\caption{\textbf{Test-Retest Reliability of Automated Judges.} Metrics are calculated across three independent inference runs to assess the stability of the judges. High Inter-Run Correlation indicates that the judge produces consistent scores for the same input across different trials.}
\label{tab:stability}

\end{table*}

\section{Sensitivity and Robustness Analysis of the Code Score}
\label{sec:code_score_sensitivity}

\begin{table*}[ht]
    \centering
\renewcommand{\arraystretch}{1.3} 

    \resizebox{0.95\textwidth}{!}{
        \begin{tabular}{l c c c c}
            \toprule
            \multirow{2}{*}{\textbf{Experiment Condition}} & \textbf{Gemini Visual Score} $\uparrow$ & \textbf{Code Score} $\uparrow$ & \textbf{CLIP Score} $\uparrow$ & \textbf{DINOv2 Score} $\uparrow$ \\
             & \small{(0-10 scale)} & \small{(0-1 scale)} & \small{(0-1 scale)} & \small{(0-1 scale)} \\
            \midrule

            \textbf{Baseline} (Original HTML) & 10.00 & 1.00 & 1.00 & 1.00 \\
            \midrule

            \makecell[l]{\textit{Exp 1: Valid Alternative}}
            & 9.266 & 0.952 & 0.969 & 0.942 \\

            \hspace{1em}\textbf{Penalty ($\Delta_1$)} & \textbf{-0.734} & \textbf{-0.048} & \textbf{-0.031} & \textbf{-0.058} \\
            \midrule

            \makecell[l]{\textit{Exp 2: Broken Implementation}}
            & 6.838 & 0.727 & 0.846 & 0.760 \\

            \hspace{1em}\textbf{Penalty ($\Delta_2$)} & \textbf{-3.162} & \textbf{-0.273} & \textbf{-0.154} & \textbf{-0.240} \\
            \midrule

            \textbf{Punishment Ratio ($\Delta_2 / \Delta_1$)} & 4.31x & \textbf{5.73x} & 5.02x & 4.17x \\
            \bottomrule
        \end{tabular}
    }
\caption{Sensitivity Analysis of Evaluation Metrics on 50 Selected Samples. We quantify the score penalty ($\Delta$) imposed by different metrics on ``Syntactic Variations'' (Exp 1, valid alternatives) versus ``Semantic Violations'' (Exp 2, breaking errors). Code Score has the largest punishment ratio among the tested metrics, while its non-zero valid-alternative penalty confirms that it remains reference-relative.}
\label{tab:metric-sensitivity}
\end{table*}

A potential concern regarding the \textsc{Code Score} is its reliance on fixed structural and stylistic weights, which can penalize valid engineering alternatives (e.g., different CSS layout techniques that yield identical visual results). The metric's model-ranking correlation with human judgment averages $\rho=0.9152$ across the four subtasks (Appendix~\ref{sec:appendix_metric_validity}), but this does not make it invariant to render-equivalent implementations. We therefore conducted a controlled sensitivity analysis to quantify its tolerance for valid variations versus its sensitivity to substantive defects.

\subsection{Experimental Design}
We selected a random subset of 50 samples from the ground truth dataset. For each sample, we generated two distinct variants of the code to simulate different implementation qualities:
\begin{itemize}
    \item \textbf{Experiment 1: Syntactic Variation (Valid Alternative).} We refactored the HTML/CSS code using alternative implementation patterns---specifically, swapping layout techniques (e.g., replacing Flexbox with CSS Grid or vice versa) and adjusting nesting structures---while rigorously preserving the original visual appearance. This condition tests the metric's ability to accept diverse but correct engineering solutions.
    \item \textbf{Experiment 2: Semantic Violation (Degradation).} We introduced structural and content-level errors, such as removing container elements, altering text content, or omitting essential style attributes. These changes were designed to cause significant semantic and visual degradation, representing ``broken'' implementations.
\end{itemize}

\subsection{Results and Discriminative Power}
The results of this analysis are summarized in Table~\ref{tab:metric-sensitivity}. We define the \textit{Penalty} ($\Delta$) as the drop in score compared to the original ground truth (Score = 1.0).

As observed, the \textsc{Code Score} retains a mean score of 0.952 for valid syntactic variations, corresponding to a residual penalty of $\Delta_1=-0.048$. In contrast, broken implementations score 0.727 on average, corresponding to a substantially larger penalty of $\Delta_2=-0.273$. The valid-alternative penalty is small relative to the broken condition but is not zero and can be comparable to differences between adjacent model point estimates.

To summarize the relative separation, we calculate the \textbf{Punishment Ratio} ($\Delta_2 / \Delta_1$), which measures how much more severely a metric penalizes the tested broken implementations than the tested valid variations. Code Score achieves a ratio of \textbf{5.73x}, compared with 5.02x for CLIP Score and 4.17x for DINOv2 Score. This sensitivity analysis supports using Code Score as a reference-relative structural/content-fidelity signal, while the residual valid-alternative penalty motivates reporting it alongside render-level, executable, and human measures rather than treating it as semantic correctness.

\section{Intervention Analysis of Perception--Implementation Decoupling}
\label{sec:intervention_analysis}

The observational analysis in Section~\ref{sec:perception_code_relationship} identifies two off-diagonal behaviors: models may perceive the queried relation correctly but implement it incorrectly ($p\checkmark c\times$), or answer the perception question incorrectly while implementing the original relation correctly ($p\times c\checkmark$). We conduct two controlled interventions to probe these behaviors without changing the generation prompt, sampling configuration, or requested full-page output.

\subsection{Crop-Assisted Evidence Localization}

For each selected $p\checkmark c\times$ case, we retain the original full-page screenshot and add an oracle crop containing the region relevant to the QA item. The model is still required to regenerate the entire page, and success is judged only on whether the queried local attribute---such as alignment, width, or relative position---is corrected. The crop therefore supplies localized visual evidence without removing global page context. As shown in Table~\ref{tab:intervention_results}, the queried implementation is corrected in 22 of 28 cases for Qwen2.5-VL-72B-Instruct and 18 of 23 cases for InternVL3-78B. This indicates that evidence localization can mitigate a substantial fraction of the implementation bottleneck for these models.

\subsection{Counterfactual Local Edits}

For each selected $p\times c\checkmark$ case, we create a counterfactual input by modifying only the queried local relation while leaving the rest of the page unchanged. The edits alter attributes such as width, position, or alignment and are newly constructed for this analysis. We then regenerate the complete page with the same prompt and generation settings. ``Not following the edit'' means that the target local relation is not implemented; it does not require the remainder of the generated page to be identical to the original output.

Gemini 2.5 Flash returns to the original or a more canonical relation in 18 of 25 cases, while GPT-5 does so in 21 of 31 cases (Table~\ref{tab:intervention_results}). Representative cases illustrate both outcomes. In one case, GPT-5 ignores a subtle downward displacement of a Search button and recreates the canonical aligned input--button pair. In a positive case with a more salient structural change, it correctly follows an upward displacement of the middle feature block. Together, these results show that $p\times c\checkmark$ on the original image is not sufficient evidence of fine-grained grounding. The newly edited inputs weaken an exact-instance memorization explanation, while the frequent return to canonical layouts is consistent with distribution-level layout priors; the experiment does not claim to identify a unique internal mechanism.

\begin{table*}[t]
    \centering
    \small
    \begin{tabular}{llrrr}
        \toprule
        \textbf{Intervention} & \textbf{Model} & \textbf{Eligible Cases} & \textbf{Target Outcome} & \textbf{Rate} \\
        \midrule
        QA-relevant crop ($p\checkmark c\times$) & Qwen2.5-VL-72B-Instruct & 28 & 22 corrected & 78.6\% \\
        QA-relevant crop ($p\checkmark c\times$) & InternVL3-78B & 23 & 18 corrected & 78.3\% \\
        \midrule
        Counterfactual local edit ($p\times c\checkmark$) & Gemini 2.5 Flash & 25 & 18 edits ignored & 72.0\% \\
        Counterfactual local edit ($p\times c\checkmark$) & GPT-5 & 31 & 21 edits ignored & 67.7\% \\
        \bottomrule
    \end{tabular}
    \caption{Controlled interventions on the two off-diagonal perception--implementation behaviors. For the crop intervention, the target outcome is correction of the queried local implementation. For the counterfactual intervention, the reported outcome is failure to follow the newly edited local relation.}
    \label{tab:intervention_results}
\end{table*}

\section{Evaluation of Newer Frontier Models}
\label{sec:frontier_model_analysis}

\subsection{Setup and Coverage}

To assess whether FullFront retains headroom for newer systems, we additionally evaluate GPT-5.5, Claude Opus 4.8, Qwen3.5-122B-A10B, and Qwen3.5-397B-A17B on the same full evaluation sets used by the corresponding main experiments: all 1,800 Perception QA items, 200 Image-to-Code items, and 200 Text-to-Code items. We use the same prompts, rendering environment, and metric definitions as in the original evaluation. The additional evaluation does not include Code Refinement or Interaction Authoring, so conclusions about the four newer models are limited to these five tested settings.

\begin{table*}[t]
    \centering
    \setlength{\tabcolsep}{3.2pt}
    \resizebox{\textwidth}{!}{
    \begin{tabular}{l ccc cccc cccc}
        \toprule
        \multirow{2}{*}{\textbf{Model}} & \multicolumn{3}{c}{\textbf{Perception QA Accuracy}} & \multicolumn{4}{c}{\textbf{Image-to-Code}} & \multicolumn{4}{c}{\textbf{Text-to-Code}} \\
        \cmidrule(lr){2-4} \cmidrule(lr){5-8} \cmidrule(lr){9-12}
        & \textbf{Real} & \textbf{Synth.} & \textbf{Multi.} & \textbf{Code} & \textbf{GVS} & \textbf{CLIP} & \textbf{DINOv2} & \textbf{Code} & \textbf{GVS} & \textbf{CLIP} & \textbf{DINOv2} \\
        \midrule
        Qwen2.5-VL-72B-Instruct & 0.4696 & 0.4950 & 0.4267 & 0.40 & 4.48 & 0.72 & 0.49 & 0.49 & 5.18 & 0.73 & 0.47 \\
        InternVL3-78B & 0.4816 & 0.5375 & 0.4600 & 0.42 & 4.47 & 0.73 & 0.53 & 0.50 & 4.91 & 0.72 & 0.44 \\
        Claude 3.7 Sonnet & 0.5464 & 0.5325 & 0.4533 & 0.64 & 8.93 & 0.89 & 0.81 & 0.64 & 8.14 & 0.87 & 0.78 \\
        GPT-5 & 0.5816 & 0.5775 & 0.5467 & 0.56 & 8.58 & 0.88 & 0.81 & 0.59 & 8.48 & 0.88 & 0.79 \\
        \midrule
        Qwen3.5-122B-A10B & 0.5848 & 0.5925 & 0.5667 & 0.55 & 7.46 & 0.84 & 0.73 & 0.57 & 7.25 & 0.82 & 0.69 \\
        Qwen3.5-397B-A17B & 0.6280 & 0.6425 & 0.6133 & 0.63 & 8.76 & 0.89 & 0.80 & 0.65 & 8.27 & 0.87 & 0.78 \\
        Claude Opus 4.8 & 0.8424 & \textbf{0.8625} & 0.8267 & \textbf{0.86} & 9.31 & 0.93 & 0.88 & \textbf{0.82} & \textbf{9.38} & \textbf{0.92} & \textbf{0.88} \\
        GPT-5.5 & \textbf{0.8680} & 0.8475 & \textbf{0.8533} & 0.83 & \textbf{9.56} & \textbf{0.95} & \textbf{0.91} & 0.76 & 9.12 & \textbf{0.92} & 0.86 \\
        \bottomrule
    \end{tabular}}
    \caption{Results for four newer frontier models and four representative original models on the shared full evaluation sets. GVS denotes Gemini Visual Score. QA accuracy, Code Score, CLIP, and DINOv2 range from 0 to 1; GVS ranges from 0 to 10. Bold indicates the best value in each column.}
    \label{tab:new_frontier_models}
\end{table*}

\subsection{Headroom Analysis}

The newer proprietary models substantially improve on the original systems, confirming that the evaluated QA, Image-to-Code, and Text-to-Code settings are less difficult for the latest frontier generation. Nevertheless, none of the reported metrics reaches its ceiling: the best QA accuracy is 0.8680, the best Code Score is 0.86, and the best Gemini Visual Score is 9.56/10. The newer open models also improve markedly. Qwen3.5-397B-A17B reaches 0.6280/0.6425/0.6133 on Real-world/Synthetic/Multi-window QA and 0.63/0.65 Code Score on Image-to-Code/Text-to-Code, while retaining larger error margins than GPT-5.5 and Claude Opus 4.8. These results show measurable headroom on the tested settings, especially for open models, but should not be extrapolated to the two untested Code Generation subtasks.

\subsection{Release Scope}
\label{sec:release_scope}

The complete camera-ready release will contain the constructed benchmark data (200 Webpage Design items, 1,800 Perception QA items, and 700 Code Generation items) and retained model outputs under the CC0 1.0 Universal license. The data-construction and evaluation code will be distributed under the MIT License. The project repository is available at \url{https://github.com/Mikivishy/FullFront}, and the dataset is hosted at \url{https://huggingface.co/datasets/Mikivis/FullFront}. The release documentation will distinguish the full benchmark from FullFront-mini and record task coverage for each evaluated model.

\clearpage
\section{Prompts}
\label{appendix_prompts}
In this section, we present the specific prompts used in our benchmark, covering both the data curation and evaluation phases.

\begin{figure*}[htbp] 
    \centering 
\begin{wideprompt}
\begin{promptbox}{Prompt -- GPT-4o Generate QA Pairs}

You are now a master of front-end page design. The user will provide you with two images: \\
- The first image: a full-page screenshot of a certain webpage.\\
- The second image: labeled boxes (icons, images, text, etc.) detected on that webpage, along with their corresponding positions, annotations, or OCR results. The two images and the corresponding boxes information are as follows.:\\

\{\textcolor{red}{Webpage Screenshot}\}\\
\{\textcolor{red}{Webpage Screenshot with Boxes}\}\\
\{\textcolor{red}{Omniparser Extracted Results}\}\\

Your goal is:\\
1. Based on the information provided by these two images, focus on analyzing the webpage's layout, visual elements, UI components, and their positions on the page.\\
2. Combine the boxes information from the second image to gain a more detailed understanding and summary of the webpage's elements.\\
3. On this basis, design and generate **two** multiple-choice questions (with correct answers). These questions should assess the understanding and ability in the following areas:\\
   - GUI Grounding: Precise grasp of the position, size, color, and relative relationships of each element on the page.\\
   - Overall understanding of page design and front-end implementation: Such as typography, font selection, color matching, and module division.\\
Please note:\\
1. The questions you design should be challenging, especially focusing on the more subtle and inconspicuous parts of the page, such as small text, slight color differences, spacing between elements, alignment, and dimensions.\\
2. You can examine how UI components are arranged on the page, interactive elements, or margins that are not easily noticeable at first glance.\\
3. Your questions should be challenging and not easily answered.\\
4. The choices you design should not have significant differences; the correct answer should not be obviously different from the other options. Each option should have a considerable level of confusion.\\
5. Each question must be accompanied by a correct answer.\\
Your output should strictly follow this format, and "correct answer" should only be the letter of the option, such as A or B or C or D, without any other elements:\\
{"Q1": "question text1", "Choices1": [ "A. choice A", "B. choice B", "C. choice C", "D. choice D" ], "A1": "correct answer", "Q2": "question text2", "Choices2": [ "A. choice A", "B. choice B", "C. choice C", "D. choice D" ], "A2": "correct answer"}
    \end{promptbox}
\end{wideprompt}
    \caption{Prompt -- GPT-4o Generate QA Pairs.}
    \label{fig:child_specifications}
\end{figure*}

\begin{figure*}[p]
    \centering
\begin{wideprompt}
\begin{promptbox}{Prompt -- GPT-4o Generate HTML-v1}
You are a web developer proficient in HTML, CSS, and JavaScript. The user will provide you with two images. Please carefully analyze the elements, layout, and other relevant details in the first image to generate the necessary HTML, CSS, and JavaScript code to replicate the webpage design shown in that image.
The second image contains various boxes detected on the webpage image, along with their position information and OCR results, as follows:\\

\{\textcolor{red}{Webpage Screenshot}\}\\
\{\textcolor{red}{Webpage Screenshot with Boxes}\}\\
\{\textcolor{red}{Omniparser Extracted Results}\}\\

If any images are required in the page code, you must classify each image into one of the following categories: People, Animal, Food, Plant, Landscape, Icon, Logo, Architecture, Technology, Transportation, Map, Texture, Art, Movie, Other. The URL for each image is in the format: \url{https://fixed_part/{category}.jpg}. For example, if you want to use an image from the Animal category, the image URL would be: \url{https://fixed_part/Animal.jpg}. Since the width and height of each image in the URLs are unknown, you need to set the image sizes manually in the HTML code to ensure that the rendered result matches the layout requirements of the webpage image.\\
Please combine the HTML, CSS, and JavaScript codes into one file and provide the full code only. The output must start with <!DOCTYPE html> and end with </html>.
    \end{promptbox}
    \end{wideprompt}
    \caption{Prompt -- GPT-4o Generate HTML-v1.}
    \label{fig:prompt_1}
\end{figure*}

\begin{figure*}[p]
    \centering
\begin{wideprompt}
\begin{promptbox}{Prompt -- Claude 3.7 Sonnet Generate HTML-v2}
The user will provide you with HTML code. Based on the existing HTML code, please make improvements and enhancements in terms of layout, text elements, visual components, color schemes, and decorative details to enhance the overall aesthetics and richness of the page. You can add new text content, appropriate icons or illustrations, or incorporate more layered styles. Please ensure that the final style of the page is unified and that the visual elements are well-coordinated, making the static page look both beautiful and not monotonous. The HTML code is as follows:'''\\

\{\textcolor{red}{HTML-v1}\}\\

If any images are required in the page code, you must classify each image into one of the following categories: People, Animal, Food, Plant, Landscape, Icon, Logo, Architecture, Technology, Transportation, Map, Texture, Art, Movie, Other. The URL for each image is in the format: \url{https://fixed_part/{category}.jpg}. For example, if you want to use an image from the Animal category, the image URL would be: \url{https://fixed_part/Animal.jpg}. Since the width and height of each image in the URLs are unknown, you need to set the image sizes manually in the HTML code to ensure that the rendered result matches the layout requirements of the webpage image. \\
Please combine the HTML, CSS, and JavaScript codes into one file and provide the full code only. The output must start with <!DOCTYPE html> and end with </html>.
    \end{promptbox}
    \end{wideprompt}
    \caption{Prompt -- Claude 3.7 Sonnet Generate HTML-v2.}
    \label{fig:prompt_2}
\end{figure*}

\begin{figure*}[p]
    \centering
\begin{wideprompt}
\begin{promptbox}{Prompt -- Claude 3.7 Sonnet Generate Webpage Description}
You are now a seasoned front-end development expert, proficient in HTML, CSS, and JavaScript. Users will provide a screenshot of a webpage and the OCR result of the webpage. The OCR result includes text content with corresponding box coordinates. You need to analyze the various elements, layout, and possible interactive effects of the webpage in detail and accurately based on this screenshot and OCR result. Please describe the page using clear and detailed language, covering the following points:\\
1. \textbf{Overall Page Layout}: Describe the general structure of the page, such as whether it uses responsive design, whether the layout is single-column or multi-column (e.g., two-column layout, three-column layout, etc.), and the approximate proportions and positional relationships of each column.\\
2. \textbf{Element Types and Attributes}:
   - \textbf{Text Elements:} Indicate the font, font size, color, boldness, italics, and other styles of the text content, as well as the text alignment (left-aligned, centered, etc.). Include the OCR text content in your description of each text element to provide complete information about the text displayed on the page.\\
   - \textbf{Image Elements:} Explain the size and position of the image, as well as the alignment of the image.\\
   - \textbf{Button Elements}: Describe the shape, size, color, background color, border style, text content, and font style of the button.\\
   - \textbf{Other Elements:} Such as forms (input boxes, drop-down menus, etc.), navigation bars, sidebars, footers, etc., provide a detailed description of their styles, positions, and functions.\\
3. \textbf{Layout Details}:\\
   - \textbf{Spacing}: Describe the spacing between elements (such as margins and padding).\\
   - \textbf{Alignment}: Explain the horizontal and vertical alignment of elements.\\
   - \textbf{Layering}: Point out the layering relationships between elements, such as which elements are on the upper layer and which are on the lower layer.\\
You need to categorize all the images on the page into one of the following fifteen categories and clearly state which category each image belongs to: People, Animal, Food, Plant, Landscape, Icon, Logo, Architecture, Technology, Transportation, Map, Texture, Art, Movie, Other.\\
Your description should be detailed enough for users to accurately recreate this static webpage using HTML, CSS, and JavaScript code based on your description. Please use professional terminology as much as possible to ensure the accuracy and operability of the description. Note that while the OCR results include box coordinates, you should incorporate the text content into your description without mentioning the coordinates.
\textbf{OCR Result}:\\

\{\textcolor{red}{Omniparser Extracted Results}\}\\
\{\textcolor{red}{Webpage Screenshot}\}\\

    \end{promptbox}
    \end{wideprompt}
    \caption{Prompt -- Claude 3.7 Sonnet Generate Webpage Description.}
    \label{fig:prompt_3}
\end{figure*}

\begin{figure*}[p]
    \centering
\begin{wideprompt}
\begin{promptbox}{Prompt -- Webpage Design Evaluation}

You are now a professional web designer. Please generate the corresponding webpage image based on the user's description of a webpage.

\{\textcolor{red}{Webpage Description}\}\\

    \end{promptbox}
    \end{wideprompt}
    \caption{Prompt -- Webpage Design Evaluation.}
    \label{fig:prompt_4}
\end{figure*}

\begin{figure*}[p]
    \centering
\begin{wideprompt}
\begin{promptbox}{Prompt -- Webpage Perception QA Evaluation}

  \textbf{Real-world QA:}
    The user will provide you with an image and a question related to that image. You must Answer with the option's letter from the given choices and put the letter in one \verb|"\\boxed{}"|, for example \verb|\\boxed{A}|.\\
    \{\textcolor{red}{Question: \{question\}}\}\\
    \{\textcolor{red}{Webpage Screenshot}\}\\
\\
\\
  \textbf{Synthetic QA:}
    The user will provide you with an image and a question related to that image. You must Answer with the option's letter from the given choices and put the letter in one \verb|"\\boxed{}"|, for example \verb|\\boxed{A}|.\\
    \{\textcolor{red}{Question: \{question\}}\}\\
    \{\textcolor{red}{Webpage Screenshot}\}\\
\\
\\
  \textbf{Multi-window QA:}
    The user will provide you with an image consisting of multiple web page screenshots, as well as a question related to the image. You must Answer with the option's letter from the given choices and put the letter in one \verb|"\\boxed{}"|, for example \verb|\\boxed{A}|.\\
    \{\textcolor{red}{Question: \{question\}}\}\\
    \{\textcolor{red}{Webpage Screenshot}\}\\
    \end{promptbox}
    \end{wideprompt}
    \caption{Webpage Perception QA Evaluation.}
    \label{fig:prompt_5}
\end{figure*}

\begin{figure*}[p]
    \centering
\begin{wideprompt}
\begin{promptbox}{Prompt -- Image to Code Evaluation}
The user will provide you with a screenshot of a webpage. Please generate the corresponding HTML, CSS, and JavaScript code based on the image to fully replicate the webpage.\\
If any images are required in the page code, you must classify each image into one of the following categories: People, Animal, Food, Plant, Landscape, Icon, Logo, Architecture, Technology, Transportation, Map, Texture, Art, Movie, Other. The URL for each image is in the format: \url{https://fixed_part/{category}.jpg}. For example, if you want to use an image from the Animal category, the image URL would be: \url{https://fixed_part/Animal.jpg}. Since the width and height of each image in the URLs are unknown, you need to set the image sizes manually in the HTML code to ensure that the rendered result matches the layout requirements of the webpage image.\\
Please place the HTML, CSS, and JavaScript code in a single file, and your response must be a code that starts with `<!DOCTYPE html>` and ends with `</html>`, without any other content.\\
    \{\textcolor{red}{Webpage Screenshot}\}\\
    \end{promptbox}
    \end{wideprompt}
    \caption{Prompt -- Image to Code Evaluation.}
    \label{fig:prompt_6}
\end{figure*}

\begin{figure*}[p]
    \centering
\begin{wideprompt}
\begin{promptbox}{Prompt -- Interaction Authoring Evaluation}

You are a web developer proficient in HTML, CSS, and JavaScript. The user will provide you with two web page screenshots. The first screenshot shows the original state of the web page, while the second screenshot shows the result after the user has performed a specific operation on a particular element on the page. Please carefully analyze the differences between the two images and determine the type of interaction. You must set the ID of the interactive element to ``\#InteractionPart'', so please ensure that this ID is unique and that the element is interactive. The possible types of interactions and the requirements for each type are as follows:\\

\{\textcolor{red}{Click Interaction Prompt}\} or \{\textcolor{red}{Hover Interaction Prompt}\}\\

You need to decide for yourself what kind of interaction to use, and you must meet the requirements for the implementation of each type of interaction.\\
If any images are required in the page code, you must classify each image into one of the following categories: People, Animal, Food, Plant, Landscape, Icon, Logo, Architecture, Technology, Transportation, Map, Texture, Art, Movie, Other. The URL for each image is in the format: \url{https://fixed_part/{category}.jpg}. For example, if you want to use an image from the Animal category, the image URL would be: \url{https://fixed_part/Animal.jpg}. Since the width and height of each image in the URLs are unknown, you need to set the image sizes manually in the HTML code to ensure that the rendered result matches the layout requirements of the webpage image.\\
Your code should accurately reflect the interaction and restore the overall static appearance of the page, ensuring the changes in the screenshots are faithfully reproduced. Please place the HTML, CSS, and JavaScript code in a single file, and your response must be a code that starts with `<!DOCTYPE html>` and ends with `</html>`, without any other content.\\

\{\textcolor{red}{Webpage Screenshot (Before Interaction)}\}\\
\{\textcolor{red}{Webpage Screenshot (After Interaction)}\}\\

    \end{promptbox}
    \end{wideprompt}
    \caption{Prompt -- Interaction Authoring Evaluation.}
    \label{fig:prompt_7}
\end{figure*}

\begin{figure*}[p]
    \centering
\begin{wideprompt}
\begin{promptbox}{Prompt -- Click Interaction}

\textbf{Interaction click 1}\\
- Function Description: When the element with id="\#InteractionPart" is clicked, a dropdown box should be displayed. The content of the dropdown should be generated based on the understanding of the page content, and its position, size, and style should be adapted according to the page structure.\\
- The element must include the attribute aria-expanded="false", which changes to aria-expanded="true" upon clicking.\\
- The dropdown content must use one of the following selectors: .dropdown-menu, .dropdown-content, or [role='menu'].\\
- Ensure that the dropdown box is hidden before clicking and visible after clicking.\\
\textbf{Interaction click 2}\\
- Function Description: Generate a checkbox that can be checked by clicking, with its id='\#InteractionPart'. The associated text content and position should be adapted based on the layout and context of the page. Ensure that the checkbox can toggle its state (checked/unchecked) when clicked.\\
- Must use an <input type="checkbox"> element or an element with the attribute [role='checkbox'].\\
- After clicking, it must display a checked state (:checked or aria-checked='true').\\
\textbf{Interaction click 3}\\
- Function Description: Implement an interactive element (id="\#InteractionPart") that changes its background color significantly when clicked. The color change should be determined based on the page design style, but it must be detectable.\\
- The initial background color must not be transparent or white.\\
- The background color must change significantly after clicking (it cannot change from one transparency to another).\\
\textbf{Interaction click 4}\\
- Function Description: Implement an interactive element (id="\#InteractionPart") that, when clicked, triggers the display of a modal window or dialog box. The content of the popup should be generated based on the existing content of the page and should match the page style.\\
- After clicking, an element that matches one of the following selectors must be displayed: .modal, .dialog, [role='dialog'], or [aria-modal='true'].\\
- The modal window must be hidden before clicking.\\
\textbf{Interaction click 5}\\
- Function Description: Implement an interactive element (id="\#InteractionPart") that, when clicked, displays a tooltip providing additional information or explanation. The position, style, and content of the tooltip should be adapted according to the needs of the page.\\
- After clicking, one of the following elements must be displayed: an element with the class .tooltip or .tip; an element with the attribute role='tooltip' or data-tooltip; an element with an ID ending in Tooltip or tooltip; an element with the class .tooltip-text inside \#InteractionPart.\\
\textbf{Interaction click 6}\\
- Function Description: Implement an interactive element (id="\#InteractionPart") that, when clicked, displays a text box or input area for users to enter content. The position and size of the text box should be adjusted according to the page design and should include a clear input prompt or label.\\
- After clicking, one of the following elements must be displayed: <input type='text'> or <input type='email'>; <textarea>; an element with the attribute [contenteditable='true']; an element with the attribute [role='textbox']; or the container with the ID newsletterContainer becomes visible.\\

    \end{promptbox}
    \end{wideprompt}
    \caption{Prompt -- Click Interaction}
    \label{fig:prompt_8}
\end{figure*}

\begin{figure*}[p]
    \centering
\begin{wideprompt}
\begin{promptbox}{Prompt -- Hover Interaction}

\textbf{Interaction hover 1}\\
- Function Description: Implement an interactive element (id="\#InteractionPart") that displays a dropdown box when the mouse hovers over it. The content of the dropdown should be generated based on the understanding of the page content, and its position and display effect should adapt to the page layout.\\
- When hovering, an element that matches one of the following selectors must be displayed: .dropdown-menu, .dropdown-content, [role='menu'], or the element with the ID placementDropdown.\\
- The dropdown element must be hidden before hovering.\\
\textbf{Interaction hover 2}\\
- Function Description: Implement an interactive element (id="\#InteractionPart") containing text that becomes bold when the mouse hovers over it. Ensure that the bold effect is clearly displayed.\\
- When hovering, the text must become bold, with a fontWeight value of $\geq 600$ or set to bold/bolder.\\
\textbf{Interaction hover 3}\\
- Function Description: Implement an interactive element (id="\#InteractionPart") containing text that adds an underline when the mouse hovers over it. The underline effect should be consistent with the page style.\\
- When hovering, the computed style of the element's textDecoration must include "underline".\\
\textbf{Interaction hover 4}\\
- Function Description: Implement an interactive element (id="\#InteractionPart") that displays a tooltip when the mouse hovers over it, providing additional information or explanation. The content and style of the tooltip should be adapted according to the page context.\\
- When hovering, at least one of the following elements must be displayed: an element with the class .tooltip or .tip; a direct child element of the interactive element (\#InteractionPart) with the class .tooltip-text; an element with the attribute [role='tooltip']; an element with the attribute [data-tooltip].\\
- The tooltip element must be invisible before hovering.\\

    \end{promptbox}
    \end{wideprompt}
    \caption{Prompt -- Hover Interaction.}
    \label{fig:prompt_9}
\end{figure*}

\begin{figure*}[p]
    \centering
\begin{wideprompt}
\begin{promptbox}{Prompt -- Text to Code Evaluation}

The user will provide you with a comprehensive textual description of a webpage, covering its layout, elements, and other relevant details. Based on this description, please generate the corresponding front-end code (including HTML, CSS, and JavaScript) to closely replicate the webpage as described.\\
If any images are required in the page code, you must classify each image into one of the following categories: People, Animal, Food, Plant, Landscape, Icon, Logo, Architecture, Technology, Transportation, Map, Texture, Art, Movie, Other. The URL for each image is in the format: \url{https://fixed_part/{category}.jpg}. For example, if you want to use an image from the Animal category, the image URL would be: \url{https://fixed_part/{Animal}.jpg}. Since the width and height of each image in the URLs are unknown, you need to set the image sizes manually in the HTML code to ensure that the rendered result matches the layout requirements of the webpage image.\\
Please place the HTML, CSS, and JavaScript code in a single file, and your response must be a code that starts with `<!DOCTYPE html>` and ends with `</html>`, without any other content.\\

    \{\textcolor{red}{Webpage Screenshot Description}\}\\

    \end{promptbox}
\end{wideprompt}
    \caption{Prompt -- Text to Code Evaluation.}
    \label{fig:prompt_10}
\end{figure*}

\begin{figure*}[p]
    \centering
\begin{wideprompt}
\begin{promptbox}{Prompt -- Code Refinement Evaluation}

The user will provide you with a screenshot of a webpage, as well as a piece of front-end code. This code is the user's initial draft based on the webpage, but it cannot fully replicate the webpage. Please generate the front-end code that can fully replicate the webpage based on the screenshot and the existing code.\\
If any images are required in the page code, you must classify each image into one of the following categories: People, Animal, Food, Plant, Landscape, Icon, Logo, Architecture, Technology, Transportation, Map, Texture, Art, Movie, Other. The URL for each image is in the format: \url{https://fixed_part/{category}.jpg}. For example, if you want to use an image from the Animal category, the image URL would be: \url{https://fixed_part/{Animal}.jpg}. Since the width and height of each image in the URLs are unknown, you need to set the image sizes manually in the HTML code to ensure that the rendered result matches the layout requirements of the webpage image.\\
Please place the HTML, CSS, and JavaScript code in a single file, and your response must be a code that starts with `<!DOCTYPE html>` and ends with `</html>`, without any other content.\\

    \{\textcolor{red}{Webpage Screenshot}\}\\
    \{\textcolor{red}{Corresponding HTML-v1}\}\\

    \end{promptbox}
\end{wideprompt}
    \caption{Prompt -- Code Refinement Evaluation.}
    \label{fig:prompt_11}
\end{figure*}

\end{document}